\documentclass[10pt,twocolumn,letterpaper]{article}
\PassOptionsToPackage{table}{xcolor}
\usepackage[pagenumbers]{cvpr}
\DeclareRobustCommand{\ARCIR}{\ensuremath{\mathrm{ARC}_{\mathrm{IR}}}}
\usepackage{multirow}
\usepackage{tabularx}
\usepackage{pgfplots}
\pgfplotsset{compat=1.18}
\usetikzlibrary{positioning,arrows.meta,fit,shapes.geometric,calc,backgrounds}
\definecolor{frbg}{HTML}{CADDF5}
\definecolor{nrbg}{HTML}{FCEAE3}
\definecolor{cvprblue}{rgb}{0.21,0.49,0.74}
\usepackage[pagebackref,breaklinks,colorlinks,allcolors=cvprblue]{hyperref}
\hypersetup{pdfauthor={Jonas Brenig, Radu Timofte},pdftitle={Compression as an Action: Agentic Joint Restoration and Compression for Degraded Images}}
\newcommand{\toggleicon}{%
  \tikz[baseline=-0.5ex]{
    \draw[rounded corners=1.8pt, draw=orange!70!black, fill=orange!15, thick] (0,0) rectangle (0.5,0.22);
    \fill[orange!70!black] (0.37,0.11) circle (0.09);
  }%
}

\title{Compression as an Action:\\ Agentic Joint Restoration and Compression for Degraded Images}
\author{Jonas Brenig \qquad Radu Timofte\\
CAIDAS \& IFI CVL, University of W\"urzburg, Germany\\
{\tt\small \{jonas.brenig, radu.timofte\}@uni-wuerzburg.de}}

\begin{document}

\maketitle
\begin{abstract}
Lossy image compression and blind image restoration are typically optimized and executed separately. We introduce \emph{Agentic Restoration and Compression} (ARC), a modular framework that treats codec selection, codec placement, and codec operating point as actions within an agentic image-restoration workflow. Compression differs from restoration because it intentionally reduces some quality measures to lower the bitrate. Within the execution–reflection loop, ARC jointly selects the codec and quality level using an objective that balances local distortion, no-reference quality, and output-relative bitrate. A budgeted search limits codec evaluations, while separate acceptance thresholds for restoration and compression permit bounded quality losses. The transmitted representation contains the codec bitstream together with the decoder-side restoration plan. On MiO100, ARC improves reconstruction quality over standalone codecs and achieves the lowest PieAPP among the evaluated compression pipelines across all three degradation groups, while performing competitively across a broad set of other metrics. Experiments on clean Kodak images expose a failure mode in which unnecessary restoration reduces fidelity. The framework supports additional codecs and rate-distortion trade-offs without retraining the restoration or compression models.
\end{abstract}

\section{Introduction}

Image compression is essential for efficient storage and transmission of visual data. Traditional codecs such as JPEG~\cite{wallaceJPEGStillPicture1992}, VVC~\cite{brossOverviewVersatileVideo2021}, and learned image compression methods~\cite{balleVariationalImageCompression2018,heELICEfficientLearned2022,luLearnedImageCompression2025} achieve increasingly strong rate-distortion performance, but apply the same fixed compression scheme regardless of an image's degradations.

Concurrently, agentic image restoration frameworks have emerged as a powerful paradigm for handling complex, real-world image degradations. Systems such as AgenticIR~\cite{zhuIntelligentAgenticSystem2025} and 4KAgent~\cite{zuo4KAgentAgenticAny2026} leverage large language models (LLMs) and vision-language models (VLMs) to dynamically analyze image degradations, plan restoration sequences, and select from a toolbox of specialized restoration models. They follow an execution-reflection paradigm, iteratively applying restoration tools, evaluating intermediate results, and rolling back unsuccessful steps.
Despite this progress, compression and restoration remain separate: fixed compression schemes cannot adapt to individual images' degradations, while agentic restoration systems ignore downstream compression requirements. Naively combining the two fails to exploit the synergy between restoration and compression decisions.

In this work, we propose \textit{Agentic Restoration and Compression} (ARC), an agentic framework for image compression in which an agent plans and executes a sequence of compression and restoration actions tailored to each input image's degradations and a user-specified rate-quality trade-off. ARC offers a practical way to select heterogeneous codecs within restoration workflows and obtains encouraging results on independent perceptual metrics.

\paragraph{Positioning.}
Interactive compression agents such as Comp-X~\cite{gaoCompXDefiningInteractive2025} select coding
modes from user requests, while prior restoration agents
can only use restoration tools
\cite{zhuIntelligentAgenticSystem2025,zuo4KAgentAgenticAny2026}.
To our knowledge, our framework is the first to schedule a
lossy codec inside an image-restoration workflow adapting to the image's degradations.

\paragraph{Contributions.}
Our main contributions are:
\begin{enumerate}
    \item \textbf{An agentic framework for joint restoration and compression.}
    The framework is agnostic to the specific restoration and compression methods, so new tools can be integrated without retraining.

    \item \textbf{Rate-penalized codec and quality selection.}
    We propose a joint codec and quality search using local
    distortion, no-reference quality, and a rate penalty.

    \item \textbf{High quality compression on degraded images.}
    On MiO100, ARC improves reconstruction quality over standalone codecs and fixed-order pipelines.

\end{enumerate}

\section{Related Work}

\subsection{Image Compression}
Traditional codecs such as JPEG~\cite{wallaceJPEGStillPicture1992}, the HEVC-intra-based BPG~\cite{bellardBPGImageFormat2015}, and the more recent VVC~\cite{brossOverviewVersatileVideo2021} remain widely used for image compression, relying on hand-crafted transform coding and entropy coding. Learned image compression (LIC) methods have since surpassed these approaches in rate-distortion performance. Most LIC models follow a nonlinear transform coding paradigm: an autoencoder maps the input image to a latent representation that is quantized and entropy-coded under a hyper-prior~\cite{balleVariationalImageCompression2018}, which estimates the distribution of the latent to drive arithmetic coding. Most subsequent methods build an autoregressive context model on top of this hyper-prior, using already-decoded parts of the latent to further refine the entropy estimate~\cite{minnenJointAutoregressiveHierarchical2018,heELICEfficientLearned2022}. Recent architectural work replaces convolutional backbones with more efficient sequence models; LALIC~\cite{fengLinearAttentionModeling2025}, for instance, uses Bi-RWKV blocks to reduce the computational cost of context modeling. One of the most capable recent works is DCAE~\cite{luLearnedImageCompression2025}, which introduces a dictionary-based entropy model.

Since most LIC models are optimized for pixel-wise distortion metrics such as MSE, their reconstructions can appear blurry at low bitrates despite high PSNR. A separate line of work instead targets perceptual quality by replacing the decoder with a generative model. Agustsson et al.~\cite{agustssonGenerativeAdversarialNetworks2019} were the first to use a GAN-based decoder for extreme image compression, and HiFiC~\cite{mentzerHighFidelityGenerativeImage2020} improved on this approach to produce high-fidelity reconstructions at low bitrates. More recently, diffusion-based decoders have replaced GANs to further improve perceptual quality. CoD~\cite{jiaCoDDiffusionFoundation2026} instead trains a diffusion foundation model from scratch specifically for compression, rather than adapting a large pretrained text-to-image model, achieving state-of-the-art perceptual quality at ultra-low bitrates while being substantially cheaper to train and faster to run than prior diffusion-based codecs.
In our framework, we focus on DCAE~\cite{luLearnedImageCompression2025} and CoD-Lite~\cite{jiaCoDLiteRealTimeDiffusionBased2026} as near state-of-the-art fidelity- and perception-optimized compression backbones, respectively, alongside VVC~\cite{brossOverviewVersatileVideo2021} as a hand-crafted baseline. However, the compression step is not tied to any of these specific codecs, and any of the above compression methods could be used as the compression action within our framework. Separately, Comp-X~\cite{gaoCompXDefiningInteractive2025} uses an LLM agent to select coding modes from user requests, but does not incorporate restoration or account for input degradations.

\subsection{Image Restoration}
A single degraded image often exhibits multiple, unknown degradations simultaneously (e.g., noise, blur, compression artifacts), motivating ``all-in-one'' restoration models that handle diverse degradations within a single network. AirNet~\cite{liAllInOneImageRestoration2022}, for instance, uses a contrastive degradation encoder to condition a shared restoration network without requiring prior knowledge of the corruption type. While these methods generalize across degradation types, they apply a single fixed forward pass and do not reason about the order or necessity of individual operations, nor account for a downstream compression objective.

Agentic restoration frameworks instead depart from this single fixed-pass paradigm by using LLMs or VLMs to plan and execute a sequence of restoration operations. AgenticIR~\cite{zhuIntelligentAgenticSystem2025} follows a five-stage perception--scheduling--execution--reflection--rescheduling loop, using a fine-tuned VLM for quality assessment and an LLM to select and reorder tools; 4KAgent~\cite{zuo4KAgentAgenticAny2026} extends this execution-reflection paradigm to any-resolution restoration and 4K super-resolution with a larger toolbox, rolling back steps that fail to improve quality. RAR~\cite{chenRestoreAssessRepeat2026} instead unifies degradation identification, restoration, and quality assessment into a single, end-to-end trainable latent-domain model, avoiding the latency of separate perception and execution modules, while Restore-R1~\cite{luRestoreR1EfficientImage2026} removes the iterative trial-and-error process altogether using a reinforcement-learning-based agent. HybridAgent~\cite{liHybridAgentsImage2026} instead restores according to a user prompt, adding a quick path for specific prompts and tools that handle multiple degradations at once. All these frameworks assume every applied step should improve measured image quality, and none incorporate a lossy compression action, whose purpose is to trade quality for bitrate rather than remove degradations.

\subsection{Joint Compression and Restoration}
Rather than treating compression and restoration as separate pipelines, Zeng et al.~\cite{zengAllinOneImageCompression2025} train a unified all-in-one framework that incorporates restoration directly into a learned codec, aggregating content and degradation representations to compress and restore degraded inputs within a single encoder-decoder pass. Such joint models are trained end-to-end for a fixed set of degradations, so extending them requires retraining.

In contrast, our framework composes existing, independently trained compression and restoration modules through agentic planning, allowing new tools to be integrated and rate-distortion trade-offs to be adjusted.

\input{framework_figure}
\section{Proposed Method}
\label{sec:methodology}

We propose an agentic framework for combined image restoration and compression (ARC): rather than applying a fixed compression pipeline, an agent plans and executes a sequence of restoration and compression actions tailored to the input image's degradations and a user-specified rate-distortion trade-off.
This allows us to exploit the synergy between restoration and compression decisions, deciding which degradations to remove before compression and which to leave in the compressed representation and restore afterwards.
The framework follows a perception-execution-reflection loop detailed in \cref{fig:framework} and consists of the following steps:

\begin{itemize}
    \item \textbf{Perception.} DepictQA~\cite{youDepictingScoresAdvancing2025} analyzes the input image and produces a list of detected degradations.
    \item \textbf{Scheduling.} The detected degradations are mapped to a fixed priority list of restoration steps and a compression step. (\cref{sec:agenda_schedule})
    \item \textbf{Execution.} The agent executes the restoration and compression steps in the order determined by the scheduler, applying each action to the current image. (\cref{sec:bitrate_control})
    \item \textbf{Reflection.} After each action, the agent evaluates the result using no-reference IQA metrics and a preference model. If the action results in a significant quality drop, it is rolled back and the plan is updated. (\cref{sec:rollback})
    \item \textbf{Transmission.} The transmitted representation contains the codec bitstream together with the decoder-side restoration plan. (\cref{sec:compressed_representation})
\end{itemize}

\input{mio100_rd_lpips_musiq}
\subsection{Selecting restoration and compression order}
\label{sec:agenda_schedule}

The order of restoration and compression operations matters because each operation changes the input seen by subsequent tools. Prior agentic restoration methods use LLMs to schedule restoration operations, guided by detected degradations and restoration knowledge supplied through prompts~\cite{zhuIntelligentAgenticSystem2025,zuo4KAgentAgenticAny2026}. Such knowledge can encode ordering preferences that can also be expressed directly as rules. We therefore use a deterministic priority list extracted from distilled restoration experience and extended to include compression. Given the degradations detected by the perception agent, the scheduler retains the corresponding restoration tasks, includes the compression step, and sorts all selected actions by priority. The resulting plan therefore adapts to the detected degradations while using a fixed relative order for restoration and compression. This makes the initial ordering reproducible; individual tools are still selected and evaluated within the execution--reflection loop. The complete toolbox and the priority list are provided in \cref{sec:toolbox}.

\subsection{Tool Selection and Rate Control}
\label{sec:bitrate_control}

For each restoration task, we evaluate the eligible tools on the current image and rank their outputs using the no-reference image quality assessment (IQA) metrics CLIPIQA~\cite{wangExploringCLIPAssessing2023}, MUSIQ~\cite{keMUSIQMultiscaleImage2021}, NIQE~\cite{mittalMakingCompletelyBlind2013} and the preference model HPSv2~\cite{wuHumanPreferenceScore2023}. We select the candidate with the best mean rank. Any tool output ranked last by any metric is excluded, unless every candidate would be excluded, to prevent the selection of a tool that introduces severe artifacts. Mean-rank ties are resolved by HPSv2. This avoids combining differently scaled metrics through manually chosen weights.

For the compression step we need a different selection strategy, since we need to account for the trade-off between quality and bitrate. We therefore rank all candidate codecs and quality levels jointly by local distortion, a no-reference quality $Q$ ensemble, and a rate term.

$Q$ combines four no-reference IQA metrics in a weighted sum: MANIQA~\cite{yangMANIQAMultidimensionAttention2022}, CLIPIQA~\cite{wangExploringCLIPAssessing2023}, MUSIQ~\cite{keMUSIQMultiscaleImage2021}, and NIQE~\cite{mittalMakingCompletelyBlind2013}.
For any given image $X$ and tool $T$, we calculate the quality score $Q(T(X))$ as the weighted sum over the $N=4$ no-reference IQA metrics $m_1, \dots, m_N$, with weights $w_1, \dots, w_N$:
\begin{equation}
    Q(T(X)) = \frac{1}{N}\sum_{n=1}^{N} w_n\, m_n(T(X))
\end{equation}

The rate a given setting achieves also depends on the restoration steps applied beforehand, so it is unknown until the codec has run on the current image. We therefore encode the current compression input $X$ with candidate codec-qualities $k \in \mathcal{K}$ and rank the reconstructions $\hat{X}_k$ jointly across codecs and quality levels. We select
\begin{equation}
    k^{\star} = \arg\min_{k \in \mathcal{K}} \; D(X, \hat{X}_k) + \lambda r_k + \beta \left(1 - Q(\hat{X}_k)\right), \label{eq:selection_objective}
\end{equation}
with the local distortion
\begin{equation}
    D(X, \hat{X}_k) = \mathrm{MSE}(X, \hat{X}_k) + \alpha \min\!\left(\frac{\mathrm{LPIPS}(X, \hat{X}_k)}{s}, 1\right).
\end{equation}
Both distortion terms reference the compression input $X$ rather than the clean image, which is unavailable at inference, so $D$ measures only the loss introduced by the codec. MSE is computed on RGB values in $[0, 1]$. The rate $r_k = b_k / P_{\text{out}}$ divides the candidate's coded bits $b_k$ by the pixel count of the final output, so that $\lambda$ expresses the same trade-off at any position in the plan, including before upscaling. We use $\alpha = 0.006$, $s = 0.3$, and $\beta = 0.01$.

The Lagrange multiplier $\lambda$ is the only rate control: larger values favor lower rates, and varying $\lambda$ yields different operating points. The selected reconstruction $\hat{X}_{k^{\star}}$ is then used in the subsequent steps.

\subsubsection{Speeding up the search}
Running all candidates is expensive, especially when a codec uses a diffusion-based decoder.
We therefore use a budgeted search that evaluates seven initial candidates and adaptively allocates up to five additional evaluations.
The initial probes are DCAE qualities 2 and 6, CoD-Lite qualities 2, 4, and 6, and VVC QPs 24 and 63.
Each probe encodes and decodes the current image at its full resolution and measures the rate, local distortion, and no-reference quality.

Using the observed candidates for each codec, we estimate the remaining settings by piecewise-linear interpolation or extrapolation of log bitrate, log MSE, log LPIPS, and the no-reference score along the quality ladder.
At each iteration, we evaluate the setting with the largest estimated improvement over the best measured candidate, considering all codecs jointly, and update the estimates.
We stop when no estimate improves on the current best objective or when 12 candidates have been evaluated, and return the best measured reconstruction.

This procedure reduces the number of codec evaluations while retaining similar mean final quality in our comparison with exhaustive search.

\subsection{Reflection and Rollback}
\label{sec:rollback}

After each restoration or compression step $T(X)$, the agent evaluates the result using the no-reference IQA metrics $Q$ and preference model introduced in \cref{sec:bitrate_control}. If the action leads to a quality degradation beyond the acceptable threshold, it is rolled back and the plan is updated by switching it with the next tool step. This makes reflection a guard against large score losses while allowing useful operations, including denoising and compression, to reduce texture-sensitive quality scores.
The rollback score $R$ is a combination of the no-reference IQA metrics and the preference model $H$ using the image caption $C_X$ (produced by a VLM) as context and is calculated on the previous image $X$ and the current image $T(X)$:

\begin{equation}
    R(T(X)) = H(T(X), C_X) + Q(T(X))
\end{equation}

We find that not all restoration steps improve quality, therefore we only roll back if the tool resulted in a large quality drop. Since compression inherently reduces quality, we allow a bounded quality drop for compression steps. In our experiments we use the quality-drop thresholds $\tau_c=0.2$ for compression and $\tau_r=0.1$ for restoration steps. The rollback decision is therefore made according to the following rule:

\begin{equation}
    \begin{cases}
        \text{rollback}, & T \in \mathcal{C} \text{ and } R(X) - R(T(X)) > \tau_c \\
        \text{rollback}, & T \not\in \mathcal{C} \text{ and } R(X) - R(T(X)) > \tau_r \\
        \text{continue}, & \text{otherwise}
    \end{cases}
\end{equation}
where $\mathcal{C}$ denotes the set of compression tools.

\subsection{Compressed Representation}\label{sec:compressed_representation}
Since our framework mixes restoration and compression steps, several restoration steps might be needed after decompression to achieve the best possible quality. To avoid re-running the costly agentic process on every decompression, we store the compressed representation of the image along with the remaining restoration plan.
The decoder executes the stored restoration plan after codec decoding, bypassing the agent's search. We evaluate the performance overhead in \cref{sec:overhead}. The full format is described in \cref{sec:supp_container}.

\begin{table*}[t]
    \centering
    \footnotesize
    \setlength{\tabcolsep}{2pt}
    \begin{tabular}{ll*{4}{>{\columncolor{frbg}}c}*{5}{>{\columncolor{nrbg}}c}cc}
        \toprule
        G & Method & PSNR $\uparrow$ & MS-SSIM $\uparrow$ & LPIPS $\downarrow$ & PieAPP $\downarrow$ & MANIQA $\uparrow$ & CLIPIQA $\uparrow$ & MUSIQ $\uparrow$ & TOPIQ $\uparrow$ & DBCNN $\uparrow$ & BPP \textdownarrow & Avg. rank $\downarrow$ \\
        \midrule
        \multirow{10}{*}{A}
         & \textcolor{gray}{\textit{4KAgent}~\cite{zuo4KAgentAgenticAny2026}} & \textcolor{gray}{\textit{21.46}} & \textcolor{gray}{\textit{0.843}} & \textcolor{gray}{\textit{0.302}} & \textcolor{gray}{\textit{1.782}} & \textcolor{gray}{\textit{0.562}} & \textcolor{gray}{\textit{0.559}} & \textcolor{gray}{\textit{63.48}} & \textcolor{gray}{\textit{0.528}} & \textcolor{gray}{\textit{0.554}} & \textcolor{gray}{\textit{N/A}} & \textcolor{gray}{\textit{N/A}} \\
         & \textcolor{gray}{\textit{\ARCIR{}}} & \textcolor{gray}{\textit{21.77}} & \textcolor{gray}{\textit{0.853}} & \textcolor{gray}{\textit{0.293}} & \textcolor{gray}{\textit{1.708}} & \textcolor{gray}{\textit{0.553}} & \textcolor{gray}{\textit{0.552}} & \textcolor{gray}{\textit{62.90}} & \textcolor{gray}{\textit{0.511}} & \textcolor{gray}{\textit{0.541}} & \textcolor{gray}{\textit{N/A}} & \textcolor{gray}{\textit{N/A}} \\
         & VVC~\cite{brossOverviewVersatileVideo2021} & 19.07\textsuperscript{7} & 0.741\textsuperscript{7} & 0.555\textsuperscript{8} & 3.847\textsuperscript{6} & 0.394\textsuperscript{7} & 0.287\textsuperscript{8} & 38.47\textsuperscript{8} & 0.319\textsuperscript{7} & 0.371\textsuperscript{8} & 0.150 & 7.33 \\
         & DCAE~\cite{luLearnedImageCompression2025} & 20.40\textsuperscript{5} & 0.805\textsuperscript{5} & 0.504\textsuperscript{6} & 3.891\textsuperscript{7} & 0.393\textsuperscript{8} & 0.402\textsuperscript{6} & 43.18\textsuperscript{6} & 0.301\textsuperscript{8} & 0.403\textsuperscript{7} & 0.150 & 6.44 \\
         & CoD-Lite~\cite{jiaCoDLiteRealTimeDiffusionBased2026} & 18.46\textsuperscript{8} & 0.711\textsuperscript{8} & 0.540\textsuperscript{7} & 3.912\textsuperscript{8} & 0.395\textsuperscript{6} & 0.379\textsuperscript{7} & 38.93\textsuperscript{7} & 0.367\textsuperscript{6} & 0.405\textsuperscript{6} & 0.150 & 7.00 \\
         & \ARCIR{} + DCAE & \textbf{21.75}\textsuperscript{1} & \textbf{0.851}\textsuperscript{1} & \textit{0.368}\textsuperscript{3} & \textit{1.943}\textsuperscript{3} & 0.505\textsuperscript{5} & 0.449\textsuperscript{5} & \underline{61.56}\textsuperscript{2} & 0.462\textsuperscript{4.5} & 0.486\textsuperscript{5} & 0.150 & \textit{3.28} \\
         & \ARCIR{} + CoD-Lite & 21.04\textsuperscript{4} & \textit{0.829}\textsuperscript{3} & \textbf{0.307}\textsuperscript{1} & \underline{1.921}\textsuperscript{2} & \textbf{0.543}\textsuperscript{1} & \textbf{0.561}\textsuperscript{1} & \textit{61.05}\textsuperscript{3} & \underline{0.477}\textsuperscript{2} & 0.491\textsuperscript{4} & 0.150 & \underline{2.33} \\
         & DCAE + \ARCIR{} & \textit{21.15}\textsuperscript{3} & 0.819\textsuperscript{4} & 0.389\textsuperscript{5} & 2.085\textsuperscript{4} & 0.508\textsuperscript{4} & \textit{0.506}\textsuperscript{3} & 60.38\textsuperscript{4} & \textit{0.474}\textsuperscript{3} & \underline{0.509}\textsuperscript{2} & 0.150 & 3.56 \\
         & CoD-Lite + \ARCIR{} & 19.93\textsuperscript{6} & 0.769\textsuperscript{6} & 0.378\textsuperscript{4} & 2.258\textsuperscript{5} & \textit{0.512}\textsuperscript{3} & 0.505\textsuperscript{4} & 58.81\textsuperscript{5} & 0.462\textsuperscript{4.5} & \textit{0.493}\textsuperscript{3} & 0.150 & 4.50 \\
         & ARC & \underline{21.19}\textsuperscript{2} & \underline{0.830}\textsuperscript{2} & \underline{0.327}\textsuperscript{2} & \textbf{1.855}\textsuperscript{1} & \underline{0.530}\textsuperscript{2} & \underline{0.544}\textsuperscript{2} & \textbf{61.71}\textsuperscript{1} & \textbf{0.490}\textsuperscript{1} & \textbf{0.519}\textsuperscript{1} & 0.150 & \textbf{1.56} \\
        \midrule
        \multirow{10}{*}{B}
         & \textcolor{gray}{\textit{4KAgent}~\cite{zuo4KAgentAgenticAny2026}} & \textcolor{gray}{\textit{20.90}} & \textcolor{gray}{\textit{0.844}} & \textcolor{gray}{\textit{0.300}} & \textcolor{gray}{\textit{1.805}} & \textcolor{gray}{\textit{0.570}} & \textcolor{gray}{\textit{0.557}} & \textcolor{gray}{\textit{62.76}} & \textcolor{gray}{\textit{0.528}} & \textcolor{gray}{\textit{0.554}} & \textcolor{gray}{\textit{N/A}} & \textcolor{gray}{\textit{N/A}} \\
         & \textcolor{gray}{\textit{\ARCIR{}}} & \textcolor{gray}{\textit{21.51}} & \textcolor{gray}{\textit{0.861}} & \textcolor{gray}{\textit{0.287}} & \textcolor{gray}{\textit{1.698}} & \textcolor{gray}{\textit{0.564}} & \textcolor{gray}{\textit{0.552}} & \textcolor{gray}{\textit{62.65}} & \textcolor{gray}{\textit{0.520}} & \textcolor{gray}{\textit{0.547}} & \textcolor{gray}{\textit{N/A}} & \textcolor{gray}{\textit{N/A}} \\
         & VVC~\cite{brossOverviewVersatileVideo2021} & 18.86\textsuperscript{7} & 0.725\textsuperscript{7} & 0.550\textsuperscript{7} & 3.425\textsuperscript{6} & 0.424\textsuperscript{6} & 0.281\textsuperscript{8} & 42.32\textsuperscript{7} & 0.345\textsuperscript{7} & 0.390\textsuperscript{8} & 0.150 & 7.00 \\
         & DCAE~\cite{luLearnedImageCompression2025} & 20.32\textsuperscript{5} & 0.793\textsuperscript{5} & 0.497\textsuperscript{6} & 3.631\textsuperscript{8} & 0.421\textsuperscript{7} & 0.402\textsuperscript{6} & 47.05\textsuperscript{6} & 0.331\textsuperscript{8} & 0.427\textsuperscript{6} & 0.150 & 6.33 \\
         & CoD-Lite~\cite{jiaCoDLiteRealTimeDiffusionBased2026} & 18.45\textsuperscript{8} & 0.692\textsuperscript{8} & 0.573\textsuperscript{8} & 3.550\textsuperscript{7} & 0.415\textsuperscript{8} & 0.385\textsuperscript{7} & 40.54\textsuperscript{8} & 0.377\textsuperscript{6} & 0.395\textsuperscript{7} & 0.150 & 7.44 \\
         & \ARCIR{} + DCAE & \textbf{21.49}\textsuperscript{1} & \textbf{0.858}\textsuperscript{1} & \textit{0.367}\textsuperscript{3} & 1.994\textsuperscript{4} & 0.509\textsuperscript{5} & 0.453\textsuperscript{5} & \textit{61.30}\textsuperscript{3} & 0.468\textsuperscript{5} & 0.490\textsuperscript{5} & 0.150 & 3.56 \\
         & \ARCIR{} + CoD-Lite & \textit{20.88}\textsuperscript{3} & \textit{0.836}\textsuperscript{3} & \textbf{0.301}\textsuperscript{1} & \underline{1.922}\textsuperscript{2} & \textbf{0.550}\textsuperscript{1} & \textbf{0.565}\textsuperscript{1} & 61.19\textsuperscript{4} & 0.484\textsuperscript{4} & 0.497\textsuperscript{4} & 0.150 & \underline{2.56} \\
         & DCAE + \ARCIR{} & 20.55\textsuperscript{4} & 0.811\textsuperscript{4} & 0.387\textsuperscript{5} & \textit{1.959}\textsuperscript{3} & \textit{0.531}\textsuperscript{3} & \textit{0.527}\textsuperscript{3} & \textbf{63.01}\textsuperscript{1} & \textbf{0.511}\textsuperscript{1} & \textbf{0.539}\textsuperscript{1} & 0.150 & \textit{2.78} \\
         & CoD-Lite + \ARCIR{} & 19.63\textsuperscript{6} & 0.769\textsuperscript{6} & 0.369\textsuperscript{4} & 2.123\textsuperscript{5} & 0.529\textsuperscript{4} & 0.512\textsuperscript{4} & 61.12\textsuperscript{5} & \textit{0.492}\textsuperscript{3} & \textit{0.525}\textsuperscript{3} & 0.150 & 4.44 \\
         & ARC & \underline{21.05}\textsuperscript{2} & \underline{0.841}\textsuperscript{2} & \underline{0.314}\textsuperscript{2} & \textbf{1.761}\textsuperscript{1} & \underline{0.547}\textsuperscript{2} & \underline{0.549}\textsuperscript{2} & \underline{62.47}\textsuperscript{2} & \underline{0.504}\textsuperscript{2} & \underline{0.536}\textsuperscript{2} & 0.150 & \textbf{1.89} \\
        \midrule
        \multirow{10}{*}{C}
         & \textcolor{gray}{\textit{4KAgent}~\cite{zuo4KAgentAgenticAny2026}} & \textcolor{gray}{\textit{19.92}} & \textcolor{gray}{\textit{0.779}} & \textcolor{gray}{\textit{0.422}} & \textcolor{gray}{\textit{2.666}} & \textcolor{gray}{\textit{0.495}} & \textcolor{gray}{\textit{0.531}} & \textcolor{gray}{\textit{55.71}} & \textcolor{gray}{\textit{0.495}} & \textcolor{gray}{\textit{0.528}} & \textcolor{gray}{\textit{N/A}} & \textcolor{gray}{\textit{N/A}} \\
         & \textcolor{gray}{\textit{\ARCIR{}}} & \textcolor{gray}{\textit{20.34}} & \textcolor{gray}{\textit{0.794}} & \textcolor{gray}{\textit{0.398}} & \textcolor{gray}{\textit{2.442}} & \textcolor{gray}{\textit{0.483}} & \textcolor{gray}{\textit{0.527}} & \textcolor{gray}{\textit{57.27}} & \textcolor{gray}{\textit{0.476}} & \textcolor{gray}{\textit{0.515}} & \textcolor{gray}{\textit{N/A}} & \textcolor{gray}{\textit{N/A}} \\
         & VVC~\cite{brossOverviewVersatileVideo2021} & 17.54\textsuperscript{7} & 0.629\textsuperscript{7} & 0.694\textsuperscript{7} & 4.617\textsuperscript{7} & 0.296\textsuperscript{7} & 0.297\textsuperscript{8} & 25.87\textsuperscript{7} & 0.307\textsuperscript{7} & 0.379\textsuperscript{6} & 0.150 & 7.00 \\
         & DCAE~\cite{luLearnedImageCompression2025} & 19.07\textsuperscript{6} & 0.746\textsuperscript{5} & 0.633\textsuperscript{6} & 4.975\textsuperscript{8} & 0.291\textsuperscript{8} & 0.376\textsuperscript{6} & 28.94\textsuperscript{6} & 0.223\textsuperscript{8} & 0.354\textsuperscript{7} & 0.150 & 6.67 \\
         & CoD-Lite~\cite{jiaCoDLiteRealTimeDiffusionBased2026} & 17.38\textsuperscript{8} & 0.607\textsuperscript{8} & 0.729\textsuperscript{8} & 4.606\textsuperscript{6} & 0.310\textsuperscript{6} & 0.324\textsuperscript{7} & 25.76\textsuperscript{8} & 0.336\textsuperscript{6} & 0.350\textsuperscript{8} & 0.150 & 7.22 \\
         & \ARCIR{} + DCAE & \textbf{20.47}\textsuperscript{1} & \textbf{0.797}\textsuperscript{1} & \textit{0.453}\textsuperscript{3} & 2.675\textsuperscript{4} & 0.440\textsuperscript{5} & 0.421\textsuperscript{5} & \textit{56.32}\textsuperscript{3} & 0.432\textsuperscript{5} & 0.477\textsuperscript{4} & 0.150 & 3.44 \\
         & \ARCIR{} + CoD-Lite & \underline{20.32}\textsuperscript{2} & \underline{0.787}\textsuperscript{2} & \textbf{0.399}\textsuperscript{1} & \textit{2.592}\textsuperscript{3} & \textit{0.478}\textsuperscript{3.5} & \textit{0.521}\textsuperscript{3} & 55.61\textsuperscript{5} & 0.440\textsuperscript{4} & 0.459\textsuperscript{5} & 0.150 & \textit{3.17} \\
         & DCAE + \ARCIR{} & 19.71\textsuperscript{4} & 0.753\textsuperscript{4} & 0.464\textsuperscript{5} & \underline{2.585}\textsuperscript{2} & \textbf{0.488}\textsuperscript{1} & \textbf{0.539}\textsuperscript{1} & \textbf{58.81}\textsuperscript{1} & \textbf{0.506}\textsuperscript{1} & \textbf{0.546}\textsuperscript{1} & 0.150 & \textbf{2.22} \\
         & CoD-Lite + \ARCIR{} & 19.49\textsuperscript{5} & 0.709\textsuperscript{6} & 0.456\textsuperscript{4} & 2.813\textsuperscript{5} & \underline{0.487}\textsuperscript{2} & 0.510\textsuperscript{4} & 56.20\textsuperscript{4} & \underline{0.481}\textsuperscript{2} & \textit{0.508}\textsuperscript{3} & 0.150 & 3.89 \\
         & ARC$^{\ddagger}$ & \textit{20.30}\textsuperscript{3} & \textit{0.786}\textsuperscript{3} & \underline{0.409}\textsuperscript{2} & \textbf{2.521}\textsuperscript{1} & \textit{0.478}\textsuperscript{3.5} & \underline{0.538}\textsuperscript{2} & \underline{57.37}\textsuperscript{2} & \textit{0.477}\textsuperscript{3} & \underline{0.514}\textsuperscript{2} & 0.126 & \underline{2.39} \\
        \bottomrule
    \end{tabular}
    \caption{\textbf{Results on MiO100.} Quality metrics are linearly interpolated in achieved BPP at 0.15 bpp. The gray italic 4KAgent and \ARCIR{} rows apply no compression. Fixed-order pipelines use \ARCIR{} without integrated compression. Avg. rank averages ranks across the nine displayed quality metrics. Superscripts give per-metric ranks, with average ranks for ties. ARC performs among the best across all groups and metrics. $^{\ddagger}$ On MiO100-C ARC uses $<0.150$ bpp even at the smallest tested $\lambda$ (highest rate), since compression is scheduled before super-resolution and higher bitrate did not improve the rate-quality tradeoff. }\label{tab:mio100_bpp015_interp}
\end{table*}

\begin{table*}[th!]
    \centering
    \footnotesize
    \setlength{\tabcolsep}{2.8pt}
    \begin{tabular}{l*{4}{>{\columncolor{frbg}}c}*{5}{>{\columncolor{nrbg}}c}cc}
        \toprule
        Method & PSNR $\uparrow$ & MS-SSIM $\uparrow$ & LPIPS $\downarrow$ & PieAPP $\downarrow$ & MANIQA $\uparrow$ & CLIPIQA $\uparrow$ & MUSIQ $\uparrow$ & TOPIQ $\uparrow$ & DBCNN $\uparrow$ & BPP \textdownarrow & Avg. rank $\downarrow$ \\
        \midrule
        VVC~\cite{brossOverviewVersatileVideo2021} & \underline{32.88}\textsuperscript{2} & \underline{0.962}\textsuperscript{2} & 0.232\textsuperscript{6} & 1.006\textsuperscript{5} & 0.665\textsuperscript{7} & 0.413\textsuperscript{8} & 71.69\textsuperscript{8} & 0.567\textsuperscript{8} & 0.591\textsuperscript{8} & 0.200 & 6.00 \\
        DCAE~\cite{luLearnedImageCompression2025} & \textbf{33.36}\textsuperscript{1} & \textbf{0.964}\textsuperscript{1} & 0.215\textsuperscript{5} & 1.014\textsuperscript{6} & \textit{0.680}\textsuperscript{4} & 0.567\textsuperscript{7} & 73.90\textsuperscript{5.5} & 0.635\textsuperscript{6} & 0.619\textsuperscript{6} & 0.200 & 4.61 \\
        CoD-Lite~\cite{jiaCoDLiteRealTimeDiffusionBased2026} & 27.33\textsuperscript{4} & 0.923\textsuperscript{4} & \textbf{0.087}\textsuperscript{1} & \textbf{0.915}\textsuperscript{1} & \underline{0.682}\textsuperscript{2.5} & \underline{0.756}\textsuperscript{2} & 73.69\textsuperscript{7} & 0.651\textsuperscript{4} & \textit{0.661}\textsuperscript{3.5} & 0.200 & \underline{3.22} \\
        \ARCIR{} + DCAE & \textit{29.91}\textsuperscript{3} & \textit{0.945}\textsuperscript{3} & 0.237\textsuperscript{7} & 1.115\textsuperscript{7} & 0.679\textsuperscript{5} & 0.570\textsuperscript{6} & \underline{74.18}\textsuperscript{2} & 0.633\textsuperscript{7} & 0.615\textsuperscript{7} & 0.200 & 5.22 \\
        \ARCIR{} + CoD-Lite & 25.34\textsuperscript{7} & 0.906\textsuperscript{7} & \underline{0.109}\textsuperscript{2} & 1.001\textsuperscript{4} & \underline{0.682}\textsuperscript{2.5} & \textbf{0.760}\textsuperscript{1} & 73.90\textsuperscript{5.5} & \textit{0.652}\textsuperscript{3} & \underline{0.664}\textsuperscript{2} & 0.200 & \textit{3.78} \\
        DCAE + \ARCIR{} & 26.06\textsuperscript{6} & 0.911\textsuperscript{6} & 0.255\textsuperscript{8} & 1.153\textsuperscript{8} & 0.662\textsuperscript{8} & 0.572\textsuperscript{5} & 73.97\textsuperscript{4} & 0.647\textsuperscript{5} & 0.647\textsuperscript{5} & 0.200 & 6.11 \\
        CoD-Lite + \ARCIR{} & 24.54\textsuperscript{8} & 0.892\textsuperscript{8} & 0.127\textsuperscript{4} & \textit{0.996}\textsuperscript{3} & 0.677\textsuperscript{6} & \textit{0.739}\textsuperscript{3} & \textit{74.02}\textsuperscript{3} & \textbf{0.660}\textsuperscript{1} & \textbf{0.675}\textsuperscript{1} & 0.200 & 4.11 \\
        \midrule
        ARC & 26.90\textsuperscript{5} & 0.914\textsuperscript{5} & \textit{0.121}\textsuperscript{3} & \underline{0.937}\textsuperscript{2} & \textbf{0.687}\textsuperscript{1} & 0.721\textsuperscript{4} & \textbf{74.43}\textsuperscript{1} & \underline{0.658}\textsuperscript{2} & \textit{0.661}\textsuperscript{3.5} & 0.200 & \textbf{2.94} \\
        \bottomrule
    \end{tabular}
    \caption{\textbf{Results on 24 clean Kodak images.} Quality metrics are linearly interpolated in achieved BPP at 0.20 bpp. Fixed-order pipelines use \ARCIR{} without integrated compression. Avg. rank averages ranks across the nine displayed quality metrics. Superscripts give per-metric ranks, with average ranks for ties. On clean images ARC generally performs worse than the standalone codecs on full-reference metrics, as it tries to correct for hallucinated degradations. }\label{tab:kodak}
\end{table*}

\section{Experiments}\label{sec:experiments}

\subsection{Experimental Setup}

We evaluate on two datasets. MiO100~\cite{zuo4KAgentAgenticAny2026}, our primary benchmark, applies synthetic mixed degradations across three groups (A/B/C) with (8/4/4) subsets of degradations, following prior agentic restoration work.
Kodak~\cite{franzenKodakLosslessTrue1999} (24 clean images, 512$\times$768) is a standard compression benchmark with no restoration required and serves as a baseline for evaluating the framework's behavior when there is nothing to restore.

We denote ARC with its compression action disabled by \ARCIR{} (image restoration only). Fixed pipelines \ARCIR{} + $X$ and $X$ + \ARCIR{} apply an external codec $X$ after or before this restoration-only variant, respectively; ARC instead schedules compression jointly with restoration.

We report on various full-reference metrics (PSNR, MS-SSIM, LPIPS~\cite{zhangUnreasonableEffectivenessDeep2018}, PieAPP~\cite{prashnaniPieAPPPerceptualImageError2018} in blue) and no-reference metrics (MANIQA~\cite{yangMANIQAMultidimensionAttention2022}, CLIPIQA~\cite{wangExploringCLIPAssessing2023}, MUSIQ~\cite{keMUSIQMultiscaleImage2021}, TOPIQ~\cite{chenTOPIQTopdownApproach2023}, DBCNN~\cite{zhangBlindImageQuality2020} in red).

\begin{table*}[t]
    \centering
    \footnotesize
    \setlength{\tabcolsep}{3pt}
    \begin{tabular}{l*{4}{>{\columncolor{frbg}}c}*{5}{>{\columncolor{nrbg}}c}c}
        \toprule
        Search & PSNR $\uparrow$ & MS-SSIM $\uparrow$ & LPIPS $\downarrow$ & PieAPP $\downarrow$ & MANIQA $\uparrow$ & CLIPIQA $\uparrow$ & MUSIQ $\uparrow$ & TOPIQ $\uparrow$ & DBCNN $\uparrow$ & BPP \\
        \midrule
        Capped (12) & 20.22 & 0.783 & \textbf{0.412} & \textbf{2.532} & \textbf{0.477} & \textbf{0.540} & 57.44 & \textbf{0.477} & \textbf{0.516} & 0.092 \\
        Full (19) & \textbf{20.23} & \textbf{0.784} & \textbf{0.412} & 2.540 & 0.476 & \textbf{0.540} & \textbf{57.45} & \textbf{0.477} & \textbf{0.516} & 0.092 \\
        \bottomrule
    \end{tabular}
    \caption{\textbf{Search ablation on MiO100-C.} The capped search evaluates at most 12 candidates; full search evaluates all 19. The capped search has minimal impact on final quality and usually selects the best candidate.}\label{tab:search}
\end{table*}

\begin{table*}[th!]
    \centering
    \footnotesize
    \setlength{\tabcolsep}{3pt}
    \begin{tabular}{l*{4}{>{\columncolor{frbg}}c}*{5}{>{\columncolor{nrbg}}c}cc}
        \toprule
        Toolbox & PSNR $\uparrow$ & MS-SSIM $\uparrow$ & LPIPS $\downarrow$ & PieAPP $\downarrow$ & MANIQA $\uparrow$ & CLIPIQA $\uparrow$ & MUSIQ $\uparrow$ & TOPIQ $\uparrow$ & DBCNN $\uparrow$ & BPP & Avg. rank $\downarrow$ \\
        \midrule
        CoD-Lite~\cite{jiaCoDLiteRealTimeDiffusionBased2026} only & 19.15\textsuperscript{4} & 0.693\textsuperscript{4} & 0.438\textsuperscript{4} & 2.682\textsuperscript{4} & \textbf{0.500}\textsuperscript{1} & \textbf{0.555}\textsuperscript{1} & \textbf{58.04}\textsuperscript{1} & \textbf{0.486}\textsuperscript{1} & 0.511\textsuperscript{4} & 0.044 & 2.67 \\
        DCAE~\cite{luLearnedImageCompression2025} only & \textbf{20.40}\textsuperscript{1} & \textbf{0.787}\textsuperscript{1} & \textit{0.427}\textsuperscript{3} & \textit{2.554}\textsuperscript{3} & 0.470\textsuperscript{4} & 0.528\textsuperscript{4} & \underline{57.48}\textsuperscript{2} & 0.475\textsuperscript{4} & \textbf{0.523}\textsuperscript{1} & 0.104 & \textit{2.56} \\
        ARC$_{ + HiFiC}$ & \underline{20.25}\textsuperscript{2} & \textit{0.781}\textsuperscript{3} & \underline{0.413}\textsuperscript{2} & \underline{2.553}\textsuperscript{2} & \underline{0.481}\textsuperscript{2} & \underline{0.544}\textsuperscript{2} & 57.33\textsuperscript{4} & \underline{0.480}\textsuperscript{2} & \underline{0.517}\textsuperscript{2} & 0.092 & \textbf{2.33} \\
        ARC (default) & \textit{20.22}\textsuperscript{3} & \underline{0.783}\textsuperscript{2} & \textbf{0.412}\textsuperscript{1} & \textbf{2.532}\textsuperscript{1} & \textit{0.477}\textsuperscript{3} & \textit{0.540}\textsuperscript{3} & \textit{57.44}\textsuperscript{3} & \textit{0.477}\textsuperscript{3} & \textit{0.516}\textsuperscript{3} & 0.092 & \underline{2.44} \\
        \bottomrule
    \end{tabular}
    \caption{\textbf{Compression-toolbox ablation on MiO100-C.} ARC includes DCAE, VVC, and CoD-Lite; ARC$_{ + HiFiC}$ adds HiFiC~\cite{mentzerHighFidelityGenerativeImage2020} to the toolbox. All runs use $\lambda = 0.01$. Avg. rank averages ranks across the nine quality metrics. Superscripts give per-metric ranks, with average ranks for ties. Adding HiFiC to the toolbox improves performance across most metrics at the cost of increased complexity. Limiting to a single codec trades overall quality, but might improve w.r.t. certain metrics.}\label{tab:ablation_toolbox}
\end{table*}

\subsection{Results for multiple degradations on MiO100}

\begin{figure*}[t]
    \centering
    \begingroup
    \setlength{\tabcolsep}{2pt}
    \footnotesize
    \newcommand{\qualcrop}[1]{%
        \sbox0{\includegraphics{#1}}%
        \includegraphics[width=\linewidth,
            trim={0.1375\ht0} {0.5375\ht0}
                {\dimexpr\wd0-0.45\ht0\relax} {0.15\ht0},clip]{#1}}
    \begin{tabular}{@{}*{6}{c}@{}}
        Input & Ground truth & DCAE only & DCAE $\to$ \ARCIR{} & \ARCIR{} $\to$ DCAE & ARC (ours) \\
        \begin{minipage}[t]{0.158\textwidth}\centering
            \qualcrop{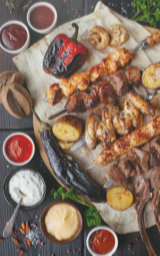}\par\smallskip
            Degraded source\par\strut\par\strut
        \end{minipage} &
        \begin{minipage}[t]{0.158\textwidth}\centering
            \qualcrop{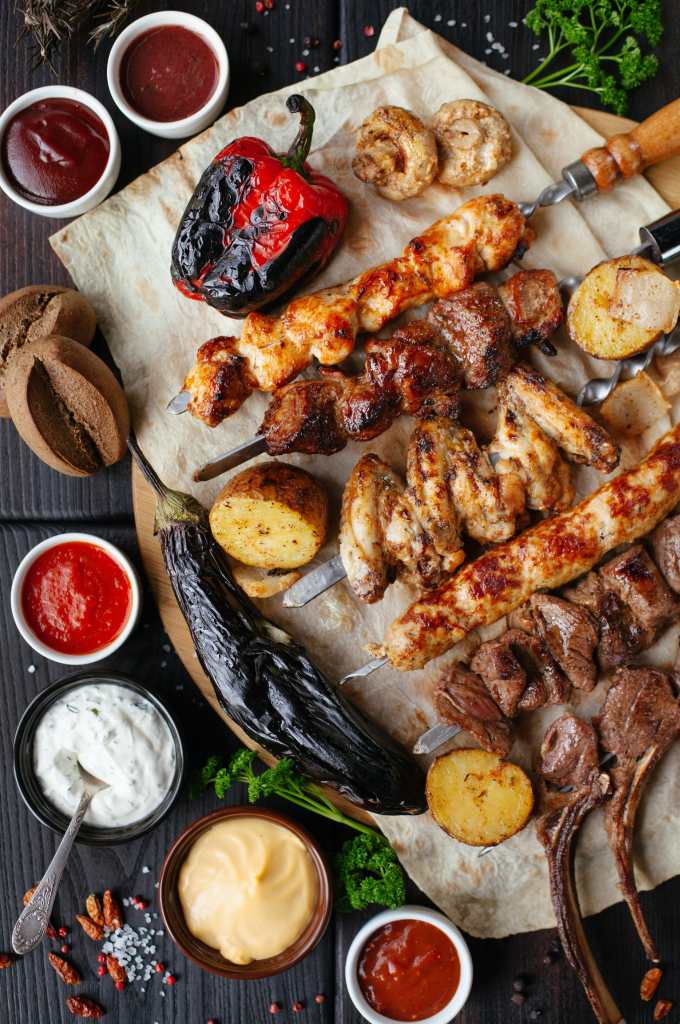}\par\smallskip
            Clean reference\par\strut\par\strut
        \end{minipage} &
        \begin{minipage}[t]{0.158\textwidth}\centering
            \qualcrop{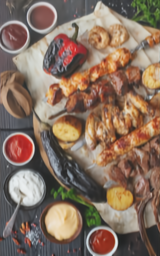}\par\smallskip
            0.085 bpp\par PSNR 18.59\,dB\par LPIPS 0.561
        \end{minipage} &
        \begin{minipage}[t]{0.158\textwidth}\centering
            \qualcrop{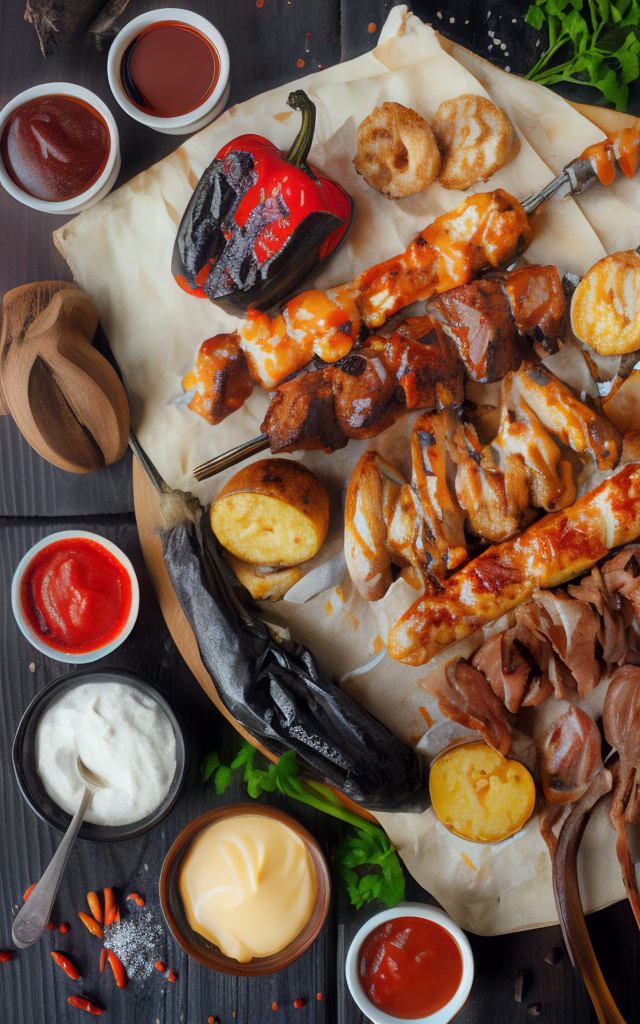}\par\smallskip
            0.085 bpp\par PSNR 19.35\,dB\par LPIPS 0.298
        \end{minipage} &
        \begin{minipage}[t]{0.158\textwidth}\centering
            \qualcrop{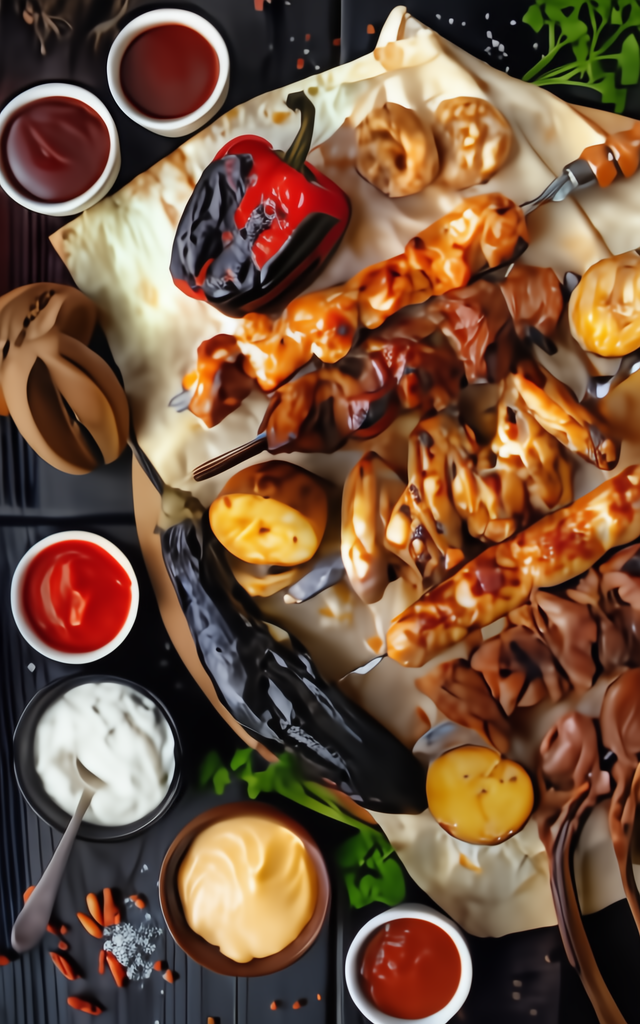}\par\smallskip
            0.132 bpp\par PSNR 20.92\,dB\par LPIPS 0.393
        \end{minipage} &
        \begin{minipage}[t]{0.158\textwidth}\centering
            \qualcrop{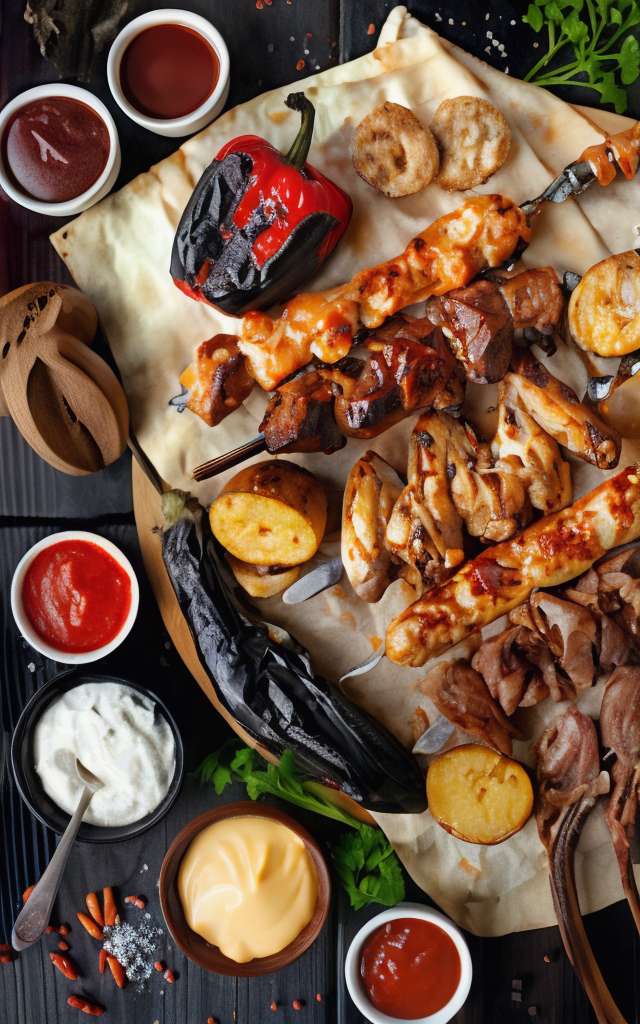}\par\smallskip
            0.095 bpp\par PSNR 20.74\,dB\par LPIPS 0.253
        \end{minipage}
    \end{tabular}
    \endgroup
    \caption{\textbf{Cropped qualitative comparison on MiO100 C2/068: haze + motion blur + low resolution.} Comparing the degraded input, clean reference, DCAE alone, both fixed orderings with restoration-only \ARCIR{}, and integrated ARC. Annotations report full-image bitrate, PSNR ($\uparrow$), and LPIPS ($\downarrow$). Compared to fixed pipelines using DCAE, our approach produces a sharper reconstruction. More exhaustive comparisons are available in \cref{sec:supp_qualitative}.}
    \label{fig:qualitative_c2_068_crop}
\end{figure*}

\Cref{fig:qualitative_c2_068_crop} compares cropped reconstructions for an example with haze, motion blur, and low resolution.

Standard codecs, trained on clean images, cannot account for degradations already present in the input, forcing users to bracket compression with a separate restoration step. \Cref{fig:rd_mio100_lpips_dbcnn} and \cref{tab:mio100_bpp015_interp} show neither ordering is safe by default: compression-then-restoration lets the codec spend bits encoding the degradations themselves, while restoration-then-compression can add detail that inflates the bitrate or gets smoothed away during compression. Depending on the degradations present in the input and the selected metric, some fixed pipelines can perform better than our agentic framework. However ARC consistently performs near the top for all metrics compared to the fixed pipelines.

Across all groups the codecs selected by the agent vary with the selected rate parameter $\lambda$. DCAE~\cite{luLearnedImageCompression2025} is selected most often (52\% - 56\%) across all rates, CoD-Lite~\cite{jiaCoDLiteRealTimeDiffusionBased2026} is picked more frequently at lower bitrates (up to 43\%), and VVC~\cite{brossOverviewVersatileVideo2021} is chosen at higher bitrates (up to 17\%).
On most images the agent schedules one (57\%) or two (26\%) restoration steps after compression.

\subsection{Results on clean images on Kodak}

Evaluation on the clean Kodak dataset in \cref{tab:kodak} shows that the perception agent can hallucinate degradations that are not present, leading to unnecessary restoration steps that reduce quality. The selected no-reference metrics used to trigger a rollback fail to flag the fidelity loss caused by unnecessary restoration, so the agent does not roll back these steps. This also applies to fixed pipelines that combine the \ARCIR{} restoration-only variant with a codec, which suffer from the same issue.

\section{Ablation Study}

We ablate several aspects of the agent's behavior on MiO100 Group C.

\subsection{Selection criteria}\label{sec:selector_weights}

We study our choice of selection mechanism  used in the compression selection objective of \cref{eq:selection_objective}, comparing the default combination of local distortion and no-reference IQA metrics against using only local distortion or only no-reference IQA metrics in \cref{tab:selector_weights}.
Even though the full-reference metrics are calculated using the possibly degraded compression input as reference, they are still informative about the quality of the compression step, since they measure how much additional distortion is introduced by the codec.
Dropping the local distortion term leads to worse performance especially for the full-reference metrics, while dropping the no-reference IQA metrics leads to worse performance for the no-reference metrics. The combination of both is therefore most balanced across full-reference metrics.

\begin{table*}[t]
    \centering
    \footnotesize
    \setlength{\tabcolsep}{3pt}
    \begin{tabular}{l*{4}{>{\columncolor{frbg}}c}*{5}{>{\columncolor{nrbg}}c}cc}
        \toprule
        Method & PSNR $\uparrow$ & MS-SSIM $\uparrow$ & LPIPS $\downarrow$ & PieAPP $\downarrow$ & MANIQA $\uparrow$ & CLIPIQA $\uparrow$ & MUSIQ $\uparrow$ & TOPIQ $\uparrow$ & DBCNN $\uparrow$ & BPP & Avg. rank $\downarrow$ \\
        \midrule
        All (ARC) & \textit{20.23}\textsuperscript{3} & \textit{0.783}\textsuperscript{3} & \textbf{0.413}\textsuperscript{1} & \underline{2.536}\textsuperscript{2} & 0.477\textsuperscript{4} & 0.540\textsuperscript{4} & 57.45\textsuperscript{4} & 0.478\textsuperscript{4} & 0.517\textsuperscript{4} & 0.090 & \underline{3.22} \\
        MSE + LPIPS & \underline{20.30}\textsuperscript{2} & \underline{0.787}\textsuperscript{2} & 0.418\textsuperscript{4} & \textit{2.539}\textsuperscript{3} & 0.471\textsuperscript{5} & 0.531\textsuperscript{5} & 57.14\textsuperscript{5} & 0.477\textsuperscript{5} & \underline{0.521}\textsuperscript{2} & 0.090 & \textit{3.67} \\
        IQA + MSE & 20.00\textsuperscript{4} & 0.765\textsuperscript{4} & \underline{0.416}\textsuperscript{2.5} & \textbf{2.533}\textsuperscript{1} & \underline{0.489}\textsuperscript{2} & \textit{0.547}\textsuperscript{3} & \underline{58.14}\textsuperscript{2} & \underline{0.483}\textsuperscript{2} & \textit{0.518}\textsuperscript{3} & 0.090 & \textbf{2.61} \\
        IQA + LPIPS & 19.99\textsuperscript{5} & 0.761\textsuperscript{5} & \underline{0.416}\textsuperscript{2.5} & 2.564\textsuperscript{5} & \textit{0.487}\textsuperscript{3} & \underline{0.549}\textsuperscript{2} & \textit{57.89}\textsuperscript{3} & \textit{0.482}\textsuperscript{3} & 0.515\textsuperscript{5} & 0.090 & 3.72 \\
        MSE only & \textbf{20.42}\textsuperscript{1} & \textbf{0.789}\textsuperscript{1} & 0.425\textsuperscript{5} & 2.548\textsuperscript{4} & 0.460\textsuperscript{6} & 0.521\textsuperscript{6} & 57.09\textsuperscript{6} & 0.468\textsuperscript{6} & \textbf{0.522}\textsuperscript{1} & 0.090 & 4.00 \\
        IQA only & 18.01\textsuperscript{6} & 0.584\textsuperscript{6} & 0.499\textsuperscript{6} & 3.047\textsuperscript{6} & \textbf{0.499}\textsuperscript{1} & \textbf{0.561}\textsuperscript{1} & \textbf{58.51}\textsuperscript{1} & \textbf{0.484}\textsuperscript{1} & 0.510\textsuperscript{6} & 0.090 & 3.78 \\
        \bottomrule
    \end{tabular}
    \caption{\textbf{Selection criteria ablation on MiO100 Group C} comparing different quality metrics for selecting the best compression codec. Quality metrics are linearly interpolated in achieved BPP at 0.09 bpp. Removing a distortion term worsens its corresponding metric, while removing no-reference IQA reduces most no-reference scores. The combination gives stable overall performance across the metrics, especially on the full-reference metrics. Superscripts indicate ranks across the six methods; Avg. rank averages the nine quality-metric ranks using the displayed values, with average ranks for ties and excluding BPP.}\label{tab:selector_weights}
\end{table*}

\subsection{Quality search}\label{sec:quality_search}

As described in \cref{sec:bitrate_control}, we limit the number of compression candidates evaluated to reduce runtime. \Cref{tab:search} compares the full search over all 19 candidates against our budgeted search that evaluates at most 12 candidates, showing that the reduced search space has minimal impact on final quality. The full candidate set comprises 6 DCAE checkpoints, 6 CoD-Lite checkpoints, and 7 VVC QPs.

\subsection{The compression toolkit}
Our main experiments restrict the toolbox to VVC, DCAE, and CoD-Lite, but any codec can be added. \Cref{tab:ablation_toolbox} compares restricting the toolbox to a single codec against the full toolbox, and against adding a fourth codec, HiFiC~\cite{mentzerHighFidelityGenerativeImage2020}. The additional tool only slightly improves the result in terms of PSNR and no-reference perceptual metrics.

\subsection{Performance overhead}\label{sec:overhead}
When it comes to time taken, adaptivity is not free: instead of a single fixed pass, the agent runs its execution-reflection loop, invoking VLM-based quality assessment and, potentially, several restoration tools around the compression action. \Cref{tab:overhead} counts the resulting tool calls: adding compression raises the average from 12 to roughly 23, with the codec-quality search accounting for about 11 of them. The actual time requirements depend significantly on the tools that are executed (e.g. compression tools usually are faster than restoration tools). However, this remains far fewer than an exhaustive search over restoration \& compression sequences would require.

During decoding only the codec and the remaining restoration steps are run, so the decoding overhead is smaller: the average number of restoration steps after compression is 1.2, and the codec is run only once.

On MiO100, the complete ARC container adds 0.64--1.45\% to the recorded codec bitrate across the three rate settings. This overhead includes the decoder-side restoration plan, container header, codec metadata, and native serialization fields absent from the recorded codec rate.

\begin{table}[t]
\centering\footnotesize
\setlength{\tabcolsep}{5pt}
\begin{tabular}{lrrr}
\toprule
Method & Tool calls & Codecs checked & Tools after compression \\
\midrule
\ARCIR{} & 12.07 & -- & -- \\
ARC & 23.25 & 11.12 & 1.19 \\
\bottomrule
\end{tabular}
\caption{Mean tool-call counts per image on MiO100.}\label{tab:overhead}
\end{table}

\section{Conclusion}

We introduced \textit{Agentic Restoration and Compression} (ARC), an agentic framework for image compression that plans and executes compression and restoration actions jointly, tailoring its strategy to the degradations present in each input image rather than applying a fixed compression pipeline. Because a compression action deliberately trades quality for bitrate rather than removing degradations, it violates the assumption -- built into prior agentic restoration systems -- that every action should improve quality, so we adapted the execution-reflection loop around it.

We evaluated the resulting framework against four classes of baselines: \ARCIR{} with no compression, a fixed compression scheme (e.g., VVC or DCAE) on its own, and both orderings of restoration and fixed compression. ARC consistently performs near the top across all metrics. The selected codec varies with the rate penalty, illustrating the benefit of supporting heterogeneous coding options. The framework can also be extended to include additional codecs or restoration tools without retraining.

\paragraph{Limitations.}
The controller uses no-reference metrics during action selection and reflection which might not be representative indicators of perceived image quality. Generative codecs and restoration tools may hallucinate details, and the planner can over-process clean inputs, as observed on Kodak. Finally, the framework requires the complete toolbox to be available both during encoding (agentic search) and decoding (post-compression restoration steps) which limits its applicability especially in resource-constrained environments.

\section*{LLM Usage Statement}
LLM-based tools Codex and Claude, as well as inline code completion (Copilot), were used to assist in writing and editing the manuscript, creating and formatting figures and for parts of the code implementation. All content and figures were verified by the authors and edited for accuracy and clarity by the authors. The proposed method itself also relies on a VLM (DepictQA~\cite{youDepictingScoresAdvancing2025}) for perception, but the LLM is not used for any other part of the method.

\bibliographystyle{ieeenat_fullname}
\bibliography{references}

\clearpage
\appendix
\maketitlesupplementary
\setcounter{figure}{0}
\setcounter{table}{0}
\renewcommand{\thefigure}{S\arabic{figure}}
\renewcommand{\thetable}{S\arabic{table}}
\renewcommand{\theHfigure}{S\arabic{figure}}
\renewcommand{\theHtable}{S\arabic{table}}

\section{Experimental Details}

Experiments were run on NVIDIA L40 GPUs. The perception is driven by DepictQA~\cite{youDepictingScoresAdvancing2025} as described in Section~\ref{sec:methodology}.

\section{Full Toolbox}\label{sec:toolbox}

Table~\ref{tab:full_toolbox} lists every tool available to the agent, expanding on Section~\ref{sec:bitrate_control}. We use the same restoration toolbox as 4KAgent~\cite{zuo4KAgentAgenticAny2026}. The agent's active compression toolbox is DCAE, VVC, CoD-Lite, and, only when explicitly enabled for the toolbox ablation (Table~\ref{tab:ablation_toolbox}), HiFiC.
We also include the priority of each restoration/compression task, with lower numbers scheduled earlier.

\begin{table*}[th]
    \centering
    \footnotesize
    \begin{tabularx}{\textwidth}{lcX}
        \toprule
        Task & Priority & Methods \\
        \midrule
        Brightening & 8 & CLAHE, Constant Shift, DiffPlugin~\cite{liuDiffPluginRevitalizingDetails2024}, FourierDiff~\cite{lvFourierPriorsGuidedDiffusion2024}, Gamma Correction, MAXIM~\cite{tuMAXIMMultiAxisMLP2022} \\
        Defocus Deblurring & 5 & ConvIR~\cite{cuiRevitalizingConvolutionalNetwork2024}, DiffPlugin~\cite{liuDiffPluginRevitalizingDetails2024}, DRBNet~\cite{ruanLearningDeblurUsing2022}, IFAN~\cite{leeIterativeFilterAdaptive2021}, LaKDNet~\cite{ruanRevisitingImageDeblurring2023}, Restormer~\cite{zamirRestormerEfficientTransformer2022} \\
        Motion Deblurring & 3 & EVSSM~\cite{kongEfficientVisualState2025}, LaKDNet~\cite{ruanRevisitingImageDeblurring2023}, MAXIM~\cite{tuMAXIMMultiAxisMLP2022}, MPRNet~\cite{zamirMultiStageProgressiveImage2021}, NAFNet~\cite{chenSimpleBaselinesImage2022}, Restormer~\cite{zamirRestormerEfficientTransformer2022}, X-Restormer~\cite{chenComparativeStudyImage2024} \\
        Dehazing & 6 & DehazeFormer~\cite{songVisionTransformersSingle2023}, DiffPlugin~\cite{liuDiffPluginRevitalizingDetails2024}, MAXIM~\cite{tuMAXIMMultiAxisMLP2022}, RIDCP~\cite{wuRIDCPRevitalizingReal2023}, X-Restormer~\cite{chenComparativeStudyImage2024} \\
        Deraining & 1 & DiffPlugin~\cite{liuDiffPluginRevitalizingDetails2024}, MAXIM~\cite{tuMAXIMMultiAxisMLP2022}, MPRNet~\cite{zamirMultiStageProgressiveImage2021}, Restormer~\cite{zamirRestormerEfficientTransformer2022}, X-Restormer~\cite{chenComparativeStudyImage2024} \\
        Denoising & 2 & MAXIM~\cite{tuMAXIMMultiAxisMLP2022}, MPRNet~\cite{zamirMultiStageProgressiveImage2021}, NAFNet~\cite{chenSimpleBaselinesImage2022}, Restormer~\cite{zamirRestormerEfficientTransformer2022}, X-Restormer~\cite{chenComparativeStudyImage2024}, SwinIR~\cite{liangSwinIRImageRestoration2021} ($\sigma$15, $\sigma$50) \\
        JPEG Artifact Removal & 0 & FBCNN~\cite{jiangFlexibleBlindJPEG2021} (q5, q90, qB), SwinIR~\cite{liangSwinIRImageRestoration2021} \\
        Super-Resolution & 7 & DiffBIR~\cite{linDiffBIRBlindImage2025}, DRCT~\cite{hsuDRCTSavingImage2024}, HAT-L~\cite{chenActivatingMorePixels2023}, HAT-GAN~\cite{chenActivatingMorePixels2023}, HMA~\cite{chuHMANetHybridMultiAxis2024}, OSEDiff~\cite{wuOneStepEffectiveDiffusion2024}, PiSA-SR~\cite{sunPixellevelSemanticlevelAdjustable2025}, SwinIR~\cite{liangSwinIRImageRestoration2021}, SwinIR (Real-ISR)~\cite{liangSwinIRImageRestoration2021}, SwinFIR~\cite{zhangSwinFIRRevisitingSwinIR2023}, X-Restormer~\cite{chenComparativeStudyImage2024} \\
        \midrule
        Compression (ARC) & 4 & DCAE~\cite{luLearnedImageCompression2025}, VVC~\cite{brossOverviewVersatileVideo2021}, CoD-Lite~\cite{jiaCoDLiteRealTimeDiffusionBased2026}, HiFiC~\cite{mentzerHighFidelityGenerativeImage2020}$^\dagger$ \\
        \bottomrule
    \end{tabularx}
    \caption{\textbf{Full restoration and compression toolbox} available to the agent, by task. Lower numbers are scheduled earlier in the default rule-based plan. $^\dagger$HiFiC is not included in the toolbox of our main results; it is only used in the toolbox ablation (Table~\ref{tab:ablation_toolbox}).}\label{tab:full_toolbox}
\end{table*}

\section{Compressed Representation}\label{sec:supp_container}

The ARC container described in Section~\ref{sec:compressed_representation} stores a fixed header, compact JSON for codec metadata and the restoration plan, and the native codec bitstream. The 31-byte header contains a 4-byte magic identifier, a 2-byte format version, a 1-byte codec identifier, two 4-byte image dimensions, two 4-byte JSON lengths, and an 8-byte bitstream length. The metadata records the checkpoint index for DCAE and CoD-Lite, and the quantization parameter and encoded dimensions for VVC. The plan is a JSON array of task and tool identifiers, in execution order. The empty array encodes a plan with zero restoration steps.

The decoder reads the codec and its operating point, decodes the bitstream, and applies each listed restoration tool to the preceding output. Learned codecs and restoration tools use compatible checkpoints installed at the decoder. The serialized plan eliminates the need to repeat perception, tool selection, and reflection during decoding. If super-resolution appears in the plan, the delivered image can have more pixels than the image at the compression step.

For the main ARC results, we compute the serialized container size from the recorded codec payload and selected restoration plan. The reported rate divides the total bit count by the number of pixels in the delivered image. On MiO100, the container adds 0.64--1.45\% relative to the mean recorded codec rate across the three $\lambda$ settings, computed as the ratio of mean overhead bpp to mean codec bpp for each setting.

\section{Post-Compression Restoration}
\begin{figure*}[th]
    \centering
    \footnotesize
    \begin{subfigure}[t]{0.32\textwidth}
        \centering
        \begin{tikzpicture}
            \begin{axis}[
                width=\textwidth,
                height=0.85\textwidth,
                xlabel={BPP},
                ylabel={PSNR (dB)},
                legend to name=postcompressionlegend,
                legend columns=2,
                legend style={font=\footnotesize, legend cell align=left},
                grid=major,
                grid style={dashed, gray!30},
                xmin=0.05, xmax=0.15,
                ymin=19, ymax=21,
                xtick={0.06,0.08,0.1,0.12,0.14},
                scaled ticks=false,
                xticklabel style={/pgf/number format/fixed, /pgf/number format/precision=2},
                yticklabel style={/pgf/number format/fixed},
            ]
            \addlegendimage{red, mark=pentagon*, thick, mark size=3pt, mark options={fill=red!50!white}}
            \addlegendentry{Full plan (ARC)}
            \addplot[gray, mark=*, thick, dashed] coordinates {(0.1263,19.55) (0.0942,19.51) (0.0667,19.42)};
            \addlegendentry{After compression}
            \addplot[red, mark=pentagon*, thick, mark size=3pt, mark options={fill=red!50!white}, forget plot] coordinates {(0.1263,20.30) (0.0942,20.23) (0.0667,20.09)};
            \end{axis}
        \end{tikzpicture}
        \caption{PSNR $\uparrow$}\label{fig:post_compression_ablation_psnr}
    \end{subfigure}
    \hfill
    \begin{subfigure}[t]{0.32\textwidth}
        \centering
        \begin{tikzpicture}
            \begin{axis}[
                width=\textwidth,
                height=0.85\textwidth,
                xlabel={BPP},
                ylabel={MS-SSIM},
                grid=major,
                grid style={dashed, gray!30},
                xmin=0.05, xmax=0.15,
                ymin=0.765, ymax=0.795,
                xtick={0.06,0.08,0.1,0.12,0.14},
                scaled ticks=false,
                xticklabel style={/pgf/number format/fixed, /pgf/number format/precision=2},
                yticklabel style={/pgf/number format/fixed},
            ]
            \addplot[gray, mark=*, thick, dashed] coordinates {(0.1263,0.7843) (0.0942,0.7817) (0.0667,0.7744)};
            \addplot[red, mark=pentagon*, thick, mark size=3pt, mark options={fill=red!50!white}] coordinates {(0.1263,0.7859) (0.0942,0.7834) (0.0667,0.7760)};
            \end{axis}
        \end{tikzpicture}
        \caption{MS-SSIM $\uparrow$}\label{fig:post_compression_ablation_msssim}
    \end{subfigure}
    \hfill
    \begin{subfigure}[t]{0.32\textwidth}
        \centering
        \begin{tikzpicture}
            \begin{axis}[
                width=\textwidth,
                height=0.85\textwidth,
                xlabel={BPP},
                ylabel={LPIPS},
                grid=major,
                grid style={dashed, gray!30},
                xmin=0.05, xmax=0.15,
                ymin=0.35, ymax=0.65,
                xtick={0.06,0.08,0.1,0.12,0.14},
                scaled ticks=false,
                xticklabel style={/pgf/number format/fixed, /pgf/number format/precision=2},
                yticklabel style={/pgf/number format/fixed},
            ]
            \addplot[gray, mark=*, thick, dashed] coordinates {(0.1263,0.5833) (0.0942,0.5836) (0.0667,0.5878)};
            \addplot[red, mark=pentagon*, thick, mark size=3pt, mark options={fill=red!50!white}] coordinates {(0.1263,0.4090) (0.0942,0.4119) (0.0667,0.4185)};
            \end{axis}
        \end{tikzpicture}
        \caption{LPIPS $\downarrow$}\label{fig:post_compression_ablation_lpips}
    \end{subfigure}

    \begin{subfigure}[t]{0.32\textwidth}
        \centering
        \begin{tikzpicture}
            \begin{axis}[
                width=\textwidth,
                height=0.85\textwidth,
                xlabel={BPP},
                ylabel={MANIQA},
                grid=major,
                grid style={dashed, gray!30},
                xmin=0.05, xmax=0.15,
                ymin=0.25, ymax=0.5,
                xtick={0.06,0.08,0.1,0.12,0.14},
                scaled ticks=false,
                xticklabel style={/pgf/number format/fixed, /pgf/number format/precision=2},
                yticklabel style={/pgf/number format/fixed},
            ]
            \addplot[gray, mark=*, thick, dashed] coordinates {(0.1263,0.3092) (0.0942,0.3042) (0.0667,0.2993)};
            \addplot[red, mark=pentagon*, thick, mark size=3pt, mark options={fill=red!50!white}] coordinates {(0.1263,0.4780) (0.0942,0.4765) (0.0667,0.4781)};
            \end{axis}
        \end{tikzpicture}
        \caption{MANIQA $\uparrow$}\label{fig:post_compression_ablation_maniqa}
    \end{subfigure}
    \hfill
    \begin{subfigure}[t]{0.32\textwidth}
        \centering
        \begin{tikzpicture}
            \begin{axis}[
                width=\textwidth,
                height=0.85\textwidth,
                xlabel={BPP},
                ylabel={CLIPIQA},
                grid=major,
                grid style={dashed, gray!30},
                xmin=0.05, xmax=0.15,
                ymin=0.3, ymax=0.6,
                xtick={0.06,0.08,0.1,0.12,0.14},
                scaled ticks=false,
                xticklabel style={/pgf/number format/fixed, /pgf/number format/precision=2},
                yticklabel style={/pgf/number format/fixed},
            ]
            \addplot[gray, mark=*, thick, dashed] coordinates {(0.1263,0.3699) (0.0942,0.3701) (0.0667,0.3688)};
            \addplot[red, mark=pentagon*, thick, mark size=3pt, mark options={fill=red!50!white}] coordinates {(0.1263,0.5384) (0.0942,0.5397) (0.0667,0.5412)};
            \end{axis}
        \end{tikzpicture}
        \caption{CLIPIQA $\uparrow$}\label{fig:post_compression_ablation_clipiqa}
    \end{subfigure}
    \hfill
    \begin{subfigure}[t]{0.32\textwidth}
        \centering
        \begin{tikzpicture}
            \begin{axis}[
                width=\textwidth,
                height=0.85\textwidth,
                xlabel={BPP},
                ylabel={MUSIQ},
                grid=major,
                grid style={dashed, gray!30},
                xmin=0.05, xmax=0.15,
                ymin=25, ymax=65,
                xtick={0.06,0.08,0.1,0.12,0.14},
                scaled ticks=false,
                xticklabel style={/pgf/number format/fixed, /pgf/number format/precision=2},
                yticklabel style={/pgf/number format/fixed},
            ]
            \addplot[gray, mark=*, thick, dashed] coordinates {(0.1263,30.45) (0.0942,30.43) (0.0667,30.35)};
            \addplot[red, mark=pentagon*, thick, mark size=3pt, mark options={fill=red!50!white}] coordinates {(0.1263,57.37) (0.0942,57.44) (0.0667,57.65)};
            \end{axis}
        \end{tikzpicture}
        \caption{MUSIQ $\uparrow$}\label{fig:post_compression_ablation_musiq}
    \end{subfigure}

    \centering
    \ref{postcompressionlegend}

    \caption{\textbf{Stopping ARC directly after compression vs.\ executing the full plan on MiO100-C.} Both stages transmit the same bitstream, so each pair shares its full-container rate. Codec outputs that are still below output resolution (compression before super-resolution) are bicubically upscaled by the integer factor before scoring.}\label{fig:post_compression_ablation}
\end{figure*}
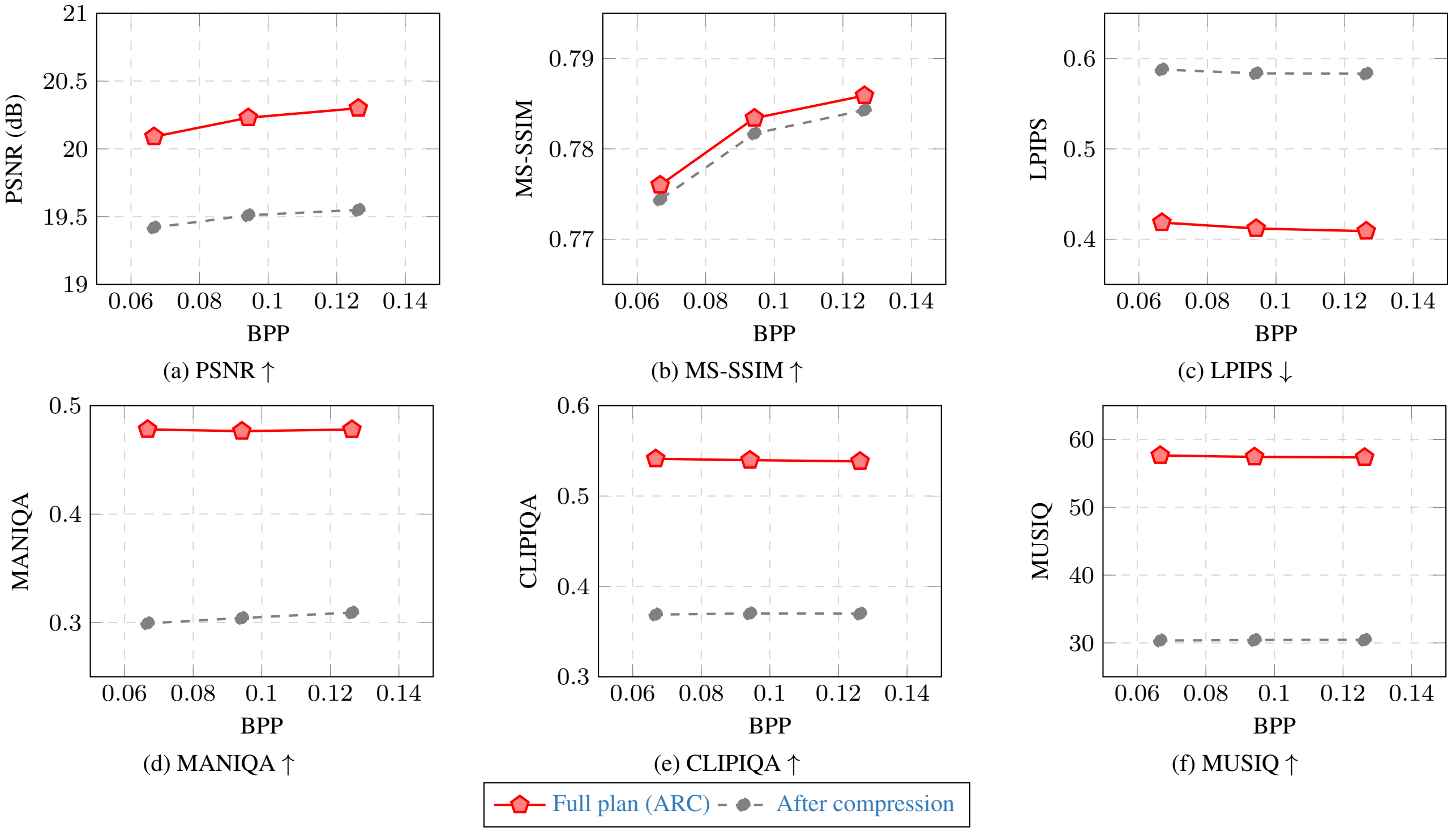

ARC usually schedules some restoration steps after compression. A decoder could skip these steps to decode faster, but the image would then remain partially restored. Figure~\ref{fig:overhead_quality} shows the reflection score on MiO100-C when stopping directly after compression and after executing the full plan. As expected, the full plan reaches a higher reflection score, since the agent scheduled these steps to improve the decoded image. Figure~\ref{fig:post_compression_ablation} shows that stopping after compression causes similar losses on the evaluation metrics.
We therefore recommend executing the full plan when decoding.

\begin{figure}[t!]
    \centering
    \footnotesize
    \begin{tikzpicture}
        \begin{axis}[
            width=\columnwidth,
            height=0.55\columnwidth,
            xlabel={BPP},
            ylabel={Reflection score},
            legend style={at={(0.98,0.5)}, anchor=east, font=\scriptsize},
            grid=major,
            grid style={dashed, gray!30},
            xmin=0.05, xmax=0.14,
            ymin=0.5, ymax=0.78,
        ]
        \addplot[gray, mark=*, thick, dashed] table[x=bpp,y=after_codec,col sep=comma] {data/completed_v3/post_compression_iqa.csv};
        \addlegendentry{After compression}
        \addplot[red, mark=pentagon*, thick, mark options={scale=1, fill=red!50!white}] table[x=bpp,y=full_plan,col sep=comma] {data/completed_v3/post_compression_iqa.csv};
        \addlegendentry{Full plan (ARC)}
        \end{axis}
    \end{tikzpicture}
    \caption{\textbf{Reflection score when stopping ARC directly after compression vs.\ executing the full plan on MiO100-C.}}\label{fig:overhead_quality}
\end{figure}
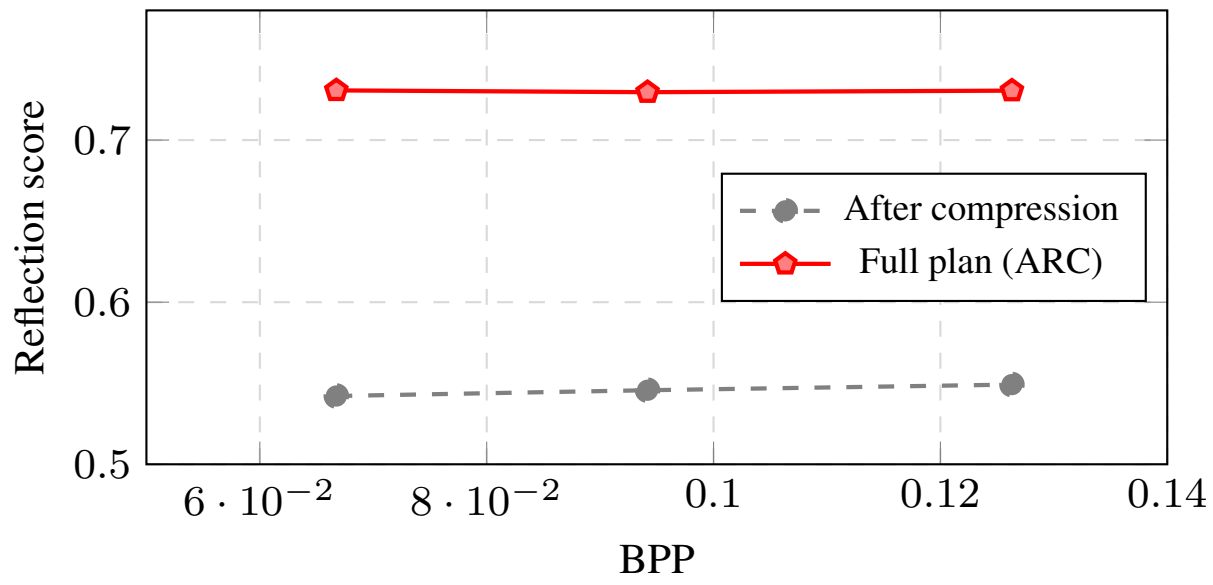

\section{Extended MiO100 Results}

\begin{table*}[th]
    \centering
    \footnotesize
    \setlength{\tabcolsep}{2pt}
    \begin{tabular}{ll*{4}{>{\columncolor{frbg}}c}*{5}{>{\columncolor{nrbg}}c}cc}
        \toprule
        G & Method & PSNR $\uparrow$ & MS-SSIM $\uparrow$ & LPIPS $\downarrow$ & PieAPP $\downarrow$ & MANIQA $\uparrow$ & CLIPIQA $\uparrow$ & MUSIQ $\uparrow$ & TOPIQ $\uparrow$ & DBCNN $\uparrow$ & BPP \textdownarrow & Avg. rank $\downarrow$ \\
        \midrule
        \multirow{13}{*}{A}
         & \textcolor{gray}{\textit{4KAgent}~\cite{zuo4KAgentAgenticAny2026}} & \textcolor{gray}{\textit{21.46}} & \textcolor{gray}{\textit{0.843}} & \textcolor{gray}{\textit{0.302}} & \textcolor{gray}{\textit{1.782}} & \textcolor{gray}{\textit{0.562}} & \textcolor{gray}{\textit{0.559}} & \textcolor{gray}{\textit{63.48}} & \textcolor{gray}{\textit{0.528}} & \textcolor{gray}{\textit{0.554}} & \textcolor{gray}{\textit{N/A}} & \textcolor{gray}{\textit{N/A}} \\
         & \textcolor{gray}{\textit{\ARCIR{}}} & \textcolor{gray}{\textit{21.77}} & \textcolor{gray}{\textit{0.853}} & \textcolor{gray}{\textit{0.293}} & \textcolor{gray}{\textit{1.708}} & \textcolor{gray}{\textit{0.553}} & \textcolor{gray}{\textit{0.552}} & \textcolor{gray}{\textit{62.90}} & \textcolor{gray}{\textit{0.511}} & \textcolor{gray}{\textit{0.541}} & \textcolor{gray}{\textit{N/A}} & \textcolor{gray}{\textit{N/A}} \\
         & VVC~\cite{brossOverviewVersatileVideo2021} & 19.07\textsuperscript{8} & 0.741\textsuperscript{10} & 0.556\textsuperscript{11} & 3.846\textsuperscript{8} & 0.393\textsuperscript{6.5} & 0.286\textsuperscript{11} & 38.43\textsuperscript{11} & 0.318\textsuperscript{7} & 0.369\textsuperscript{11} & 0.145 & 9.28 \\
         & DCAE~\cite{luLearnedImageCompression2025} & 20.40\textsuperscript{5} & 0.805\textsuperscript{5} & 0.504\textsuperscript{6} & 3.890\textsuperscript{9} & 0.393\textsuperscript{6.5} & 0.401\textsuperscript{6} & 43.18\textsuperscript{6} & 0.301\textsuperscript{8} & 0.403\textsuperscript{7} & 0.147 & 6.50 \\
         & CoD-Lite~\cite{jiaCoDLiteRealTimeDiffusionBased2026} & 18.36\textsuperscript{11} & 0.705\textsuperscript{11} & 0.541\textsuperscript{10} & 3.919\textsuperscript{11} & 0.392\textsuperscript{8} & 0.384\textsuperscript{7.5} & 38.86\textsuperscript{10} & 0.368\textsuperscript{6} & 0.404\textsuperscript{6} & 0.091 & 8.94 \\
         & AllInOne~\cite{zengAllinOneImageCompression2025} & 18.60\textsuperscript{10} & 0.780\textsuperscript{8} & 0.514\textsuperscript{9} & 3.915\textsuperscript{10} & 0.249\textsuperscript{10} & 0.384\textsuperscript{7.5} & 41.69\textsuperscript{8} & 0.295\textsuperscript{9} & 0.401\textsuperscript{8} & 0.336 & 8.83 \\
         & AirNet + DCAE & 19.20\textsuperscript{7} & 0.788\textsuperscript{6} & 0.507\textsuperscript{7} & 3.674\textsuperscript{6} & 0.256\textsuperscript{9} & 0.357\textsuperscript{10} & 42.48\textsuperscript{7} & 0.289\textsuperscript{10} & 0.389\textsuperscript{9} & 0.079 & 7.89 \\
         & DCAE + AirNet & 18.63\textsuperscript{9} & 0.782\textsuperscript{7} & 0.513\textsuperscript{8} & 3.709\textsuperscript{7} & 0.242\textsuperscript{11} & 0.359\textsuperscript{9} & 41.24\textsuperscript{9} & 0.281\textsuperscript{11} & 0.381\textsuperscript{10} & 0.081 & 9.00 \\
         & \ARCIR{} + DCAE & \textbf{21.75}\textsuperscript{1} & \textbf{0.851}\textsuperscript{1} & \textit{0.368}\textsuperscript{3} & \textit{1.945}\textsuperscript{3} & 0.504\textsuperscript{5} & 0.448\textsuperscript{5} & \textbf{61.54}\textsuperscript{1} & 0.462\textsuperscript{4} & 0.486\textsuperscript{5} & 0.148 & \textit{3.11} \\
         & \ARCIR{} + CoD-Lite & \textit{21.00}\textsuperscript{3} & \underline{0.827}\textsuperscript{2} & \textbf{0.308}\textsuperscript{1} & \underline{1.931}\textsuperscript{2} & \textbf{0.542}\textsuperscript{1} & \textbf{0.561}\textsuperscript{1} & \textit{60.95}\textsuperscript{3} & \underline{0.476}\textsuperscript{2} & 0.490\textsuperscript{4} & 0.125 & \underline{2.11} \\
         & DCAE + \ARCIR{} & \underline{21.15}\textsuperscript{2} & 0.818\textsuperscript{4} & 0.390\textsuperscript{5} & 2.085\textsuperscript{4} & 0.507\textsuperscript{4} & \textit{0.505}\textsuperscript{3} & 60.37\textsuperscript{4} & \textit{0.474}\textsuperscript{3} & \underline{0.509}\textsuperscript{2} & 0.147 & 3.44 \\
         & CoD-Lite + \ARCIR{} & 19.71\textsuperscript{6} & 0.757\textsuperscript{9} & 0.386\textsuperscript{4} & 2.314\textsuperscript{5} & \textit{0.508}\textsuperscript{3} & 0.501\textsuperscript{4} & 58.53\textsuperscript{5} & 0.459\textsuperscript{5} & \textit{0.491}\textsuperscript{3} & 0.096 & 4.89 \\
         & ARC & 20.93\textsuperscript{4} & \textit{0.819}\textsuperscript{3} & \underline{0.332}\textsuperscript{2} & \textbf{1.893}\textsuperscript{1} & \underline{0.526}\textsuperscript{2} & \underline{0.545}\textsuperscript{2} & \underline{61.53}\textsuperscript{2} & \textbf{0.487}\textsuperscript{1} & \textbf{0.517}\textsuperscript{1} & 0.103 & \textbf{2.00} \\
        \midrule
        \multirow{13}{*}{B}
         & \textcolor{gray}{\textit{4KAgent}~\cite{zuo4KAgentAgenticAny2026}} & \textcolor{gray}{\textit{20.90}} & \textcolor{gray}{\textit{0.844}} & \textcolor{gray}{\textit{0.300}} & \textcolor{gray}{\textit{1.805}} & \textcolor{gray}{\textit{0.570}} & \textcolor{gray}{\textit{0.557}} & \textcolor{gray}{\textit{62.76}} & \textcolor{gray}{\textit{0.528}} & \textcolor{gray}{\textit{0.554}} & \textcolor{gray}{\textit{N/A}} & \textcolor{gray}{\textit{N/A}} \\
         & \textcolor{gray}{\textit{\ARCIR{}}} & \textcolor{gray}{\textit{21.51}} & \textcolor{gray}{\textit{0.861}} & \textcolor{gray}{\textit{0.287}} & \textcolor{gray}{\textit{1.698}} & \textcolor{gray}{\textit{0.564}} & \textcolor{gray}{\textit{0.552}} & \textcolor{gray}{\textit{62.65}} & \textcolor{gray}{\textit{0.520}} & \textcolor{gray}{\textit{0.547}} & \textcolor{gray}{\textit{N/A}} & \textcolor{gray}{\textit{N/A}} \\
         & VVC~\cite{brossOverviewVersatileVideo2021} & 18.86\textsuperscript{9} & 0.725\textsuperscript{10} & 0.551\textsuperscript{10} & 3.426\textsuperscript{7} & 0.422\textsuperscript{6} & 0.280\textsuperscript{11} & 42.23\textsuperscript{10} & 0.342\textsuperscript{7} & 0.387\textsuperscript{11} & 0.143 & 9.00 \\
         & DCAE~\cite{luLearnedImageCompression2025} & 20.24\textsuperscript{5} & 0.789\textsuperscript{7} & 0.507\textsuperscript{9} & 3.662\textsuperscript{11} & 0.397\textsuperscript{8} & 0.393\textsuperscript{6} & 46.14\textsuperscript{8} & 0.322\textsuperscript{10} & 0.424\textsuperscript{6} & 0.117 & 7.78 \\
         & CoD-Lite~\cite{jiaCoDLiteRealTimeDiffusionBased2026} & 18.37\textsuperscript{10} & 0.684\textsuperscript{11} & 0.574\textsuperscript{11} & 3.553\textsuperscript{10} & 0.410\textsuperscript{7} & 0.390\textsuperscript{7} & 40.46\textsuperscript{11} & 0.377\textsuperscript{6} & 0.396\textsuperscript{9.5} & 0.091 & 9.17 \\
         & AllInOne~\cite{zengAllinOneImageCompression2025} & 18.89\textsuperscript{8} & 0.788\textsuperscript{8} & 0.475\textsuperscript{7} & 3.549\textsuperscript{9} & 0.269\textsuperscript{9.5} & 0.369\textsuperscript{8} & 47.03\textsuperscript{6} & 0.328\textsuperscript{9} & 0.415\textsuperscript{7} & 0.291 & 7.94 \\
         & AirNet + DCAE & 18.33\textsuperscript{11} & 0.801\textsuperscript{5} & 0.465\textsuperscript{6} & 3.436\textsuperscript{8} & 0.269\textsuperscript{9.5} & 0.357\textsuperscript{10} & 46.30\textsuperscript{7} & 0.329\textsuperscript{8} & 0.409\textsuperscript{8} & 0.110 & 8.06 \\
         & DCAE + AirNet & 19.22\textsuperscript{7} & 0.795\textsuperscript{6} & 0.476\textsuperscript{8} & 3.247\textsuperscript{6} & 0.252\textsuperscript{11} & 0.361\textsuperscript{9} & 45.08\textsuperscript{9} & 0.316\textsuperscript{11} & 0.396\textsuperscript{9.5} & 0.104 & 8.50 \\
         & \ARCIR{} + DCAE & \textbf{21.49}\textsuperscript{1} & \textbf{0.858}\textsuperscript{1} & \textit{0.367}\textsuperscript{3} & 1.997\textsuperscript{4} & 0.508\textsuperscript{5} & 0.452\textsuperscript{5} & \textit{61.27}\textsuperscript{3} & 0.467\textsuperscript{5} & 0.490\textsuperscript{5} & 0.148 & 3.56 \\
         & \ARCIR{} + CoD-Lite & \textit{20.85}\textsuperscript{3} & \underline{0.834}\textsuperscript{2} & \textbf{0.301}\textsuperscript{1} & \underline{1.933}\textsuperscript{2} & \textbf{0.549}\textsuperscript{1} & \textbf{0.565}\textsuperscript{1} & 61.12\textsuperscript{4} & 0.483\textsuperscript{4} & 0.496\textsuperscript{4} & 0.125 & \underline{2.44} \\
         & DCAE + \ARCIR{} & 20.40\textsuperscript{4} & 0.803\textsuperscript{4} & 0.400\textsuperscript{5} & \textit{1.991}\textsuperscript{3} & 0.512\textsuperscript{4} & \textit{0.521}\textsuperscript{3} & \textbf{62.39}\textsuperscript{1} & \underline{0.501}\textsuperscript{2} & \textbf{0.535}\textsuperscript{1} & 0.117 & \textit{3.00} \\
         & CoD-Lite + \ARCIR{} & 19.41\textsuperscript{6} & 0.756\textsuperscript{9} & 0.378\textsuperscript{4} & 2.168\textsuperscript{5} & \textit{0.525}\textsuperscript{3} & 0.509\textsuperscript{4} & 60.98\textsuperscript{5} & \textit{0.489}\textsuperscript{3} & \textit{0.524}\textsuperscript{3} & 0.096 & 4.67 \\
         & ARC & \underline{20.87}\textsuperscript{2} & \textit{0.832}\textsuperscript{3} & \underline{0.318}\textsuperscript{2} & \textbf{1.785}\textsuperscript{1} & \underline{0.544}\textsuperscript{2} & \underline{0.549}\textsuperscript{2} & \underline{62.30}\textsuperscript{2} & \textbf{0.503}\textsuperscript{1} & \underline{0.534}\textsuperscript{2} & 0.110 & \textbf{1.89} \\
        \midrule
        \multirow{13}{*}{C}
         & \textcolor{gray}{\textit{4KAgent}~\cite{zuo4KAgentAgenticAny2026}} & \textcolor{gray}{\textit{19.92}} & \textcolor{gray}{\textit{0.779}} & \textcolor{gray}{\textit{0.422}} & \textcolor{gray}{\textit{2.666}} & \textcolor{gray}{\textit{0.495}} & \textcolor{gray}{\textit{0.531}} & \textcolor{gray}{\textit{55.71}} & \textcolor{gray}{\textit{0.495}} & \textcolor{gray}{\textit{0.528}} & \textcolor{gray}{\textit{N/A}} & \textcolor{gray}{\textit{N/A}} \\
         & \textcolor{gray}{\textit{\ARCIR{}}} & \textcolor{gray}{\textit{20.34}} & \textcolor{gray}{\textit{0.794}} & \textcolor{gray}{\textit{0.398}} & \textcolor{gray}{\textit{2.442}} & \textcolor{gray}{\textit{0.483}} & \textcolor{gray}{\textit{0.527}} & \textcolor{gray}{\textit{57.27}} & \textcolor{gray}{\textit{0.476}} & \textcolor{gray}{\textit{0.515}} & \textcolor{gray}{\textit{N/A}} & \textcolor{gray}{\textit{N/A}} \\
         & VVC~\cite{brossOverviewVersatileVideo2021} & 17.63\textsuperscript{9} & 0.632\textsuperscript{10} & 0.697\textsuperscript{10} & 4.627\textsuperscript{7} & 0.291\textsuperscript{7} & 0.283\textsuperscript{11} & 25.81\textsuperscript{10} & 0.293\textsuperscript{7} & 0.361\textsuperscript{6} & 0.097 & 8.56 \\
         & DCAE~\cite{luLearnedImageCompression2025} & 19.13\textsuperscript{6} & 0.745\textsuperscript{5} & 0.629\textsuperscript{7} & 5.014\textsuperscript{11} & 0.287\textsuperscript{8} & 0.374\textsuperscript{6} & 29.65\textsuperscript{6} & 0.207\textsuperscript{8} & 0.345\textsuperscript{8} & 0.084 & 7.22 \\
         & CoD-Lite~\cite{jiaCoDLiteRealTimeDiffusionBased2026} & 17.36\textsuperscript{10} & 0.601\textsuperscript{11} & 0.729\textsuperscript{11} & 4.603\textsuperscript{6} & 0.309\textsuperscript{6} & 0.335\textsuperscript{8} & 25.80\textsuperscript{11} & 0.339\textsuperscript{6} & 0.353\textsuperscript{7} & 0.063 & 8.44 \\
         & AllInOne~\cite{zengAllinOneImageCompression2025} & 17.64\textsuperscript{8} & 0.724\textsuperscript{7} & 0.632\textsuperscript{8} & 5.007\textsuperscript{10} & 0.198\textsuperscript{11} & 0.322\textsuperscript{10} & 28.10\textsuperscript{9} & 0.190\textsuperscript{11} & 0.331\textsuperscript{11} & 0.297 & 9.44 \\
         & AirNet + DCAE & 17.91\textsuperscript{7} & 0.738\textsuperscript{6} & 0.614\textsuperscript{6} & 4.815\textsuperscript{8} & 0.199\textsuperscript{10} & 0.336\textsuperscript{7} & 29.38\textsuperscript{7} & 0.201\textsuperscript{10} & 0.334\textsuperscript{10} & 0.081 & 7.89 \\
         & DCAE + AirNet & 17.29\textsuperscript{11} & 0.717\textsuperscript{8} & 0.638\textsuperscript{9} & 4.858\textsuperscript{9} & 0.205\textsuperscript{9} & 0.329\textsuperscript{9} & 28.96\textsuperscript{8} & 0.202\textsuperscript{9} & 0.341\textsuperscript{9} & 0.068 & 9.00 \\
         & \ARCIR{} + DCAE & \textbf{20.48}\textsuperscript{1} & \textbf{0.797}\textsuperscript{1} & \textit{0.458}\textsuperscript{3} & 2.703\textsuperscript{4} & 0.432\textsuperscript{5} & 0.414\textsuperscript{5} & 55.86\textsuperscript{4} & 0.426\textsuperscript{5} & 0.473\textsuperscript{4} & 0.129 & 3.56 \\
         & \ARCIR{} + CoD-Lite & \underline{20.32}\textsuperscript{2} & \underline{0.787}\textsuperscript{2} & \textbf{0.399}\textsuperscript{1} & \underline{2.600}\textsuperscript{2} & \textit{0.477}\textsuperscript{3} & \textit{0.521}\textsuperscript{3} & 55.52\textsuperscript{5} & 0.439\textsuperscript{4} & 0.458\textsuperscript{5} & 0.125 & \textit{3.00} \\
         & DCAE + \ARCIR{} & 19.70\textsuperscript{4} & 0.748\textsuperscript{4} & 0.477\textsuperscript{5} & \textit{2.646}\textsuperscript{3} & 0.473\textsuperscript{4} & \underline{0.531}\textsuperscript{2} & \textbf{58.64}\textsuperscript{1} & \textbf{0.490}\textsuperscript{1} & \textbf{0.536}\textsuperscript{1} & 0.084 & \underline{2.78} \\
         & CoD-Lite + \ARCIR{} & 19.14\textsuperscript{5} & 0.679\textsuperscript{9} & 0.472\textsuperscript{4} & 2.900\textsuperscript{5} & \textbf{0.482}\textsuperscript{1} & 0.503\textsuperscript{4} & \textit{55.95}\textsuperscript{3} & \textit{0.472}\textsuperscript{3} & \textit{0.501}\textsuperscript{3} & 0.066 & 4.11 \\
         & ARC & \textit{20.30}\textsuperscript{3} & \textit{0.786}\textsuperscript{3} & \underline{0.409}\textsuperscript{2} & \textbf{2.521}\textsuperscript{1} & \underline{0.478}\textsuperscript{2} & \textbf{0.538}\textsuperscript{1} & \underline{57.37}\textsuperscript{2} & \underline{0.477}\textsuperscript{2} & \underline{0.514}\textsuperscript{2} & 0.126 & \textbf{2.00} \\
        \bottomrule
    \end{tabular}
    \caption{\textbf{Results on MiO100 at the lower interpolation points, with actual BPP.} Each method from Table~\ref{tab:mio100_bpp015_interp} shows its measured operating point closest to 0.15\,bpp at or below it, i.e., the lower input of the interpolation. AllInOne and the AirNet combinations are not in the main table and show their own measured points.}\label{tab:mio100_bpp01}
\end{table*}

\begin{table*}[th]
    \centering
    \footnotesize
    \setlength{\tabcolsep}{2pt}
    \begin{tabular}{ll*{4}{>{\columncolor{frbg}}c}*{5}{>{\columncolor{nrbg}}c}cc}
        \toprule
        G & Method & PSNR $\uparrow$ & MS-SSIM $\uparrow$ & LPIPS $\downarrow$ & PieAPP $\downarrow$ & MANIQA $\uparrow$ & CLIPIQA $\uparrow$ & MUSIQ $\uparrow$ & TOPIQ $\uparrow$ & DBCNN $\uparrow$ & BPP \textdownarrow & Avg. rank $\downarrow$ \\
        \midrule
        \multirow{13}{*}{A}
         & \textcolor{gray}{\textit{4KAgent}~\cite{zuo4KAgentAgenticAny2026}} & \textcolor{gray}{\textit{21.46}} & \textcolor{gray}{\textit{0.843}} & \textcolor{gray}{\textit{0.302}} & \textcolor{gray}{\textit{1.782}} & \textcolor{gray}{\textit{0.562}} & \textcolor{gray}{\textit{0.559}} & \textcolor{gray}{\textit{63.48}} & \textcolor{gray}{\textit{0.528}} & \textcolor{gray}{\textit{0.554}} & \textcolor{gray}{\textit{N/A}} & \textcolor{gray}{\textit{N/A}} \\
         & \textcolor{gray}{\textit{\ARCIR{}}} & \textcolor{gray}{\textit{21.77}} & \textcolor{gray}{\textit{0.853}} & \textcolor{gray}{\textit{0.293}} & \textcolor{gray}{\textit{1.708}} & \textcolor{gray}{\textit{0.553}} & \textcolor{gray}{\textit{0.552}} & \textcolor{gray}{\textit{62.90}} & \textcolor{gray}{\textit{0.511}} & \textcolor{gray}{\textit{0.541}} & \textcolor{gray}{\textit{N/A}} & \textcolor{gray}{\textit{N/A}} \\
         & VVC~\cite{brossOverviewVersatileVideo2021} & 18.99\textsuperscript{8} & 0.739\textsuperscript{10} & 0.536\textsuperscript{10} & 3.866\textsuperscript{8} & 0.418\textsuperscript{6} & 0.320\textsuperscript{11} & 39.75\textsuperscript{10} & 0.348\textsuperscript{7} & 0.412\textsuperscript{6} & 0.292 & 8.44 \\
         & DCAE~\cite{luLearnedImageCompression2025} & 20.34\textsuperscript{6} & 0.807\textsuperscript{6} & 0.498\textsuperscript{7.5} & 3.932\textsuperscript{11} & 0.408\textsuperscript{7} & 0.412\textsuperscript{6} & 42.96\textsuperscript{6} & 0.312\textsuperscript{8} & 0.410\textsuperscript{7} & 0.239 & 7.17 \\
         & CoD-Lite~\cite{jiaCoDLiteRealTimeDiffusionBased2026} & 18.84\textsuperscript{9} & 0.734\textsuperscript{11} & 0.537\textsuperscript{11} & 3.888\textsuperscript{9} & 0.407\textsuperscript{8} & 0.362\textsuperscript{10} & 39.16\textsuperscript{11} & 0.361\textsuperscript{6} & 0.407\textsuperscript{8} & 0.365 & 9.22 \\
         & AllInOne~\cite{zengAllinOneImageCompression2025} & 18.60\textsuperscript{11} & 0.780\textsuperscript{9} & 0.514\textsuperscript{9} & 3.915\textsuperscript{10} & 0.249\textsuperscript{10} & 0.384\textsuperscript{8} & 41.69\textsuperscript{9} & 0.295\textsuperscript{10.5} & 0.401\textsuperscript{10} & 0.336 & 9.61 \\
         & AirNet + DCAE & 19.25\textsuperscript{7} & 0.803\textsuperscript{7} & 0.482\textsuperscript{6} & 3.555\textsuperscript{6} & 0.255\textsuperscript{9} & 0.395\textsuperscript{7} & 42.57\textsuperscript{7} & 0.306\textsuperscript{9} & 0.403\textsuperscript{9} & 0.207 & 7.44 \\
         & DCAE + AirNet & 18.75\textsuperscript{10} & 0.791\textsuperscript{8} & 0.498\textsuperscript{7.5} & 3.666\textsuperscript{7} & 0.247\textsuperscript{11} & 0.381\textsuperscript{9} & 41.73\textsuperscript{8} & 0.295\textsuperscript{10.5} & 0.394\textsuperscript{11} & 0.137 & 9.11 \\
         & \ARCIR{} + DCAE & \textbf{21.76}\textsuperscript{1} & \textbf{0.852}\textsuperscript{1} & 0.355\textsuperscript{4} & \textit{1.890}\textsuperscript{3} & \textit{0.529}\textsuperscript{3.5} & 0.467\textsuperscript{5} & \underline{62.38}\textsuperscript{2} & 0.476\textsuperscript{4} & 0.493\textsuperscript{5} & 0.205 & \textit{3.17} \\
         & \ARCIR{} + CoD-Lite & \underline{21.58}\textsuperscript{2} & \underline{0.849}\textsuperscript{2} & \textbf{0.298}\textsuperscript{1} & \textbf{1.770}\textsuperscript{1} & \textbf{0.560}\textsuperscript{1} & \textbf{0.555}\textsuperscript{1} & \textbf{62.39}\textsuperscript{1} & \underline{0.490}\textsuperscript{2} & 0.500\textsuperscript{4} & 0.500 & \textbf{1.67} \\
         & DCAE + \ARCIR{} & 21.16\textsuperscript{4} & 0.824\textsuperscript{4} & 0.370\textsuperscript{5} & 2.085\textsuperscript{5} & 0.525\textsuperscript{5} & \textit{0.521}\textsuperscript{3} & 60.65\textsuperscript{4} & \textit{0.484}\textsuperscript{3} & \underline{0.518}\textsuperscript{2} & 0.239 & 3.89 \\
         & CoD-Lite + \ARCIR{} & 20.86\textsuperscript{5} & 0.821\textsuperscript{5} & \textit{0.341}\textsuperscript{3} & 2.018\textsuperscript{4} & \textit{0.529}\textsuperscript{3.5} & 0.518\textsuperscript{4} & 60.01\textsuperscript{5} & 0.475\textsuperscript{5} & \textit{0.503}\textsuperscript{3} & 0.383 & 4.17 \\
         & ARC & \textit{21.26}\textsuperscript{3} & \textit{0.833}\textsuperscript{3} & \underline{0.325}\textsuperscript{2} & \underline{1.846}\textsuperscript{2} & \underline{0.531}\textsuperscript{2} & \underline{0.543}\textsuperscript{2} & \textit{61.76}\textsuperscript{3} & \textbf{0.491}\textsuperscript{1} & \textbf{0.520}\textsuperscript{1} & 0.161 & \underline{2.11} \\
        \midrule
        \multirow{13}{*}{B}
         & \textcolor{gray}{\textit{4KAgent}~\cite{zuo4KAgentAgenticAny2026}} & \textcolor{gray}{\textit{20.90}} & \textcolor{gray}{\textit{0.844}} & \textcolor{gray}{\textit{0.300}} & \textcolor{gray}{\textit{1.805}} & \textcolor{gray}{\textit{0.570}} & \textcolor{gray}{\textit{0.557}} & \textcolor{gray}{\textit{62.76}} & \textcolor{gray}{\textit{0.528}} & \textcolor{gray}{\textit{0.554}} & \textcolor{gray}{\textit{N/A}} & \textcolor{gray}{\textit{N/A}} \\
         & \textcolor{gray}{\textit{\ARCIR{}}} & \textcolor{gray}{\textit{21.51}} & \textcolor{gray}{\textit{0.861}} & \textcolor{gray}{\textit{0.287}} & \textcolor{gray}{\textit{1.698}} & \textcolor{gray}{\textit{0.564}} & \textcolor{gray}{\textit{0.552}} & \textcolor{gray}{\textit{62.65}} & \textcolor{gray}{\textit{0.520}} & \textcolor{gray}{\textit{0.547}} & \textcolor{gray}{\textit{N/A}} & \textcolor{gray}{\textit{N/A}} \\
         & VVC~\cite{brossOverviewVersatileVideo2021} & 18.89\textsuperscript{8.5} & 0.729\textsuperscript{10} & 0.530\textsuperscript{10} & 3.422\textsuperscript{8} & 0.456\textsuperscript{6} & 0.313\textsuperscript{11} & 44.02\textsuperscript{10} & 0.387\textsuperscript{6} & 0.442\textsuperscript{6} & 0.289 & 8.39 \\
         & DCAE~\cite{luLearnedImageCompression2025} & 20.34\textsuperscript{6} & 0.795\textsuperscript{8} & 0.494\textsuperscript{9} & 3.623\textsuperscript{11} & 0.428\textsuperscript{8} & 0.404\textsuperscript{6} & 47.30\textsuperscript{6} & 0.333\textsuperscript{10} & 0.427\textsuperscript{7} & 0.159 & 7.89 \\
         & CoD-Lite~\cite{jiaCoDLiteRealTimeDiffusionBased2026} & 18.73\textsuperscript{10} & 0.721\textsuperscript{11} & 0.569\textsuperscript{11} & 3.536\textsuperscript{9} & 0.432\textsuperscript{7} & 0.366\textsuperscript{10} & 40.84\textsuperscript{11} & 0.377\textsuperscript{7} & 0.390\textsuperscript{11} & 0.365 & 9.67 \\
         & AllInOne~\cite{zengAllinOneImageCompression2025} & 18.89\textsuperscript{8.5} & 0.788\textsuperscript{9} & 0.475\textsuperscript{8} & 3.549\textsuperscript{10} & 0.269\textsuperscript{10} & 0.369\textsuperscript{9} & 47.03\textsuperscript{8} & 0.328\textsuperscript{11} & 0.415\textsuperscript{9} & 0.291 & 9.17 \\
         & AirNet + DCAE & 18.49\textsuperscript{11} & 0.813\textsuperscript{6} & 0.441\textsuperscript{6} & 3.325\textsuperscript{7} & 0.272\textsuperscript{9} & 0.376\textsuperscript{8} & 47.26\textsuperscript{7} & 0.344\textsuperscript{8} & 0.417\textsuperscript{8} & 0.177 & 7.78 \\
         & DCAE + AirNet & 19.41\textsuperscript{7} & 0.805\textsuperscript{7} & 0.456\textsuperscript{7} & 3.183\textsuperscript{6} & 0.264\textsuperscript{11} & 0.387\textsuperscript{7} & 46.71\textsuperscript{9} & 0.340\textsuperscript{9} & 0.414\textsuperscript{10} & 0.178 & 8.11 \\
         & \ARCIR{} + DCAE & \textbf{21.50}\textsuperscript{1} & \textbf{0.860}\textsuperscript{1} & 0.352\textsuperscript{4} & \textit{1.924}\textsuperscript{3} & 0.533\textsuperscript{5} & 0.470\textsuperscript{5} & 62.22\textsuperscript{4} & 0.481\textsuperscript{5} & 0.498\textsuperscript{5} & 0.203 & 3.67 \\
         & \ARCIR{} + CoD-Lite & \underline{21.34}\textsuperscript{2} & \underline{0.857}\textsuperscript{2} & \textbf{0.292}\textsuperscript{1} & \underline{1.770}\textsuperscript{2} & \textbf{0.567}\textsuperscript{1} & \textbf{0.558}\textsuperscript{1} & \textit{62.23}\textsuperscript{3} & 0.495\textsuperscript{4} & 0.505\textsuperscript{4} & 0.500 & \underline{2.22} \\
         & DCAE + \ARCIR{} & 20.60\textsuperscript{4.5} & 0.814\textsuperscript{5} & 0.383\textsuperscript{5} & 1.950\textsuperscript{5} & 0.537\textsuperscript{4} & \textit{0.528}\textsuperscript{3} & \textbf{63.19}\textsuperscript{1} & \textbf{0.514}\textsuperscript{1} & \textbf{0.540}\textsuperscript{1} & 0.159 & \textit{3.28} \\
         & CoD-Lite + \ARCIR{} & 20.60\textsuperscript{4.5} & 0.824\textsuperscript{4} & \textit{0.333}\textsuperscript{3} & 1.931\textsuperscript{4} & \textit{0.545}\textsuperscript{3} & 0.526\textsuperscript{4} & 61.72\textsuperscript{5} & \textit{0.503}\textsuperscript{3} & \textit{0.530}\textsuperscript{3} & 0.383 & 3.72 \\
         & ARC & \textit{21.17}\textsuperscript{3} & \textit{0.846}\textsuperscript{3} & \underline{0.311}\textsuperscript{2} & \textbf{1.745}\textsuperscript{1} & \underline{0.550}\textsuperscript{2} & \underline{0.549}\textsuperscript{2} & \underline{62.58}\textsuperscript{2} & \underline{0.505}\textsuperscript{2} & \underline{0.537}\textsuperscript{2} & 0.176 & \textbf{2.11} \\
        \midrule
        \multirow{13}{*}{C}
         & \textcolor{gray}{\textit{4KAgent}~\cite{zuo4KAgentAgenticAny2026}} & \textcolor{gray}{\textit{19.92}} & \textcolor{gray}{\textit{0.779}} & \textcolor{gray}{\textit{0.422}} & \textcolor{gray}{\textit{2.666}} & \textcolor{gray}{\textit{0.495}} & \textcolor{gray}{\textit{0.531}} & \textcolor{gray}{\textit{55.71}} & \textcolor{gray}{\textit{0.495}} & \textcolor{gray}{\textit{0.528}} & \textcolor{gray}{\textit{N/A}} & \textcolor{gray}{\textit{N/A}} \\
         & \textcolor{gray}{\textit{\ARCIR{}}} & \textcolor{gray}{\textit{20.34}} & \textcolor{gray}{\textit{0.794}} & \textcolor{gray}{\textit{0.398}} & \textcolor{gray}{\textit{2.442}} & \textcolor{gray}{\textit{0.483}} & \textcolor{gray}{\textit{0.527}} & \textcolor{gray}{\textit{57.27}} & \textcolor{gray}{\textit{0.476}} & \textcolor{gray}{\textit{0.515}} & \textcolor{gray}{\textit{N/A}} & \textcolor{gray}{\textit{N/A}} \\
         & VVC~\cite{brossOverviewVersatileVideo2021} & 17.46\textsuperscript{9} & 0.625\textsuperscript{10} & 0.691\textsuperscript{10} & 4.608\textsuperscript{6} & 0.301\textsuperscript{7} & 0.311\textsuperscript{10} & 25.92\textsuperscript{10} & 0.322\textsuperscript{7} & 0.398\textsuperscript{6} & 0.202 & 8.33 \\
         & DCAE~\cite{luLearnedImageCompression2025} & 19.05\textsuperscript{6} & 0.746\textsuperscript{6} & 0.635\textsuperscript{8} & 4.959\textsuperscript{10} & 0.293\textsuperscript{8} & 0.377\textsuperscript{6} & 28.67\textsuperscript{7} & 0.229\textsuperscript{8} & 0.358\textsuperscript{7} & 0.175 & 7.33 \\
         & CoD-Lite~\cite{jiaCoDLiteRealTimeDiffusionBased2026} & 17.40\textsuperscript{10} & 0.615\textsuperscript{11} & 0.729\textsuperscript{11} & 4.609\textsuperscript{7} & 0.312\textsuperscript{6} & 0.310\textsuperscript{11} & 25.72\textsuperscript{11} & 0.333\textsuperscript{6} & 0.347\textsuperscript{9} & 0.253 & 9.11 \\
         & AllInOne~\cite{zengAllinOneImageCompression2025} & 17.64\textsuperscript{8} & 0.724\textsuperscript{8} & 0.632\textsuperscript{7} & 5.007\textsuperscript{11} & 0.198\textsuperscript{10} & 0.322\textsuperscript{9} & 28.10\textsuperscript{8} & 0.190\textsuperscript{11} & 0.331\textsuperscript{11} & 0.297 & 9.22 \\
         & AirNet + DCAE & 17.91\textsuperscript{7} & 0.738\textsuperscript{7} & 0.614\textsuperscript{6} & 4.815\textsuperscript{9} & 0.199\textsuperscript{9} & 0.336\textsuperscript{8} & 29.38\textsuperscript{6} & 0.201\textsuperscript{10} & 0.334\textsuperscript{10} & 0.081 & 8.00 \\
         & DCAE + AirNet & 17.29\textsuperscript{11} & 0.716\textsuperscript{9} & 0.673\textsuperscript{9} & 4.763\textsuperscript{8} & 0.195\textsuperscript{11} & 0.346\textsuperscript{7} & 27.97\textsuperscript{9} & 0.221\textsuperscript{9} & 0.350\textsuperscript{8} & 0.210 & 9.00 \\
         & \ARCIR{} + DCAE & \textbf{20.45}\textsuperscript{1} & \textbf{0.797}\textsuperscript{1} & 0.445\textsuperscript{4} & 2.626\textsuperscript{4} & 0.456\textsuperscript{5} & 0.434\textsuperscript{5} & \textit{57.13}\textsuperscript{3} & 0.444\textsuperscript{5} & 0.485\textsuperscript{4} & 0.188 & 3.56 \\
         & \ARCIR{} + CoD-Lite & \underline{20.37}\textsuperscript{2} & \underline{0.794}\textsuperscript{2} & \textbf{0.399}\textsuperscript{1} & \textbf{2.481}\textsuperscript{1} & \textit{0.488}\textsuperscript{3} & 0.513\textsuperscript{4} & 56.83\textsuperscript{4} & 0.454\textsuperscript{4} & 0.470\textsuperscript{5} & 0.500 & \textit{2.89} \\
         & DCAE + \ARCIR{} & 19.72\textsuperscript{5} & 0.754\textsuperscript{4} & 0.459\textsuperscript{5} & \textit{2.561}\textsuperscript{3} & \underline{0.493}\textsuperscript{2} & \textbf{0.543}\textsuperscript{1} & \textbf{58.87}\textsuperscript{1} & \textbf{0.511}\textsuperscript{1} & \textbf{0.550}\textsuperscript{1} & 0.175 & \textbf{2.56} \\
         & CoD-Lite + \ARCIR{} & 19.97\textsuperscript{4} & 0.750\textsuperscript{5} & \textit{0.434}\textsuperscript{3} & 2.692\textsuperscript{5} & \textbf{0.494}\textsuperscript{1} & \textit{0.518}\textsuperscript{3} & 56.55\textsuperscript{5} & \underline{0.493}\textsuperscript{2} & \underline{0.518}\textsuperscript{2} & 0.266 & 3.33 \\
         & ARC$^{\ddagger}$ & \textit{20.30}\textsuperscript{3} & \textit{0.786}\textsuperscript{3} & \underline{0.409}\textsuperscript{2} & \underline{2.521}\textsuperscript{2} & 0.478\textsuperscript{4} & \underline{0.538}\textsuperscript{2} & \underline{57.37}\textsuperscript{2} & \textit{0.477}\textsuperscript{3} & \textit{0.514}\textsuperscript{3} & 0.126 & \underline{2.67} \\
        \bottomrule
    \end{tabular}
    \caption{\textbf{Results on MiO100 at the upper interpolation points, with actual BPP.} Each method from Table~\ref{tab:mio100_bpp015_interp} shows its measured operating point closest to 0.15\,bpp at or above it, i.e., the upper input of the interpolation. $^{\ddagger}$~No measured point lies on that side of 0.15\,bpp; the nearest point is shown, as in Table~\ref{tab:mio100_bpp015_interp}. AllInOne and the AirNet combinations are not in the main table and show their own measured points. }\label{tab:mio100_bpp02}
\end{table*}

\subsection{Additional Rate-Distortion Curves}\label{sec:supp_rd_curves}

Figure~\ref{fig:rd_mio100_lpips_dbcnn} shows LPIPS and DBCNN. Figures~\ref{fig:rd_mio100}--\ref{fig:rd_mio100_topiq} show the MiO100 rate-distortion sweeps for PSNR, MS-SSIM, MANIQA, CLIPIQA, PieAPP, MUSIQ, and TOPIQ.

\begin{figure*}[th]
    \centering
    \footnotesize
    \begin{subfigure}[t]{0.32\textwidth}
        \centering
        \begin{tikzpicture}
            \begin{axis}[
                width=\textwidth,
                height=0.8\textwidth,
                xlabel near ticks,
                ylabel near ticks,
                xlabel={BPP},
                ylabel={PSNR (dB)},
                grid=major,
                grid style={dashed, gray!30},
                xmin=0, xmax=0.53,
                ymin=14, ymax=22,
            ]
            \addplot[black, mark=*, thick] table[x=bpp,y=psnr,col sep=comma] {data/rd_curves/a_vvc.csv};
            \addplot[blue, mark=*, thick] table[x=bpp,y=psnr,col sep=comma] {data/rd_curves/a_dcae.csv};
            \addplot[teal, mark=*, thick] table[x=bpp,y=psnr,col sep=comma] {data/rd_curves/a_cod_lite.csv};
            \addplot[orange, mark=*, thick] table[x=bpp,y=psnr,col sep=comma] {data/rd_curves/a_allinone.csv};
            \addplot[blue, mark=square*, thick] table[x=bpp,y=psnr,col sep=comma] {data/rd_curves/a_arcir_dcae_v3.csv};
            \addplot[teal, mark=square*, thick] table[x=bpp,y=psnr,col sep=comma] {data/rd_curves/a_arcir_cod_lite_v3.csv};
            \addplot[violet, mark=square*, thick] table[x=bpp,y=psnr,col sep=comma] {data/rd_curves/a_airnet_dcae.csv};
            \addplot[blue, mark=triangle*, thick, dashed] table[x=bpp,y=psnr,col sep=comma] {data/rd_curves/a_dcae_arcir_v3.csv};
            \addplot[teal, mark=triangle*, thick, dashed] table[x=bpp,y=psnr,col sep=comma] {data/rd_curves/a_cod_lite_arcir_v3.csv};
            \addplot[violet, mark=triangle*, thick, dashed] table[x=bpp,y=psnr,col sep=comma] {data/rd_curves/a_dcae_airnet.csv};
            \addplot[red, mark=pentagon*, thick, mark size=3pt, mark options={fill=red!50!white}] table[x=bpp,y=psnr,col sep=comma] {data/rd_curves/a_arc_v3.csv};
\end{axis}
        \end{tikzpicture}
        \caption{Group A}\label{fig:rd_mio100_a}
    \end{subfigure}
    \hfill
    \begin{subfigure}[t]{0.32\textwidth}
        \centering
        \begin{tikzpicture}
            \begin{axis}[
                width=\textwidth,
                height=0.8\textwidth,
                xlabel near ticks,
                ylabel near ticks,
                xlabel={BPP},
                ylabel={PSNR (dB)},
                grid=major,
                grid style={dashed, gray!30},
                xmin=0, xmax=0.53,
                ymin=14, ymax=22,
            ]
            \addplot[black, mark=*, thick] table[x=bpp,y=psnr,col sep=comma] {data/rd_curves/b_vvc.csv};
            \addplot[blue, mark=*, thick] table[x=bpp,y=psnr,col sep=comma] {data/rd_curves/b_dcae.csv};
            \addplot[teal, mark=*, thick] table[x=bpp,y=psnr,col sep=comma] {data/rd_curves/b_cod_lite.csv};
            \addplot[orange, mark=*, thick] table[x=bpp,y=psnr,col sep=comma] {data/rd_curves/b_allinone.csv};
            \addplot[blue, mark=square*, thick] table[x=bpp,y=psnr,col sep=comma] {data/rd_curves/b_arcir_dcae_v3.csv};
            \addplot[teal, mark=square*, thick] table[x=bpp,y=psnr,col sep=comma] {data/rd_curves/b_arcir_cod_lite_v3.csv};
            \addplot[violet, mark=square*, thick] table[x=bpp,y=psnr,col sep=comma] {data/rd_curves/b_airnet_dcae.csv};
            \addplot[blue, mark=triangle*, thick, dashed] table[x=bpp,y=psnr,col sep=comma] {data/rd_curves/b_dcae_arcir_v3.csv};
            \addplot[teal, mark=triangle*, thick, dashed] table[x=bpp,y=psnr,col sep=comma] {data/rd_curves/b_cod_lite_arcir_v3.csv};
            \addplot[violet, mark=triangle*, thick, dashed] table[x=bpp,y=psnr,col sep=comma] {data/rd_curves/b_dcae_airnet.csv};
            \addplot[red, mark=pentagon*, thick, mark size=3pt, mark options={fill=red!50!white}] table[x=bpp,y=psnr,col sep=comma] {data/rd_curves/b_arc_v3.csv};
\end{axis}
        \end{tikzpicture}
        \caption{Group B}\label{fig:rd_mio100_b}
    \end{subfigure}
    \hfill
    \begin{subfigure}[t]{0.32\textwidth}
        \centering
        \begin{tikzpicture}
            \begin{axis}[
                width=\textwidth,
                height=0.8\textwidth,
                xlabel near ticks,
                ylabel near ticks,
                xlabel={BPP},
                ylabel={PSNR (dB)},
                legend to name=miorlegend,
                legend columns=4,
                legend style={font=\footnotesize, legend cell align=left},
                grid=major,
                grid style={dashed, gray!30},
                xmin=0, xmax=0.53,
                ymin=14, ymax=22,
            ]
            \addlegendimage{red, mark=pentagon*, thick, mark size=3pt, mark options={fill=red!50!white}}
            \addlegendentry{ARC}
            \addplot[black, mark=*, thick] table[x=bpp,y=psnr,col sep=comma] {data/rd_curves/c_vvc.csv};
            \addlegendentry{VVC~\cite{brossOverviewVersatileVideo2021}}
            \addplot[blue, mark=*, thick] table[x=bpp,y=psnr,col sep=comma] {data/rd_curves/c_dcae.csv};
            \addlegendentry{DCAE~\cite{luLearnedImageCompression2025}}
            \addplot[teal, mark=*, thick] table[x=bpp,y=psnr,col sep=comma] {data/rd_curves/c_cod_lite.csv};
            \addlegendentry{CoD-Lite~\cite{jiaCoDLiteRealTimeDiffusionBased2026}}
            \addplot[orange, mark=*, thick] table[x=bpp,y=psnr,col sep=comma] {data/rd_curves/c_allinone.csv};
            \addlegendentry{AllInOne}
            \addplot[blue, mark=square*, thick] table[x=bpp,y=psnr,col sep=comma] {data/rd_curves/c_arcir_dcae_v3.csv};
            \addlegendentry{\ARCIR{} + DCAE}
            \addplot[teal, mark=square*, thick] table[x=bpp,y=psnr,col sep=comma] {data/rd_curves/c_arcir_cod_lite_v3.csv};
            \addlegendentry{\ARCIR{} + CoD-Lite}
            \addplot[violet, mark=square*, thick] table[x=bpp,y=psnr,col sep=comma] {data/rd_curves/c_airnet_dcae.csv};
            \addlegendentry{AirNet+DCAE}
            \addplot[blue, mark=triangle*, thick, dashed] table[x=bpp,y=psnr,col sep=comma] {data/rd_curves/c_dcae_arcir_v3.csv};
            \addlegendentry{DCAE + \ARCIR{}}
            \addplot[teal, mark=triangle*, thick, dashed] table[x=bpp,y=psnr,col sep=comma] {data/rd_curves/c_cod_lite_arcir_v3.csv};
            \addlegendentry{CoD-Lite + \ARCIR{}}
            \addplot[violet, mark=triangle*, thick, dashed] table[x=bpp,y=psnr,col sep=comma] {data/rd_curves/c_dcae_airnet.csv};
            \addlegendentry{DCAE+AirNet}
            \addplot[red, mark=pentagon*, thick, mark size=3pt, mark options={fill=red!50!white}, forget plot] table[x=bpp,y=psnr,col sep=comma] {data/rd_curves/c_arc_v3.csv};
\end{axis}
        \end{tikzpicture}
        \caption{Group C}\label{fig:rd_mio100_c}
    \end{subfigure}

    \centering
    \ref{miorlegend}

    \caption{\textbf{Rate-distortion performance on MiO100} for PSNR $\uparrow$ vs. BPP.}\label{fig:rd_mio100}
\end{figure*}
\begin{figure*}[t]
    \centering
    \footnotesize
    \begin{subfigure}[t]{0.32\textwidth}
        \centering
        \begin{tikzpicture}
            \begin{axis}[
                width=\textwidth,
                height=0.8\textwidth,
                xlabel near ticks,
                ylabel near ticks,
                xlabel={BPP},
                ylabel={MS-SSIM},
                grid=major,
                grid style={dashed, gray!30},
                xmin=0, xmax=0.53,
                ymin=0.4, ymax=0.9,
            ]
            \addplot[black, mark=*, thick] table[x=bpp,y=msssim,col sep=comma] {data/rd_curves/a_vvc.csv};
            \addplot[blue, mark=*, thick] table[x=bpp,y=msssim,col sep=comma] {data/rd_curves/a_dcae.csv};
            \addplot[teal, mark=*, thick] table[x=bpp,y=msssim,col sep=comma] {data/rd_curves/a_cod_lite.csv};
            \addplot[orange, mark=*, thick] table[x=bpp,y=msssim,col sep=comma] {data/rd_curves/a_allinone.csv};
            \addplot[blue, mark=square*, thick] table[x=bpp,y=msssim,col sep=comma] {data/rd_curves/a_arcir_dcae_v3.csv};
            \addplot[teal, mark=square*, thick] table[x=bpp,y=msssim,col sep=comma] {data/rd_curves/a_arcir_cod_lite_v3.csv};
            \addplot[violet, mark=square*, thick] table[x=bpp,y=msssim,col sep=comma] {data/rd_curves/a_airnet_dcae.csv};
            \addplot[blue, mark=triangle*, thick, dashed] table[x=bpp,y=msssim,col sep=comma] {data/rd_curves/a_dcae_arcir_v3.csv};
            \addplot[teal, mark=triangle*, thick, dashed] table[x=bpp,y=msssim,col sep=comma] {data/rd_curves/a_cod_lite_arcir_v3.csv};
            \addplot[violet, mark=triangle*, thick, dashed] table[x=bpp,y=msssim,col sep=comma] {data/rd_curves/a_dcae_airnet.csv};
            \addplot[red, mark=pentagon*, thick, mark size=3pt, mark options={fill=red!50!white}] table[x=bpp,y=msssim,col sep=comma] {data/rd_curves/a_arc_v3.csv};
\end{axis}
        \end{tikzpicture}
        \caption{Group A}\label{fig:rd_mio100_msssim_a}
    \end{subfigure}
    \hfill
    \begin{subfigure}[t]{0.32\textwidth}
        \centering
        \begin{tikzpicture}
            \begin{axis}[
                width=\textwidth,
                height=0.8\textwidth,
                xlabel near ticks,
                ylabel near ticks,
                xlabel={BPP},
                ylabel={MS-SSIM},
                grid=major,
                grid style={dashed, gray!30},
                xmin=0, xmax=0.53,
                ymin=0.4, ymax=0.9,
            ]
            \addplot[black, mark=*, thick] table[x=bpp,y=msssim,col sep=comma] {data/rd_curves/b_vvc.csv};
            \addplot[blue, mark=*, thick] table[x=bpp,y=msssim,col sep=comma] {data/rd_curves/b_dcae.csv};
            \addplot[teal, mark=*, thick] table[x=bpp,y=msssim,col sep=comma] {data/rd_curves/b_cod_lite.csv};
            \addplot[orange, mark=*, thick] table[x=bpp,y=msssim,col sep=comma] {data/rd_curves/b_allinone.csv};
            \addplot[blue, mark=square*, thick] table[x=bpp,y=msssim,col sep=comma] {data/rd_curves/b_arcir_dcae_v3.csv};
            \addplot[teal, mark=square*, thick] table[x=bpp,y=msssim,col sep=comma] {data/rd_curves/b_arcir_cod_lite_v3.csv};
            \addplot[violet, mark=square*, thick] table[x=bpp,y=msssim,col sep=comma] {data/rd_curves/b_airnet_dcae.csv};
            \addplot[blue, mark=triangle*, thick, dashed] table[x=bpp,y=msssim,col sep=comma] {data/rd_curves/b_dcae_arcir_v3.csv};
            \addplot[teal, mark=triangle*, thick, dashed] table[x=bpp,y=msssim,col sep=comma] {data/rd_curves/b_cod_lite_arcir_v3.csv};
            \addplot[violet, mark=triangle*, thick, dashed] table[x=bpp,y=msssim,col sep=comma] {data/rd_curves/b_dcae_airnet.csv};
            \addplot[red, mark=pentagon*, thick, mark size=3pt, mark options={fill=red!50!white}] table[x=bpp,y=msssim,col sep=comma] {data/rd_curves/b_arc_v3.csv};
\end{axis}
        \end{tikzpicture}
        \caption{Group B}\label{fig:rd_mio100_msssim_b}
    \end{subfigure}
    \hfill
    \begin{subfigure}[t]{0.32\textwidth}
        \centering
        \begin{tikzpicture}
            \begin{axis}[
                width=\textwidth,
                height=0.8\textwidth,
                xlabel near ticks,
                ylabel near ticks,
                xlabel={BPP},
                ylabel={MS-SSIM},
                grid=major,
                grid style={dashed, gray!30},
                xmin=0, xmax=0.53,
                ymin=0.4, ymax=0.9,
            ]
            \addplot[black, mark=*, thick] table[x=bpp,y=msssim,col sep=comma] {data/rd_curves/c_vvc.csv};
            \addplot[blue, mark=*, thick] table[x=bpp,y=msssim,col sep=comma] {data/rd_curves/c_dcae.csv};
            \addplot[teal, mark=*, thick] table[x=bpp,y=msssim,col sep=comma] {data/rd_curves/c_cod_lite.csv};
            \addplot[orange, mark=*, thick] table[x=bpp,y=msssim,col sep=comma] {data/rd_curves/c_allinone.csv};
            \addplot[blue, mark=square*, thick] table[x=bpp,y=msssim,col sep=comma] {data/rd_curves/c_arcir_dcae_v3.csv};
            \addplot[teal, mark=square*, thick] table[x=bpp,y=msssim,col sep=comma] {data/rd_curves/c_arcir_cod_lite_v3.csv};
            \addplot[violet, mark=square*, thick] table[x=bpp,y=msssim,col sep=comma] {data/rd_curves/c_airnet_dcae.csv};
            \addplot[blue, mark=triangle*, thick, dashed] table[x=bpp,y=msssim,col sep=comma] {data/rd_curves/c_dcae_arcir_v3.csv};
            \addplot[teal, mark=triangle*, thick, dashed] table[x=bpp,y=msssim,col sep=comma] {data/rd_curves/c_cod_lite_arcir_v3.csv};
            \addplot[violet, mark=triangle*, thick, dashed] table[x=bpp,y=msssim,col sep=comma] {data/rd_curves/c_dcae_airnet.csv};
            \addplot[red, mark=pentagon*, thick, mark size=3pt, mark options={fill=red!50!white}] table[x=bpp,y=msssim,col sep=comma] {data/rd_curves/c_arc_v3.csv};
\end{axis}
        \end{tikzpicture}
        \caption{Group C}\label{fig:rd_mio100_msssim_c}
    \end{subfigure}

    \centering
    \ref{miorlegend}

    \caption{\textbf{Rate-distortion performance on MiO100} for MS-SSIM $\uparrow$ vs. BPP.}\label{fig:rd_mio100_msssim}
\end{figure*}
\begin{figure*}[t]
    \centering
    \footnotesize
    \begin{subfigure}[t]{0.32\textwidth}
        \centering
        \begin{tikzpicture}
            \begin{axis}[
                width=\textwidth,
                height=0.8\textwidth,
                xlabel near ticks,
                ylabel near ticks,
                xlabel={BPP},
                ylabel={MANIQA},
                grid=major,
                grid style={dashed, gray!30},
                xmin=0, xmax=0.53,
                ymin=0.2, ymax=0.6,
            ]
            \addplot[black, mark=*, thick] table[x=bpp,y=maniqa,col sep=comma] {data/rd_curves/a_vvc.csv};
            \addplot[blue, mark=*, thick] table[x=bpp,y=maniqa,col sep=comma] {data/rd_curves/a_dcae.csv};
            \addplot[teal, mark=*, thick] table[x=bpp,y=maniqa,col sep=comma] {data/rd_curves/a_cod_lite.csv};
            \addplot[orange, mark=*, thick] table[x=bpp,y=maniqa,col sep=comma] {data/rd_curves/a_allinone.csv};
            \addplot[blue, mark=square*, thick] table[x=bpp,y=maniqa,col sep=comma] {data/rd_curves/a_arcir_dcae_v3.csv};
            \addplot[teal, mark=square*, thick] table[x=bpp,y=maniqa,col sep=comma] {data/rd_curves/a_arcir_cod_lite_v3.csv};
            \addplot[blue, mark=triangle*, thick, dashed] table[x=bpp,y=maniqa,col sep=comma] {data/rd_curves/a_dcae_arcir_v3.csv};
            \addplot[teal, mark=triangle*, thick, dashed] table[x=bpp,y=maniqa,col sep=comma] {data/rd_curves/a_cod_lite_arcir_v3.csv};
            \addplot[red, mark=pentagon*, thick, mark size=3pt, mark options={fill=red!50!white}] table[x=bpp,y=maniqa,col sep=comma] {data/rd_curves/a_arc_v3.csv};
\end{axis}
        \end{tikzpicture}
        \caption{Group A}\label{fig:rd_mio100_maniqa_a}
    \end{subfigure}
    \hfill
    \begin{subfigure}[t]{0.32\textwidth}
        \centering
        \begin{tikzpicture}
            \begin{axis}[
                width=\textwidth,
                height=0.8\textwidth,
                xlabel near ticks,
                ylabel near ticks,
                xlabel={BPP},
                ylabel={MANIQA},
                grid=major,
                grid style={dashed, gray!30},
                xmin=0, xmax=0.53,
                ymin=0.2, ymax=0.6,
            ]
            \addplot[black, mark=*, thick] table[x=bpp,y=maniqa,col sep=comma] {data/rd_curves/b_vvc.csv};
            \addplot[blue, mark=*, thick] table[x=bpp,y=maniqa,col sep=comma] {data/rd_curves/b_dcae.csv};
            \addplot[teal, mark=*, thick] table[x=bpp,y=maniqa,col sep=comma] {data/rd_curves/b_cod_lite.csv};
            \addplot[orange, mark=*, thick] table[x=bpp,y=maniqa,col sep=comma] {data/rd_curves/b_allinone.csv};
            \addplot[blue, mark=square*, thick] table[x=bpp,y=maniqa,col sep=comma] {data/rd_curves/b_arcir_dcae_v3.csv};
            \addplot[teal, mark=square*, thick] table[x=bpp,y=maniqa,col sep=comma] {data/rd_curves/b_arcir_cod_lite_v3.csv};
            \addplot[blue, mark=triangle*, thick, dashed] table[x=bpp,y=maniqa,col sep=comma] {data/rd_curves/b_dcae_arcir_v3.csv};
            \addplot[teal, mark=triangle*, thick, dashed] table[x=bpp,y=maniqa,col sep=comma] {data/rd_curves/b_cod_lite_arcir_v3.csv};
            \addplot[red, mark=pentagon*, thick, mark size=3pt, mark options={fill=red!50!white}] table[x=bpp,y=maniqa,col sep=comma] {data/rd_curves/b_arc_v3.csv};
\end{axis}
        \end{tikzpicture}
        \caption{Group B}\label{fig:rd_mio100_maniqa_b}
    \end{subfigure}
    \hfill
    \begin{subfigure}[t]{0.32\textwidth}
        \centering
        \begin{tikzpicture}
            \begin{axis}[
                width=\textwidth,
                height=0.8\textwidth,
                xlabel near ticks,
                ylabel near ticks,
                xlabel={BPP},
                ylabel={MANIQA},
                grid=major,
                grid style={dashed, gray!30},
                xmin=0, xmax=0.53,
                ymin=0.2, ymax=0.6,
            ]
            \addplot[black, mark=*, thick] table[x=bpp,y=maniqa,col sep=comma] {data/rd_curves/c_vvc.csv};
            \addplot[blue, mark=*, thick] table[x=bpp,y=maniqa,col sep=comma] {data/rd_curves/c_dcae.csv};
            \addplot[teal, mark=*, thick] table[x=bpp,y=maniqa,col sep=comma] {data/rd_curves/c_cod_lite.csv};
            \addplot[orange, mark=*, thick] table[x=bpp,y=maniqa,col sep=comma] {data/rd_curves/c_allinone.csv};
            \addplot[blue, mark=square*, thick] table[x=bpp,y=maniqa,col sep=comma] {data/rd_curves/c_arcir_dcae_v3.csv};
            \addplot[teal, mark=square*, thick] table[x=bpp,y=maniqa,col sep=comma] {data/rd_curves/c_arcir_cod_lite_v3.csv};
            \addplot[blue, mark=triangle*, thick, dashed] table[x=bpp,y=maniqa,col sep=comma] {data/rd_curves/c_dcae_arcir_v3.csv};
            \addplot[teal, mark=triangle*, thick, dashed] table[x=bpp,y=maniqa,col sep=comma] {data/rd_curves/c_cod_lite_arcir_v3.csv};
            \addplot[red, mark=pentagon*, thick, mark size=3pt, mark options={fill=red!50!white}] table[x=bpp,y=maniqa,col sep=comma] {data/rd_curves/c_arc_v3.csv};
\end{axis}
        \end{tikzpicture}
        \caption{Group C}\label{fig:rd_mio100_maniqa_c}
    \end{subfigure}

    \centering
    \ref{miorlegend}

    \caption{\textbf{Rate-distortion performance on MiO100} for MANIQA $\uparrow$ vs. BPP.}\label{fig:rd_mio100_maniqa}
\end{figure*}
\begin{figure*}[t]
    \centering
    \footnotesize
    \begin{subfigure}[t]{0.32\textwidth}
        \centering
        \begin{tikzpicture}
            \begin{axis}[
                width=\textwidth,
                height=0.8\textwidth,
                xlabel near ticks,
                ylabel near ticks,
                xlabel={BPP},
                ylabel={CLIPIQA},
                grid=major,
                grid style={dashed, gray!30},
                xmin=0, xmax=0.53,
                ymin=0.15, ymax=0.6,
            ]
            \addplot[black, mark=*, thick] table[x=bpp,y=clipiqa,col sep=comma] {data/rd_curves/a_vvc.csv};
            \addplot[blue, mark=*, thick] table[x=bpp,y=clipiqa,col sep=comma] {data/rd_curves/a_dcae.csv};
            \addplot[teal, mark=*, thick] table[x=bpp,y=clipiqa,col sep=comma] {data/rd_curves/a_cod_lite.csv};
            \addplot[orange, mark=*, thick] table[x=bpp,y=clipiqa,col sep=comma] {data/rd_curves/a_allinone.csv};
            \addplot[blue, mark=square*, thick] table[x=bpp,y=clipiqa,col sep=comma] {data/rd_curves/a_arcir_dcae_v3.csv};
            \addplot[teal, mark=square*, thick] table[x=bpp,y=clipiqa,col sep=comma] {data/rd_curves/a_arcir_cod_lite_v3.csv};
            \addplot[blue, mark=triangle*, thick, dashed] table[x=bpp,y=clipiqa,col sep=comma] {data/rd_curves/a_dcae_arcir_v3.csv};
            \addplot[teal, mark=triangle*, thick, dashed] table[x=bpp,y=clipiqa,col sep=comma] {data/rd_curves/a_cod_lite_arcir_v3.csv};
            \addplot[red, mark=pentagon*, thick, mark size=3pt, mark options={fill=red!50!white}] table[x=bpp,y=clipiqa,col sep=comma] {data/rd_curves/a_arc_v3.csv};
\end{axis}
        \end{tikzpicture}
        \caption{Group A}\label{fig:rd_mio100_clipiqa_a}
    \end{subfigure}
    \hfill
    \begin{subfigure}[t]{0.32\textwidth}
        \centering
        \begin{tikzpicture}
            \begin{axis}[
                width=\textwidth,
                height=0.8\textwidth,
                xlabel near ticks,
                ylabel near ticks,
                xlabel={BPP},
                ylabel={CLIPIQA},
                grid=major,
                grid style={dashed, gray!30},
                xmin=0, xmax=0.53,
                ymin=0.15, ymax=0.6,
            ]
            \addplot[black, mark=*, thick] table[x=bpp,y=clipiqa,col sep=comma] {data/rd_curves/b_vvc.csv};
            \addplot[blue, mark=*, thick] table[x=bpp,y=clipiqa,col sep=comma] {data/rd_curves/b_dcae.csv};
            \addplot[teal, mark=*, thick] table[x=bpp,y=clipiqa,col sep=comma] {data/rd_curves/b_cod_lite.csv};
            \addplot[orange, mark=*, thick] table[x=bpp,y=clipiqa,col sep=comma] {data/rd_curves/b_allinone.csv};
            \addplot[blue, mark=square*, thick] table[x=bpp,y=clipiqa,col sep=comma] {data/rd_curves/b_arcir_dcae_v3.csv};
            \addplot[teal, mark=square*, thick] table[x=bpp,y=clipiqa,col sep=comma] {data/rd_curves/b_arcir_cod_lite_v3.csv};
            \addplot[blue, mark=triangle*, thick, dashed] table[x=bpp,y=clipiqa,col sep=comma] {data/rd_curves/b_dcae_arcir_v3.csv};
            \addplot[teal, mark=triangle*, thick, dashed] table[x=bpp,y=clipiqa,col sep=comma] {data/rd_curves/b_cod_lite_arcir_v3.csv};
            \addplot[red, mark=pentagon*, thick, mark size=3pt, mark options={fill=red!50!white}] table[x=bpp,y=clipiqa,col sep=comma] {data/rd_curves/b_arc_v3.csv};
\end{axis}
        \end{tikzpicture}
        \caption{Group B}\label{fig:rd_mio100_clipiqa_b}
    \end{subfigure}
    \hfill
    \begin{subfigure}[t]{0.32\textwidth}
        \centering
        \begin{tikzpicture}
            \begin{axis}[
                width=\textwidth,
                height=0.8\textwidth,
                xlabel near ticks,
                ylabel near ticks,
                xlabel={BPP},
                ylabel={CLIPIQA},
                grid=major,
                grid style={dashed, gray!30},
                xmin=0, xmax=0.53,
                ymin=0.15, ymax=0.6,
            ]
            \addplot[black, mark=*, thick] table[x=bpp,y=clipiqa,col sep=comma] {data/rd_curves/c_vvc.csv};
            \addplot[blue, mark=*, thick] table[x=bpp,y=clipiqa,col sep=comma] {data/rd_curves/c_dcae.csv};
            \addplot[teal, mark=*, thick] table[x=bpp,y=clipiqa,col sep=comma] {data/rd_curves/c_cod_lite.csv};
            \addplot[orange, mark=*, thick] table[x=bpp,y=clipiqa,col sep=comma] {data/rd_curves/c_allinone.csv};
            \addplot[blue, mark=square*, thick] table[x=bpp,y=clipiqa,col sep=comma] {data/rd_curves/c_arcir_dcae_v3.csv};
            \addplot[teal, mark=square*, thick] table[x=bpp,y=clipiqa,col sep=comma] {data/rd_curves/c_arcir_cod_lite_v3.csv};
            \addplot[blue, mark=triangle*, thick, dashed] table[x=bpp,y=clipiqa,col sep=comma] {data/rd_curves/c_dcae_arcir_v3.csv};
            \addplot[teal, mark=triangle*, thick, dashed] table[x=bpp,y=clipiqa,col sep=comma] {data/rd_curves/c_cod_lite_arcir_v3.csv};
            \addplot[red, mark=pentagon*, thick, mark size=3pt, mark options={fill=red!50!white}] table[x=bpp,y=clipiqa,col sep=comma] {data/rd_curves/c_arc_v3.csv};
\end{axis}
        \end{tikzpicture}
        \caption{Group C}\label{fig:rd_mio100_clipiqa_c}
    \end{subfigure}

    \centering
    \ref{miorlegend}

    \caption{\textbf{Rate-distortion performance on MiO100} for CLIPIQA $\uparrow$ vs. BPP.}\label{fig:rd_mio100_clipiqa}
\end{figure*}
\begin{figure*}[t]
    \centering
    \footnotesize
    \begin{subfigure}[t]{0.32\textwidth}
        \centering
        \begin{tikzpicture}
            \begin{axis}[
                width=\textwidth,
                height=0.8\textwidth,
                xlabel near ticks,
                ylabel near ticks,
                xlabel={BPP},
                ylabel={PieAPP},
                grid=major,
                grid style={dashed, gray!30},
                xmin=0, xmax=0.53,
                ymin=1.5, ymax=5.0,
            ]
            \addplot[black, mark=*, thick] table[x=bpp,y=pieapp,col sep=comma] {data/rd_curves/a_vvc.csv};
            \addplot[blue, mark=*, thick] table[x=bpp,y=pieapp,col sep=comma] {data/rd_curves/a_dcae.csv};
            \addplot[teal, mark=*, thick] table[x=bpp,y=pieapp,col sep=comma] {data/rd_curves/a_cod_lite.csv};
            \addplot[blue, mark=square*, thick] table[x=bpp,y=pieapp,col sep=comma] {data/rd_curves/a_arcir_dcae_v3.csv};
            \addplot[teal, mark=square*, thick] table[x=bpp,y=pieapp,col sep=comma] {data/rd_curves/a_arcir_cod_lite_v3.csv};
            \addplot[blue, mark=triangle*, thick, dashed] table[x=bpp,y=pieapp,col sep=comma] {data/rd_curves/a_dcae_arcir_v3.csv};
            \addplot[teal, mark=triangle*, thick, dashed] table[x=bpp,y=pieapp,col sep=comma] {data/rd_curves/a_cod_lite_arcir_v3.csv};
            \addplot[red, mark=pentagon*, thick, mark size=3pt, mark options={fill=red!50!white}] table[x=bpp,y=pieapp,col sep=comma] {data/rd_curves/a_arc_v3.csv};
\end{axis}
        \end{tikzpicture}
        \caption{Group A}\label{fig:rd_mio100_pieapp_a}
    \end{subfigure}
    \hfill
    \begin{subfigure}[t]{0.32\textwidth}
        \centering
        \begin{tikzpicture}
            \begin{axis}[
                width=\textwidth,
                height=0.8\textwidth,
                xlabel near ticks,
                ylabel near ticks,
                xlabel={BPP},
                ylabel={PieAPP},
                grid=major,
                grid style={dashed, gray!30},
                xmin=0, xmax=0.53,
                ymin=1.5, ymax=5.0,
            ]
            \addplot[black, mark=*, thick] table[x=bpp,y=pieapp,col sep=comma] {data/rd_curves/b_vvc.csv};
            \addplot[blue, mark=*, thick] table[x=bpp,y=pieapp,col sep=comma] {data/rd_curves/b_dcae.csv};
            \addplot[teal, mark=*, thick] table[x=bpp,y=pieapp,col sep=comma] {data/rd_curves/b_cod_lite.csv};
            \addplot[blue, mark=square*, thick] table[x=bpp,y=pieapp,col sep=comma] {data/rd_curves/b_arcir_dcae_v3.csv};
            \addplot[teal, mark=square*, thick] table[x=bpp,y=pieapp,col sep=comma] {data/rd_curves/b_arcir_cod_lite_v3.csv};
            \addplot[blue, mark=triangle*, thick, dashed] table[x=bpp,y=pieapp,col sep=comma] {data/rd_curves/b_dcae_arcir_v3.csv};
            \addplot[teal, mark=triangle*, thick, dashed] table[x=bpp,y=pieapp,col sep=comma] {data/rd_curves/b_cod_lite_arcir_v3.csv};
            \addplot[red, mark=pentagon*, thick, mark size=3pt, mark options={fill=red!50!white}] table[x=bpp,y=pieapp,col sep=comma] {data/rd_curves/b_arc_v3.csv};
\end{axis}
        \end{tikzpicture}
        \caption{Group B}\label{fig:rd_mio100_pieapp_b}
    \end{subfigure}
    \hfill
    \begin{subfigure}[t]{0.32\textwidth}
        \centering
        \begin{tikzpicture}
            \begin{axis}[
                width=\textwidth,
                height=0.8\textwidth,
                xlabel near ticks,
                ylabel near ticks,
                xlabel={BPP},
                ylabel={PieAPP},
                grid=major,
                grid style={dashed, gray!30},
                xmin=0, xmax=0.53,
                ymin=1.5, ymax=5.0,
            ]
            \addplot[black, mark=*, thick] table[x=bpp,y=pieapp,col sep=comma] {data/rd_curves/c_vvc.csv};
            \addplot[blue, mark=*, thick] table[x=bpp,y=pieapp,col sep=comma] {data/rd_curves/c_dcae.csv};
            \addplot[teal, mark=*, thick] table[x=bpp,y=pieapp,col sep=comma] {data/rd_curves/c_cod_lite.csv};
            \addplot[blue, mark=square*, thick] table[x=bpp,y=pieapp,col sep=comma] {data/rd_curves/c_arcir_dcae_v3.csv};
            \addplot[teal, mark=square*, thick] table[x=bpp,y=pieapp,col sep=comma] {data/rd_curves/c_arcir_cod_lite_v3.csv};
            \addplot[blue, mark=triangle*, thick, dashed] table[x=bpp,y=pieapp,col sep=comma] {data/rd_curves/c_dcae_arcir_v3.csv};
            \addplot[teal, mark=triangle*, thick, dashed] table[x=bpp,y=pieapp,col sep=comma] {data/rd_curves/c_cod_lite_arcir_v3.csv};
            \addplot[red, mark=pentagon*, thick, mark size=3pt, mark options={fill=red!50!white}] table[x=bpp,y=pieapp,col sep=comma] {data/rd_curves/c_arc_v3.csv};
\end{axis}
        \end{tikzpicture}
        \caption{Group C}\label{fig:rd_mio100_pieapp_c}
    \end{subfigure}

    \centering
    \ref{miorlegend}

    \caption{\textbf{Rate-distortion performance on MiO100} for PieAPP $\downarrow$ vs. BPP.}\label{fig:rd_mio100_pieapp}
\end{figure*}
\begin{figure*}[t]
    \centering
    \footnotesize
    \begin{subfigure}[t]{0.32\textwidth}
        \centering
        \begin{tikzpicture}
            \begin{axis}[
                width=\textwidth,
                height=0.8\textwidth,
                xlabel near ticks,
                ylabel near ticks,
                xlabel={BPP},
                ylabel={MUSIQ},
                grid=major,
                grid style={dashed, gray!30},
                xmin=0, xmax=0.53,
                ymin=20, ymax=65,
            ]
            \addplot[black, mark=*, thick] table[x=bpp,y=musiq,col sep=comma] {data/rd_curves/a_vvc.csv};
            \addplot[blue, mark=*, thick] table[x=bpp,y=musiq,col sep=comma] {data/rd_curves/a_dcae.csv};
            \addplot[teal, mark=*, thick] table[x=bpp,y=musiq,col sep=comma] {data/rd_curves/a_cod_lite.csv};
            \addplot[orange, mark=*, thick] table[x=bpp,y=musiq,col sep=comma] {data/rd_curves/a_allinone.csv};
            \addplot[blue, mark=square*, thick] table[x=bpp,y=musiq,col sep=comma] {data/rd_curves/a_arcir_dcae_v3.csv};
            \addplot[teal, mark=square*, thick] table[x=bpp,y=musiq,col sep=comma] {data/rd_curves/a_arcir_cod_lite_v3.csv};
            \addplot[violet, mark=square*, thick] table[x=bpp,y=musiq,col sep=comma] {data/rd_curves/a_airnet_dcae.csv};
            \addplot[blue, mark=triangle*, thick, dashed] table[x=bpp,y=musiq,col sep=comma] {data/rd_curves/a_dcae_arcir_v3.csv};
            \addplot[teal, mark=triangle*, thick, dashed] table[x=bpp,y=musiq,col sep=comma] {data/rd_curves/a_cod_lite_arcir_v3.csv};
            \addplot[violet, mark=triangle*, thick, dashed] table[x=bpp,y=musiq,col sep=comma] {data/rd_curves/a_dcae_airnet.csv};
            \addplot[red, mark=pentagon*, thick, mark size=3pt, mark options={fill=red!50!white}] table[x=bpp,y=musiq,col sep=comma] {data/rd_curves/a_arc_v3.csv};
\end{axis}
        \end{tikzpicture}
        \caption{Group A}\label{fig:rd_mio100_musiq_a}
    \end{subfigure}
    \hfill
    \begin{subfigure}[t]{0.32\textwidth}
        \centering
        \begin{tikzpicture}
            \begin{axis}[
                width=\textwidth,
                height=0.8\textwidth,
                xlabel near ticks,
                ylabel near ticks,
                xlabel={BPP},
                ylabel={MUSIQ},
                grid=major,
                grid style={dashed, gray!30},
                xmin=0, xmax=0.53,
                ymin=20, ymax=65,
            ]
            \addplot[black, mark=*, thick] table[x=bpp,y=musiq,col sep=comma] {data/rd_curves/b_vvc.csv};
            \addplot[blue, mark=*, thick] table[x=bpp,y=musiq,col sep=comma] {data/rd_curves/b_dcae.csv};
            \addplot[teal, mark=*, thick] table[x=bpp,y=musiq,col sep=comma] {data/rd_curves/b_cod_lite.csv};
            \addplot[orange, mark=*, thick] table[x=bpp,y=musiq,col sep=comma] {data/rd_curves/b_allinone.csv};
            \addplot[blue, mark=square*, thick] table[x=bpp,y=musiq,col sep=comma] {data/rd_curves/b_arcir_dcae_v3.csv};
            \addplot[teal, mark=square*, thick] table[x=bpp,y=musiq,col sep=comma] {data/rd_curves/b_arcir_cod_lite_v3.csv};
            \addplot[violet, mark=square*, thick] table[x=bpp,y=musiq,col sep=comma] {data/rd_curves/b_airnet_dcae.csv};
            \addplot[blue, mark=triangle*, thick, dashed] table[x=bpp,y=musiq,col sep=comma] {data/rd_curves/b_dcae_arcir_v3.csv};
            \addplot[teal, mark=triangle*, thick, dashed] table[x=bpp,y=musiq,col sep=comma] {data/rd_curves/b_cod_lite_arcir_v3.csv};
            \addplot[violet, mark=triangle*, thick, dashed] table[x=bpp,y=musiq,col sep=comma] {data/rd_curves/b_dcae_airnet.csv};
            \addplot[red, mark=pentagon*, thick, mark size=3pt, mark options={fill=red!50!white}] table[x=bpp,y=musiq,col sep=comma] {data/rd_curves/b_arc_v3.csv};
\end{axis}
        \end{tikzpicture}
        \caption{Group B}\label{fig:rd_mio100_musiq_b}
    \end{subfigure}
    \hfill
    \begin{subfigure}[t]{0.32\textwidth}
        \centering
        \begin{tikzpicture}
            \begin{axis}[
                width=\textwidth,
                height=0.8\textwidth,
                xlabel near ticks,
                ylabel near ticks,
                xlabel={BPP},
                ylabel={MUSIQ},
                grid=major,
                grid style={dashed, gray!30},
                xmin=0, xmax=0.53,
                ymin=20, ymax=65,
            ]
            \addplot[black, mark=*, thick] table[x=bpp,y=musiq,col sep=comma] {data/rd_curves/c_vvc.csv};
            \addplot[blue, mark=*, thick] table[x=bpp,y=musiq,col sep=comma] {data/rd_curves/c_dcae.csv};
            \addplot[teal, mark=*, thick] table[x=bpp,y=musiq,col sep=comma] {data/rd_curves/c_cod_lite.csv};
            \addplot[orange, mark=*, thick] table[x=bpp,y=musiq,col sep=comma] {data/rd_curves/c_allinone.csv};
            \addplot[blue, mark=square*, thick] table[x=bpp,y=musiq,col sep=comma] {data/rd_curves/c_arcir_dcae_v3.csv};
            \addplot[teal, mark=square*, thick] table[x=bpp,y=musiq,col sep=comma] {data/rd_curves/c_arcir_cod_lite_v3.csv};
            \addplot[violet, mark=square*, thick] table[x=bpp,y=musiq,col sep=comma] {data/rd_curves/c_airnet_dcae.csv};
            \addplot[blue, mark=triangle*, thick, dashed] table[x=bpp,y=musiq,col sep=comma] {data/rd_curves/c_dcae_arcir_v3.csv};
            \addplot[teal, mark=triangle*, thick, dashed] table[x=bpp,y=musiq,col sep=comma] {data/rd_curves/c_cod_lite_arcir_v3.csv};
            \addplot[violet, mark=triangle*, thick, dashed] table[x=bpp,y=musiq,col sep=comma] {data/rd_curves/c_dcae_airnet.csv};
            \addplot[red, mark=pentagon*, thick, mark size=3pt, mark options={fill=red!50!white}] table[x=bpp,y=musiq,col sep=comma] {data/rd_curves/c_arc_v3.csv};
\end{axis}
        \end{tikzpicture}
        \caption{Group C}\label{fig:rd_mio100_musiq_c}
    \end{subfigure}

    \centering
    \ref{miorlegend}

    \caption{\textbf{Rate-distortion performance on MiO100} for MUSIQ $\uparrow$ vs. BPP.}\label{fig:rd_mio100_musiq}
\end{figure*}
\begin{figure*}[t]
    \centering
    \footnotesize
    \begin{subfigure}[t]{0.32\textwidth}
        \centering
        \begin{tikzpicture}
            \begin{axis}[
                width=\textwidth,
                height=0.8\textwidth,
                xlabel near ticks,
                ylabel near ticks,
                xlabel={BPP},
                ylabel={TOPIQ},
                grid=major,
                grid style={dashed, gray!30},
                xmin=0, xmax=0.53,
                ymin=0.15, ymax=0.55,
            ]
            \addplot[black, mark=*, thick] table[x=bpp,y=topiq,col sep=comma] {data/rd_curves/a_vvc.csv};
            \addplot[blue, mark=*, thick] table[x=bpp,y=topiq,col sep=comma] {data/rd_curves/a_dcae.csv};
            \addplot[teal, mark=*, thick] table[x=bpp,y=topiq,col sep=comma] {data/rd_curves/a_cod_lite.csv};
            \addplot[orange, mark=*, thick] table[x=bpp,y=topiq,col sep=comma] {data/rd_curves/a_allinone.csv};
            \addplot[blue, mark=square*, thick] table[x=bpp,y=topiq,col sep=comma] {data/rd_curves/a_arcir_dcae_v3.csv};
            \addplot[teal, mark=square*, thick] table[x=bpp,y=topiq,col sep=comma] {data/rd_curves/a_arcir_cod_lite_v3.csv};
            \addplot[violet, mark=square*, thick] table[x=bpp,y=topiq,col sep=comma] {data/rd_curves/a_airnet_dcae.csv};
            \addplot[blue, mark=triangle*, thick, dashed] table[x=bpp,y=topiq,col sep=comma] {data/rd_curves/a_dcae_arcir_v3.csv};
            \addplot[teal, mark=triangle*, thick, dashed] table[x=bpp,y=topiq,col sep=comma] {data/rd_curves/a_cod_lite_arcir_v3.csv};
            \addplot[violet, mark=triangle*, thick, dashed] table[x=bpp,y=topiq,col sep=comma] {data/rd_curves/a_dcae_airnet.csv};
            \addplot[red, mark=pentagon*, thick, mark size=3pt, mark options={fill=red!50!white}] table[x=bpp,y=topiq,col sep=comma] {data/rd_curves/a_arc_v3.csv};
\end{axis}
        \end{tikzpicture}
        \caption{Group A}\label{fig:rd_mio100_topiq_a}
    \end{subfigure}
    \hfill
    \begin{subfigure}[t]{0.32\textwidth}
        \centering
        \begin{tikzpicture}
            \begin{axis}[
                width=\textwidth,
                height=0.8\textwidth,
                xlabel near ticks,
                ylabel near ticks,
                xlabel={BPP},
                ylabel={TOPIQ},
                grid=major,
                grid style={dashed, gray!30},
                xmin=0, xmax=0.53,
                ymin=0.15, ymax=0.55,
            ]
            \addplot[black, mark=*, thick] table[x=bpp,y=topiq,col sep=comma] {data/rd_curves/b_vvc.csv};
            \addplot[blue, mark=*, thick] table[x=bpp,y=topiq,col sep=comma] {data/rd_curves/b_dcae.csv};
            \addplot[teal, mark=*, thick] table[x=bpp,y=topiq,col sep=comma] {data/rd_curves/b_cod_lite.csv};
            \addplot[orange, mark=*, thick] table[x=bpp,y=topiq,col sep=comma] {data/rd_curves/b_allinone.csv};
            \addplot[blue, mark=square*, thick] table[x=bpp,y=topiq,col sep=comma] {data/rd_curves/b_arcir_dcae_v3.csv};
            \addplot[teal, mark=square*, thick] table[x=bpp,y=topiq,col sep=comma] {data/rd_curves/b_arcir_cod_lite_v3.csv};
            \addplot[violet, mark=square*, thick] table[x=bpp,y=topiq,col sep=comma] {data/rd_curves/b_airnet_dcae.csv};
            \addplot[blue, mark=triangle*, thick, dashed] table[x=bpp,y=topiq,col sep=comma] {data/rd_curves/b_dcae_arcir_v3.csv};
            \addplot[teal, mark=triangle*, thick, dashed] table[x=bpp,y=topiq,col sep=comma] {data/rd_curves/b_cod_lite_arcir_v3.csv};
            \addplot[violet, mark=triangle*, thick, dashed] table[x=bpp,y=topiq,col sep=comma] {data/rd_curves/b_dcae_airnet.csv};
            \addplot[red, mark=pentagon*, thick, mark size=3pt, mark options={fill=red!50!white}] table[x=bpp,y=topiq,col sep=comma] {data/rd_curves/b_arc_v3.csv};
\end{axis}
        \end{tikzpicture}
        \caption{Group B}\label{fig:rd_mio100_topiq_b}
    \end{subfigure}
    \hfill
    \begin{subfigure}[t]{0.32\textwidth}
        \centering
        \begin{tikzpicture}
            \begin{axis}[
                width=\textwidth,
                height=0.8\textwidth,
                xlabel near ticks,
                ylabel near ticks,
                xlabel={BPP},
                ylabel={TOPIQ},
                grid=major,
                grid style={dashed, gray!30},
                xmin=0, xmax=0.53,
                ymin=0.15, ymax=0.55,
            ]
            \addplot[black, mark=*, thick] table[x=bpp,y=topiq,col sep=comma] {data/rd_curves/c_vvc.csv};
            \addplot[blue, mark=*, thick] table[x=bpp,y=topiq,col sep=comma] {data/rd_curves/c_dcae.csv};
            \addplot[teal, mark=*, thick] table[x=bpp,y=topiq,col sep=comma] {data/rd_curves/c_cod_lite.csv};
            \addplot[orange, mark=*, thick] table[x=bpp,y=topiq,col sep=comma] {data/rd_curves/c_allinone.csv};
            \addplot[blue, mark=square*, thick] table[x=bpp,y=topiq,col sep=comma] {data/rd_curves/c_arcir_dcae_v3.csv};
            \addplot[teal, mark=square*, thick] table[x=bpp,y=topiq,col sep=comma] {data/rd_curves/c_arcir_cod_lite_v3.csv};
            \addplot[violet, mark=square*, thick] table[x=bpp,y=topiq,col sep=comma] {data/rd_curves/c_airnet_dcae.csv};
            \addplot[blue, mark=triangle*, thick, dashed] table[x=bpp,y=topiq,col sep=comma] {data/rd_curves/c_dcae_arcir_v3.csv};
            \addplot[teal, mark=triangle*, thick, dashed] table[x=bpp,y=topiq,col sep=comma] {data/rd_curves/c_cod_lite_arcir_v3.csv};
            \addplot[violet, mark=triangle*, thick, dashed] table[x=bpp,y=topiq,col sep=comma] {data/rd_curves/c_dcae_airnet.csv};
            \addplot[red, mark=pentagon*, thick, mark size=3pt, mark options={fill=red!50!white}] table[x=bpp,y=topiq,col sep=comma] {data/rd_curves/c_arc_v3.csv};
\end{axis}
        \end{tikzpicture}
        \caption{Group C}\label{fig:rd_mio100_topiq_c}
    \end{subfigure}

    \centering
    \ref{miorlegend}

    \caption{\textbf{Rate-distortion performance on MiO100} for TOPIQ $\uparrow$ vs. BPP.}\label{fig:rd_mio100_topiq}
\end{figure*}

\subsection{Measured Operating Points for MiO100}
Tables~\ref{tab:mio100_bpp01} and \ref{tab:mio100_bpp02} list the measured MiO100 operating points from which Table~\ref{tab:mio100_bpp015_interp} is interpolated: for each method, the closest point at or below and at or above 0.15\,bpp.
These also contain additional results for combining DCAE with a traditional restoration model (AirNet~\cite{liAllInOneImageRestoration2022}), and a joint compression and restoration method (AllInOne~\cite{zengAllinOneImageCompression2025}).
Both are single-pass ``all-in-one'' networks trained to handle individual degradation types rather than the compound combinations MiO100 pairs together.
As a consequence, AllInOne and the AirNet-based combinations (AirNet+DCAE, DCAE+AirNet) trail ARC in Tables~\ref{tab:mio100_bpp01} and \ref{tab:mio100_bpp02}.

\subsection{Codec Selection by Degradation}
Complementing the discussion of agent behavior in Section~\ref{sec:experiments}, Table~\ref{tab:codec_by_degradation} gives the full per-degradation codec-selection breakdown at all three rate parameters. DCAE is selected most often overall, especially for blur combined with JPEG artifacts; CoD-Lite gains share as $\lambda$ increases (lower rates); VVC is chosen mainly for degradations that include low resolution, and mostly at the smallest $\lambda$.

\begin{table*}[t]
    \centering
    \footnotesize
    \setlength{\tabcolsep}{3pt}
    \begin{tabular}{llrrrrrrrrrr}
        \toprule
         & & & \multicolumn{3}{c}{$\lambda=0.005$} & \multicolumn{3}{c}{$\lambda=0.01$} & \multicolumn{3}{c}{$\lambda=0.02$} \\
        \cmidrule(lr){4-6} \cmidrule(lr){7-9} \cmidrule(lr){10-12}
        Group & Degradation & $n$ & CoD-Lite & DCAE & VVC & CoD-Lite & DCAE & VVC & CoD-Lite & DCAE & VVC \\
        \midrule
        \multirow{8}{*}{A}
        & dark+noise & 80 & 52\% & 46\% & 1\% & 60\% & 39\% & 1\% & 62\% & 36\% & 1\% \\
        & defocus blur+haze & 80 & 52\% & 48\% & -- & 59\% & 41\% & -- & 70\% & 30\% & -- \\
        & defocus blur+jpeg compression artifact & 80 & 9\% & 90\% & 1\% & 16\% & 84\% & -- & 30\% & 70\% & -- \\
        & motion blur+dark & 80 & 35\% & 65\% & -- & 56\% & 44\% & -- & 75\% & 25\% & -- \\
        & motion blur+low resolution & 80 & 5\% & 50\% & 45\% & 8\% & 68\% & 25\% & 14\% & 81\% & 5\% \\
        & noise+jpeg compression artifact & 80 & 38\% & 59\% & 4\% & 41\% & 59\% & -- & 59\% & 41\% & -- \\
        & rain+haze & 80 & 51\% & 48\% & 1\% & 68\% & 32\% & -- & 84\% & 16\% & -- \\
        & rain+low resolution & 80 & 1\% & 51\% & 48\% & 5\% & 71\% & 24\% & 6\% & 82\% & 11\% \\
        \midrule
        \multirow{4}{*}{B}
        & defocus blur+low resolution & 100 & 4\% & 54\% & 42\% & 7\% & 67\% & 26\% & 9\% & 78\% & 13\% \\
        & haze+noise & 100 & 48\% & 51\% & 1\% & 49\% & 50\% & 1\% & 56\% & 44\% & -- \\
        & motion blur+jpeg compression artifact & 100 & 7\% & 92\% & 1\% & 12\% & 88\% & -- & 41\% & 58\% & 1\% \\
        & rain+dark & 100 & 33\% & 65\% & 2\% & 55\% & 44\% & 1\% & 67\% & 33\% & -- \\
        \midrule
        \multirow{4}{*}{C}
        & dark+defocus blur+jpeg compression artifact & 100 & 18\% & 82\% & -- & 20\% & 79\% & 1\% & 32\% & 68\% & -- \\
        & haze+motion blur+low resolution & 100 & 9\% & 23\% & 68\% & 16\% & 39\% & 45\% & 29\% & 53\% & 18\% \\
        & motion blur+defocus blur+noise & 100 & 67\% & 33\% & -- & 68\% & 32\% & -- & 65\% & 35\% & -- \\
        & rain+noise+low resolution & 100 & 4\% & 39\% & 57\% & 5\% & 59\% & 36\% & 7\% & 72\% & 21\% \\
        \midrule
        \multicolumn{2}{l}{All} & 1440 & 27\% & 56\% & 17\% & 33\% & 56\% & 10\% & 43\% & 52\% & 5\% \\
        \bottomrule
    \end{tabular}
    \caption{\textbf{Codec-selection frequency of ARC} by MiO100 degradation at each rate parameter $\lambda$ (larger $\lambda$ gives lower rates). $n$ is the number of evaluated views per degradation; percentages are rounded per row and may not sum to 100.}\label{tab:codec_by_degradation}
\end{table*}

\section{Extended Kodak Results}

\begin{table*}[th!]
    \centering
    \footnotesize
    \setlength{\tabcolsep}{2.8pt}
    \begin{tabular}{l*{4}{>{\columncolor{frbg}}c}*{5}{>{\columncolor{nrbg}}c}cc}
        \toprule
        Method & PSNR $\uparrow$ & MS-SSIM $\uparrow$ & LPIPS $\downarrow$ & PieAPP $\downarrow$ & MANIQA $\uparrow$ & CLIPIQA $\uparrow$ & MUSIQ $\uparrow$ & TOPIQ $\uparrow$ & DBCNN $\uparrow$ & BPP \textdownarrow & Avg. rank $\downarrow$ \\
        \midrule
        VVC~\cite{brossOverviewVersatileVideo2021} & \textbf{32.86}\textsuperscript{1} & \textbf{0.962}\textsuperscript{1} & 0.233\textsuperscript{5} & \underline{1.008}\textsuperscript{2} & 0.664\textsuperscript{6} & 0.413\textsuperscript{8} & 71.68\textsuperscript{8} & 0.566\textsuperscript{8} & 0.591\textsuperscript{8} & 0.199 & 5.22 \\
        DCAE~\cite{luLearnedImageCompression2025} & \underline{31.60}\textsuperscript{2} & \underline{0.950}\textsuperscript{2} & 0.274\textsuperscript{7} & 1.260\textsuperscript{7} & 0.642\textsuperscript{7} & 0.517\textsuperscript{7} & 72.70\textsuperscript{6} & 0.606\textsuperscript{7} & 0.593\textsuperscript{7} & 0.128 & 5.78 \\
        CoD-Lite~\cite{jiaCoDLiteRealTimeDiffusionBased2026} & 26.44\textsuperscript{4} & 0.909\textsuperscript{5} & \textbf{0.097}\textsuperscript{1} & \textbf{1.005}\textsuperscript{1} & \underline{0.675}\textsuperscript{2.5} & \underline{0.757}\textsuperscript{2} & 73.58\textsuperscript{5} & 0.650\textsuperscript{4} & 0.660\textsuperscript{4} & 0.125 & \textbf{3.17} \\
        \ARCIR{} + DCAE & \textit{29.50}\textsuperscript{3} & \textit{0.941}\textsuperscript{3} & 0.255\textsuperscript{6} & 1.174\textsuperscript{6} & 0.669\textsuperscript{5} & 0.556\textsuperscript{5} & \textit{73.89}\textsuperscript{3} & 0.625\textsuperscript{5} & 0.608\textsuperscript{6} & 0.170 & 4.67 \\
        \ARCIR{} + CoD-Lite & 24.61\textsuperscript{7} & 0.894\textsuperscript{7} & \underline{0.118}\textsuperscript{2} & 1.080\textsuperscript{5} & \underline{0.675}\textsuperscript{2.5} & \textbf{0.761}\textsuperscript{1} & 73.79\textsuperscript{4} & \textit{0.651}\textsuperscript{3} & \underline{0.663}\textsuperscript{2} & 0.125 & \underline{3.72} \\
        DCAE + \ARCIR{} & 26.07\textsuperscript{5} & 0.911\textsuperscript{4} & 0.301\textsuperscript{8} & 1.274\textsuperscript{8} & 0.633\textsuperscript{8} & 0.539\textsuperscript{6} & 72.64\textsuperscript{7} & 0.623\textsuperscript{6} & 0.628\textsuperscript{5} & 0.128 & 6.33 \\
        CoD-Lite + \ARCIR{} & 23.64\textsuperscript{8} & 0.877\textsuperscript{8} & 0.138\textsuperscript{4} & 1.078\textsuperscript{4} & \textit{0.670}\textsuperscript{4} & \textit{0.737}\textsuperscript{3.5} & \underline{73.93}\textsuperscript{2} & \textbf{0.658}\textsuperscript{1} & \textbf{0.675}\textsuperscript{1} & 0.125 & \textit{3.94} \\
        \midrule
        ARC & 25.34\textsuperscript{6} & 0.901\textsuperscript{6} & \textit{0.121}\textsuperscript{3} & \textit{1.039}\textsuperscript{3} & \textbf{0.679}\textsuperscript{1} & \textit{0.737}\textsuperscript{3.5} & \textbf{74.18}\textsuperscript{1} & \underline{0.655}\textsuperscript{2} & \textit{0.662}\textsuperscript{3} & 0.148 & \textbf{3.17} \\
        \bottomrule
    \end{tabular}
    \caption{\textbf{Results on 24 clean Kodak images at the lower interpolation points, with actual BPP.} Each method shows its measured operating point closest to 0.2\,bpp at or below it, i.e., the lower input of the interpolation in Table~\ref{tab:kodak}.}\label{tab:kodak_measured_l002}
\end{table*}

\begin{table*}[th!]
    \centering
    \footnotesize
    \setlength{\tabcolsep}{2.8pt}
    \begin{tabular}{l*{4}{>{\columncolor{frbg}}c}*{5}{>{\columncolor{nrbg}}c}cc}
        \toprule
        Method & PSNR $\uparrow$ & MS-SSIM $\uparrow$ & LPIPS $\downarrow$ & PieAPP $\downarrow$ & MANIQA $\uparrow$ & CLIPIQA $\uparrow$ & MUSIQ $\uparrow$ & TOPIQ $\uparrow$ & DBCNN $\uparrow$ & BPP \textdownarrow & Avg. rank $\downarrow$ \\
        \midrule
        VVC~\cite{brossOverviewVersatileVideo2021} & \textbf{35.77}\textsuperscript{1} & \textbf{0.982}\textsuperscript{1} & 0.142\textsuperscript{5} & \underline{0.642}\textsuperscript{2} & \textbf{0.714}\textsuperscript{1} & 0.535\textsuperscript{8} & 74.18\textsuperscript{5} & 0.637\textsuperscript{7} & 0.654\textsuperscript{5} & 0.402 & 3.89 \\
        DCAE~\cite{luLearnedImageCompression2025} & \underline{33.44}\textsuperscript{2} & \textit{0.964}\textsuperscript{3} & 0.213\textsuperscript{7} & 1.003\textsuperscript{7} & 0.682\textsuperscript{7} & 0.569\textsuperscript{7} & 73.95\textsuperscript{8} & 0.636\textsuperscript{8} & 0.620\textsuperscript{8} & 0.203 & 6.33 \\
        CoD-Lite~\cite{jiaCoDLiteRealTimeDiffusionBased2026} & \textit{30.90}\textsuperscript{3} & \underline{0.975}\textsuperscript{2} & \textbf{0.048}\textsuperscript{1} & \textbf{0.553}\textsuperscript{1} & \underline{0.709}\textsuperscript{2} & \underline{0.752}\textsuperscript{2} & 74.12\textsuperscript{6} & \textit{0.658}\textsuperscript{3.5} & \textit{0.663}\textsuperscript{3} & 0.500 & \textbf{2.61} \\
        \ARCIR{} + DCAE & 30.81\textsuperscript{4} & 0.954\textsuperscript{5} & 0.198\textsuperscript{6} & 0.987\textsuperscript{6} & 0.702\textsuperscript{5} & 0.600\textsuperscript{5} & \textbf{74.79}\textsuperscript{1} & 0.648\textsuperscript{5.5} & 0.632\textsuperscript{7} & 0.265 & 4.94 \\
        \ARCIR{} + CoD-Lite & 28.24\textsuperscript{6} & 0.956\textsuperscript{4} & \underline{0.073}\textsuperscript{2} & 0.688\textsuperscript{4} & \textit{0.708}\textsuperscript{3} & \textbf{0.756}\textsuperscript{1} & 74.37\textsuperscript{4} & \textit{0.658}\textsuperscript{3.5} & \underline{0.667}\textsuperscript{2} & 0.500 & \underline{3.28} \\
        DCAE + \ARCIR{} & 26.06\textsuperscript{8} & 0.911\textsuperscript{8} & 0.253\textsuperscript{8} & 1.148\textsuperscript{8} & 0.663\textsuperscript{8} & 0.574\textsuperscript{6} & 74.03\textsuperscript{7} & 0.648\textsuperscript{5.5} & 0.648\textsuperscript{6} & 0.203 & 7.17 \\
        CoD-Lite + \ARCIR{} & 28.15\textsuperscript{7} & 0.949\textsuperscript{6} & \textit{0.081}\textsuperscript{3} & \textit{0.666}\textsuperscript{3} & 0.704\textsuperscript{4} & \textit{0.746}\textsuperscript{3} & \textit{74.40}\textsuperscript{3} & \textbf{0.665}\textsuperscript{1} & \textbf{0.674}\textsuperscript{1} & 0.500 & \textit{3.44} \\
        \midrule
        ARC & 28.93\textsuperscript{5} & 0.931\textsuperscript{7} & 0.121\textsuperscript{4} & 0.805\textsuperscript{5} & 0.698\textsuperscript{6} & 0.702\textsuperscript{4} & \underline{74.75}\textsuperscript{2} & \underline{0.662}\textsuperscript{2} & 0.660\textsuperscript{4} & 0.268 & 4.33 \\
        \bottomrule
    \end{tabular}
    \caption{\textbf{Results on 24 clean Kodak images at the upper interpolation points, with actual BPP.} Each method shows its measured operating point closest to 0.2\,bpp at or above it, i.e., the upper input of the interpolation in Table~\ref{tab:kodak}.}\label{tab:kodak_measured}
\end{table*}

Complementing Table~\ref{tab:kodak}, Figure~\ref{fig:rd_kodak} sweeps the same
Kodak methods across their available bpp range. Table~\ref{tab:kodak} interpolates each method at 0.2\,bpp between its closest measured points below and above that rate, which Tables~\ref{tab:kodak_measured_l002} and~\ref{tab:kodak_measured} list.
Because the different qualities available for each method can be quite coarse, the rows usually lie at different rates, so ranks within these two tables are not matched-rate comparisons.

\begin{figure*}[t]
    \centering
    \footnotesize
    \begin{subfigure}[t]{0.32\textwidth}
        \centering
        \begin{tikzpicture}
            \begin{axis}[
                width=\textwidth,
                height=\textwidth,
                xlabel near ticks,
                ylabel near ticks,
                xlabel={BPP},
                ylabel={PSNR (dB)},
                grid=major,
                grid style={dashed, gray!30},
                xmin=0, xmax=0.56,
                ymin=22, ymax=37,
            ]
            \addplot[black, mark=*, thick] table[x=bpp,y=psnr,col sep=comma] {data/rd_curves/kodak_vvc.csv};
            \addplot[blue, mark=*, thick] table[x=bpp,y=psnr,col sep=comma] {data/rd_curves/kodak_dcae.csv};
            \addplot[teal, mark=*, thick] table[x=bpp,y=psnr,col sep=comma] {data/rd_curves/kodak_cod_lite.csv};
            \addplot[blue, mark=square*, thick] table[x=bpp,y=psnr,col sep=comma] {data/rd_curves/kodak_arcir_dcae_v3.csv};
            \addplot[teal, mark=square*, thick] table[x=bpp,y=psnr,col sep=comma] {data/rd_curves/kodak_arcir_cod_lite_v3.csv};
            \addplot[blue, mark=triangle*, thick, dashed] table[x=bpp,y=psnr,col sep=comma] {data/rd_curves/kodak_dcae_arcir_v3.csv};
            \addplot[teal, mark=triangle*, thick, dashed] table[x=bpp,y=psnr,col sep=comma] {data/rd_curves/kodak_cod_lite_arcir_v3.csv};
            \addplot[red, mark=pentagon*, thick, mark size=3pt, mark options={fill=red!50!white}] table[x=bpp,y=psnr,col sep=comma] {data/rd_curves/kodak_arc_v3.csv};
\end{axis}
        \end{tikzpicture}
        \caption{PSNR $\uparrow$}\label{fig:rd_kodak_psnr}
    \end{subfigure}
    \hfill
    \begin{subfigure}[t]{0.32\textwidth}
        \centering
        \begin{tikzpicture}
            \begin{axis}[
                width=\textwidth,
                height=\textwidth,
                xlabel near ticks,
                ylabel near ticks,
                xlabel={BPP},
                ylabel={LPIPS},
                legend to name=kodakrdlegend,
                legend columns=4,
                legend style={font=\footnotesize, legend cell align=left},
                grid=major,
                grid style={dashed, gray!30},
                xmin=0, xmax=0.56,
                ymin=0.04, ymax=0.46,
            ]
            \addlegendimage{red, mark=pentagon*, thick, mark size=3pt, mark options={fill=red!50!white}}
            \addlegendentry{ARC}
            \addplot[black, mark=*, thick] table[x=bpp,y=lpips,col sep=comma] {data/rd_curves/kodak_vvc.csv};
            \addlegendentry{VVC~\cite{brossOverviewVersatileVideo2021}}
            \addplot[blue, mark=*, thick] table[x=bpp,y=lpips,col sep=comma] {data/rd_curves/kodak_dcae.csv};
            \addlegendentry{DCAE~\cite{luLearnedImageCompression2025}}
            \addplot[teal, mark=*, thick] table[x=bpp,y=lpips,col sep=comma] {data/rd_curves/kodak_cod_lite.csv};
            \addlegendentry{CoD-Lite~\cite{jiaCoDLiteRealTimeDiffusionBased2026}}
            \addplot[blue, mark=square*, thick] table[x=bpp,y=lpips,col sep=comma] {data/rd_curves/kodak_arcir_dcae_v3.csv};
            \addlegendentry{\ARCIR{} + DCAE}
            \addplot[teal, mark=square*, thick] table[x=bpp,y=lpips,col sep=comma] {data/rd_curves/kodak_arcir_cod_lite_v3.csv};
            \addlegendentry{\ARCIR{} + CoD-Lite}
            \addplot[blue, mark=triangle*, thick, dashed] table[x=bpp,y=lpips,col sep=comma] {data/rd_curves/kodak_dcae_arcir_v3.csv};
            \addlegendentry{DCAE + \ARCIR{}}
            \addplot[teal, mark=triangle*, thick, dashed] table[x=bpp,y=lpips,col sep=comma] {data/rd_curves/kodak_cod_lite_arcir_v3.csv};
            \addlegendentry{CoD-Lite + \ARCIR{}}
            \addplot[red, mark=pentagon*, thick, mark size=3pt, mark options={fill=red!50!white}, forget plot] table[x=bpp,y=lpips,col sep=comma] {data/rd_curves/kodak_arc_v3.csv};
\end{axis}
        \end{tikzpicture}
        \caption{LPIPS $\downarrow$}\label{fig:rd_kodak_lpips}
    \end{subfigure}
    \hfill
    \begin{subfigure}[t]{0.32\textwidth}
        \centering
        \begin{tikzpicture}
            \begin{axis}[
                width=\textwidth,
                height=\textwidth,
                xlabel near ticks,
                ylabel near ticks,
                xlabel={BPP},
                ylabel={DBCNN},
                grid=major,
                grid style={dashed, gray!30},
                xmin=0, xmax=0.56,
                ymin=0.35, ymax=0.7,
            ]
            \addplot[black, mark=*, thick] table[x=bpp,y=dbcnn,col sep=comma] {data/rd_curves/kodak_vvc.csv};
            \addplot[blue, mark=*, thick] table[x=bpp,y=dbcnn,col sep=comma] {data/rd_curves/kodak_dcae.csv};
            \addplot[teal, mark=*, thick] table[x=bpp,y=dbcnn,col sep=comma] {data/rd_curves/kodak_cod_lite.csv};
            \addplot[blue, mark=square*, thick] table[x=bpp,y=dbcnn,col sep=comma] {data/rd_curves/kodak_arcir_dcae_v3.csv};
            \addplot[teal, mark=square*, thick] table[x=bpp,y=dbcnn,col sep=comma] {data/rd_curves/kodak_arcir_cod_lite_v3.csv};
            \addplot[blue, mark=triangle*, thick, dashed] table[x=bpp,y=dbcnn,col sep=comma] {data/rd_curves/kodak_dcae_arcir_v3.csv};
            \addplot[teal, mark=triangle*, thick, dashed] table[x=bpp,y=dbcnn,col sep=comma] {data/rd_curves/kodak_cod_lite_arcir_v3.csv};
            \addplot[red, mark=pentagon*, thick, mark size=3pt, mark options={fill=red!50!white}] table[x=bpp,y=dbcnn,col sep=comma] {data/rd_curves/kodak_arc_v3.csv};
\end{axis}
        \end{tikzpicture}
        \caption{DBCNN $\uparrow$}\label{fig:rd_kodak_dbcnn}
    \end{subfigure}

    \centering
    \ref{kodakrdlegend}

    \caption{\textbf{Rate-distortion performance on 24 clean Kodak images} for PSNR $\uparrow$, LPIPS $\downarrow$, and DBCNN $\uparrow$ vs. BPP.}\label{fig:rd_kodak}
\end{figure*}

\section{Qualitative Comparisons}\label{sec:supp_qualitative}

\begin{figure*}[t]
\centering
\begingroup
\setlength{\tabcolsep}{1.5pt}
\footnotesize
\begin{tabular}{@{}*{5}{c}@{}}
\multicolumn{5}{@{}l}{\rule{0pt}{2.5ex}\textbf{B2/013: haze + noise}\quad Plan: Denoise (SwinIR) $\to$ \textbf{Compress (CoD-Lite)} $\to$ Dehaze (X-Restormer)}\\[3pt]
\begin{minipage}[t]{0.190\textwidth}\centering
\parbox[c][\linewidth][c]{\linewidth}{\centering\includegraphics[width=\linewidth,height=\linewidth,keepaspectratio]{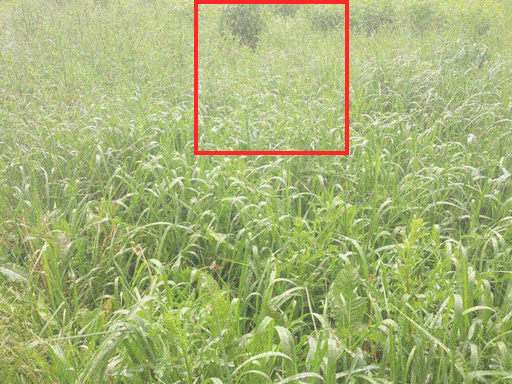}}\par\smallskip
Input\par Full frame\par \strut
\end{minipage}
 &
\begin{minipage}[t]{0.190\textwidth}\centering
\parbox[c][\linewidth][c]{\linewidth}{\centering\includegraphics[width=\linewidth,height=\linewidth,keepaspectratio]{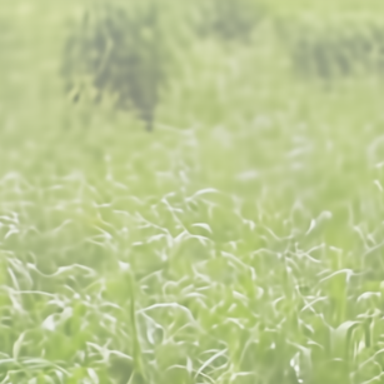}}\par\smallskip
DCAE only\par 0.184 bpp, 17.29\,dB\par LPIPS 0.460
\end{minipage}
 &
\begin{minipage}[t]{0.190\textwidth}\centering
\parbox[c][\linewidth][c]{\linewidth}{\centering\includegraphics[width=\linewidth,height=\linewidth,keepaspectratio]{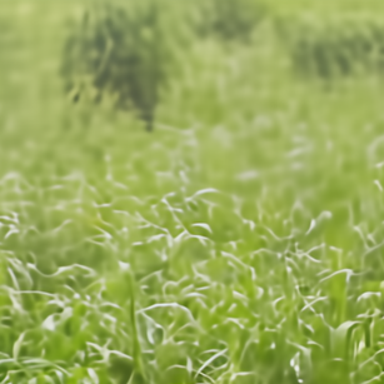}}\par\smallskip
DCAE $\to$ \ARCIR{}\par 0.184 bpp, \textbf{22.10\,dB}\par LPIPS 0.394
\end{minipage}
 &
\begin{minipage}[t]{0.190\textwidth}\centering
\parbox[c][\linewidth][c]{\linewidth}{\centering\includegraphics[width=\linewidth,height=\linewidth,keepaspectratio]{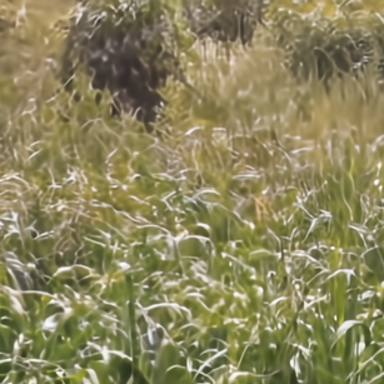}}\par\smallskip
\ARCIR{} $\to$ DCAE\par 0.324 bpp, 19.52\,dB\par LPIPS 0.363
\end{minipage}
 &
\begin{minipage}[t]{0.190\textwidth}\centering
\parbox[c][\linewidth][c]{\linewidth}{\centering\includegraphics[width=\linewidth,height=\linewidth,keepaspectratio]{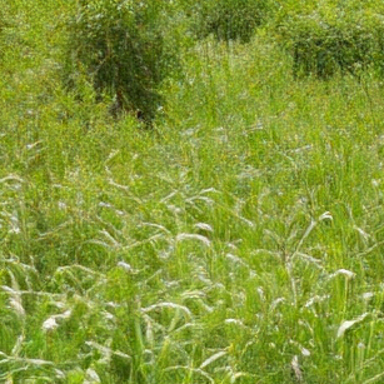}}\par\smallskip
\textbf{ARC (ours)}\par \textbf{0.125 bpp}, 19.14\,dB\par \textbf{LPIPS 0.215}
\end{minipage}\\[3pt]
\begin{minipage}[t]{0.190\textwidth}\centering
\parbox[c][\linewidth][c]{\linewidth}{\centering\includegraphics[width=\linewidth,height=\linewidth,keepaspectratio]{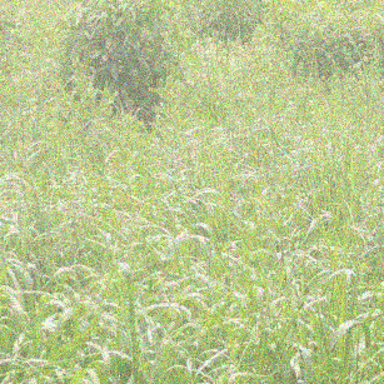}}\par\smallskip
Input crop\par Degraded\par \strut
\end{minipage}
 &
\begin{minipage}[t]{0.190\textwidth}\centering
\parbox[c][\linewidth][c]{\linewidth}{\centering\includegraphics[width=\linewidth,height=\linewidth,keepaspectratio]{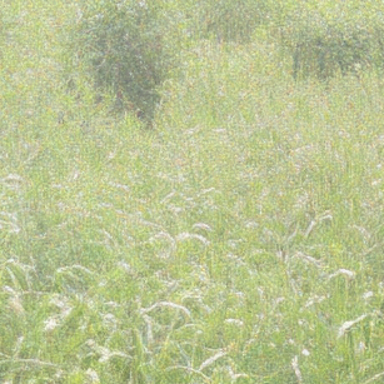}}\par\smallskip
CoD-Lite only\par \textbf{0.125 bpp}, 16.68\,dB\par LPIPS 0.439
\end{minipage}
 &
\begin{minipage}[t]{0.190\textwidth}\centering
\parbox[c][\linewidth][c]{\linewidth}{\centering\includegraphics[width=\linewidth,height=\linewidth,keepaspectratio]{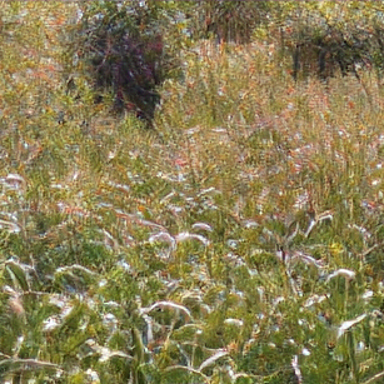}}\par\smallskip
CoD-Lite $\to$ \ARCIR{}\par \textbf{0.125 bpp}, 16.46\,dB\par LPIPS 0.371
\end{minipage}
 &
\begin{minipage}[t]{0.190\textwidth}\centering
\parbox[c][\linewidth][c]{\linewidth}{\centering\includegraphics[width=\linewidth,height=\linewidth,keepaspectratio]{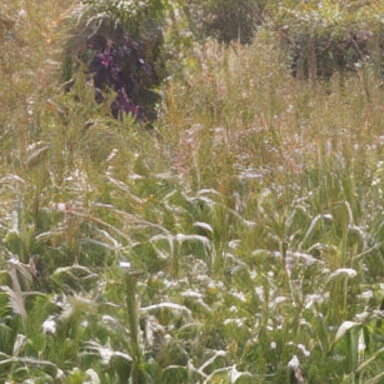}}\par\smallskip
\ARCIR{} $\to$ CoD-Lite\par \textbf{0.125 bpp}, 17.92\,dB\par LPIPS 0.295
\end{minipage}
 &
\begin{minipage}[t]{0.190\textwidth}\centering
\parbox[c][\linewidth][c]{\linewidth}{\centering\includegraphics[width=\linewidth,height=\linewidth,keepaspectratio]{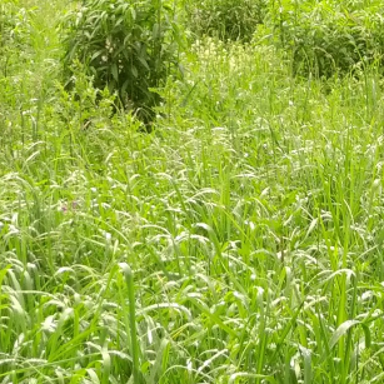}}\par\smallskip
Ground truth\par Clean reference\par \strut
\end{minipage}\\[3pt]
\multicolumn{5}{@{}l}{\rule{0pt}{4ex}\textbf{C2/068: haze + motion blur + low resolution}\quad Plan: Motion deblur (MPRNet) $\to$ \textbf{Compress (DCAE)} $\to$ Dehaze (DehazeFormer) $\to$ SR (DiffBIR)}\\[3pt]
\begin{minipage}[t]{0.190\textwidth}\centering
\parbox[c][\linewidth][c]{\linewidth}{\centering\includegraphics[width=\linewidth,height=\linewidth,keepaspectratio]{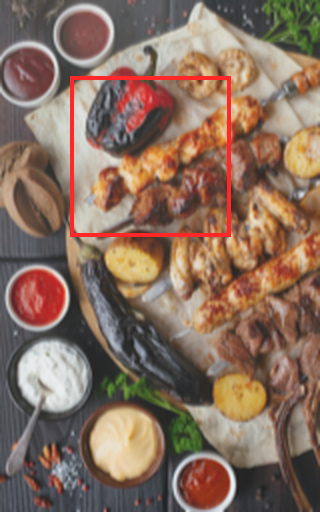}}\par\smallskip
Input\par Full frame\par \strut
\end{minipage}
 &
\begin{minipage}[t]{0.190\textwidth}\centering
\parbox[c][\linewidth][c]{\linewidth}{\centering\includegraphics[width=\linewidth,height=\linewidth,keepaspectratio]{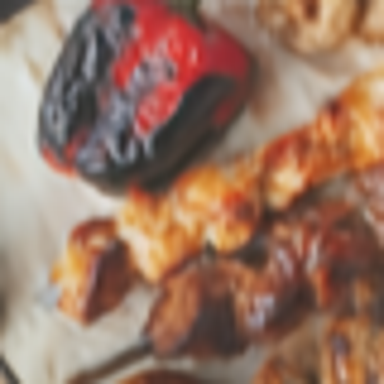}}\par\smallskip
DCAE only\par 0.085 bpp, 18.59\,dB\par LPIPS 0.561
\end{minipage}
 &
\begin{minipage}[t]{0.190\textwidth}\centering
\parbox[c][\linewidth][c]{\linewidth}{\centering\includegraphics[width=\linewidth,height=\linewidth,keepaspectratio]{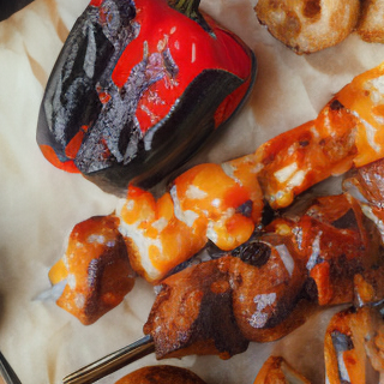}}\par\smallskip
DCAE $\to$ \ARCIR{}\par 0.085 bpp, 19.35\,dB\par LPIPS 0.298
\end{minipage}
 &
\begin{minipage}[t]{0.190\textwidth}\centering
\parbox[c][\linewidth][c]{\linewidth}{\centering\includegraphics[width=\linewidth,height=\linewidth,keepaspectratio]{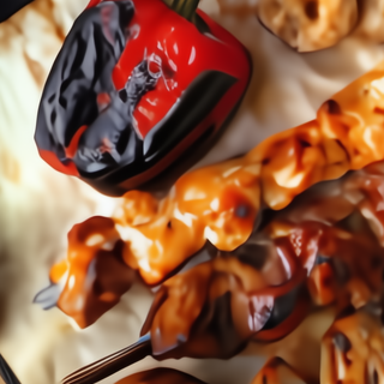}}\par\smallskip
\ARCIR{} $\to$ DCAE\par 0.132 bpp, \textbf{20.92\,dB}\par LPIPS 0.393
\end{minipage}
 &
\begin{minipage}[t]{0.190\textwidth}\centering
\parbox[c][\linewidth][c]{\linewidth}{\centering\includegraphics[width=\linewidth,height=\linewidth,keepaspectratio]{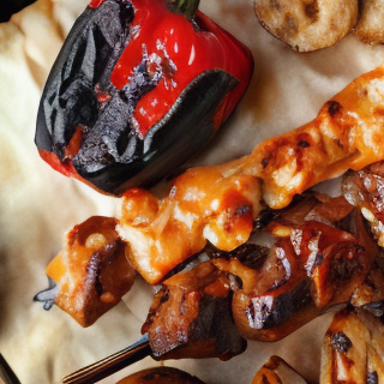}}\par\smallskip
\textbf{ARC (ours)}\par 0.095 bpp, 20.74\,dB\par \textbf{LPIPS 0.253}
\end{minipage}\\[3pt]
\begin{minipage}[t]{0.190\textwidth}\centering
\parbox[c][\linewidth][c]{\linewidth}{\centering\includegraphics[width=\linewidth,height=\linewidth,keepaspectratio]{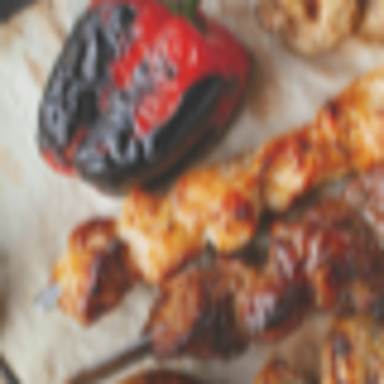}}\par\smallskip
Input crop\par Degraded\par \strut
\end{minipage}
 &
\begin{minipage}[t]{0.190\textwidth}\centering
\parbox[c][\linewidth][c]{\linewidth}{\centering\includegraphics[width=\linewidth,height=\linewidth,keepaspectratio]{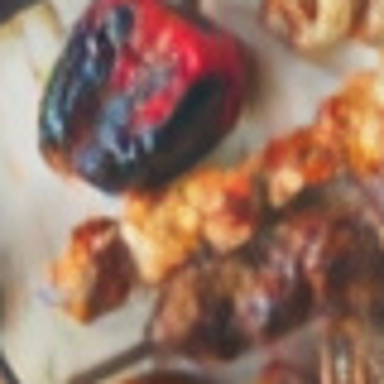}}\par\smallskip
CoD-Lite only\par \textbf{0.029 bpp}, 14.07\,dB\par LPIPS 0.760
\end{minipage}
 &
\begin{minipage}[t]{0.190\textwidth}\centering
\parbox[c][\linewidth][c]{\linewidth}{\centering\includegraphics[width=\linewidth,height=\linewidth,keepaspectratio]{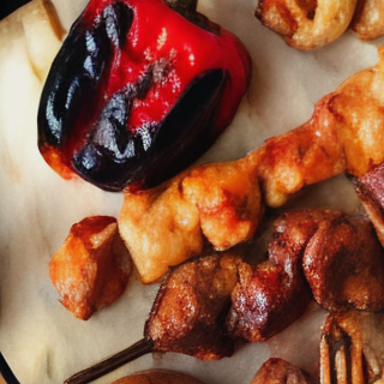}}\par\smallskip
CoD-Lite $\to$ \ARCIR{}\par 0.031 bpp, 18.42\,dB\par LPIPS 0.321
\end{minipage}
 &
\begin{minipage}[t]{0.190\textwidth}\centering
\parbox[c][\linewidth][c]{\linewidth}{\centering\includegraphics[width=\linewidth,height=\linewidth,keepaspectratio]{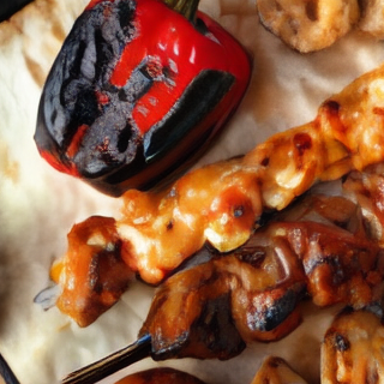}}\par\smallskip
\ARCIR{} $\to$ CoD-Lite\par 0.125 bpp, 20.66\,dB\par LPIPS 0.255
\end{minipage}
 &
\begin{minipage}[t]{0.190\textwidth}\centering
\parbox[c][\linewidth][c]{\linewidth}{\centering\includegraphics[width=\linewidth,height=\linewidth,keepaspectratio]{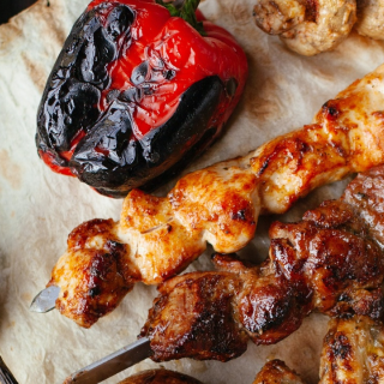}}\par\smallskip
Ground truth\par Clean reference\par \strut
\end{minipage}\\[3pt]
\end{tabular}
\endgroup
\caption{\textbf{Qualitative comparisons on MiO100.} \textbf{Bold}: lowest bpp, highest PSNR and lowest LPIPS per example. Each example shows the DCAE baselines in the first row and the CoD-Lite baselines in the second, with ARC directly above the ground truth. The row header gives the plan ARC executed, including the codec selected for the compression step. Low-resolution input crops are upsampled bicubically for display.}
\label{fig:qualitative_comparison_v3_grid}
\end{figure*}

\begin{figure*}[t]
\centering
\begingroup
\setlength{\tabcolsep}{1.5pt}
\footnotesize
\begin{tabular}{@{}*{5}{c}@{}}
\multicolumn{5}{@{}l}{\rule{0pt}{2.5ex}\textbf{B1/063: defocus blur + low resolution}\quad Plan: \textbf{Compress (DCAE)} $\to$ Defocus deblur (Restormer) $\to$ SR (PiSA-SR)}\\[3pt]
\begin{minipage}[t]{0.190\textwidth}\centering
\parbox[c][\linewidth][c]{\linewidth}{\centering\includegraphics[width=\linewidth,height=\linewidth,keepaspectratio]{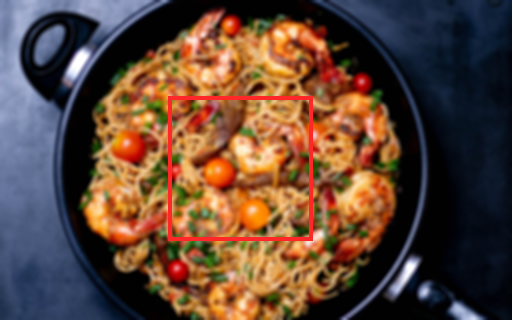}}\par\smallskip
Input\par Full frame\par \strut
\end{minipage}
 &
\begin{minipage}[t]{0.190\textwidth}\centering
\parbox[c][\linewidth][c]{\linewidth}{\centering\includegraphics[width=\linewidth,height=\linewidth,keepaspectratio]{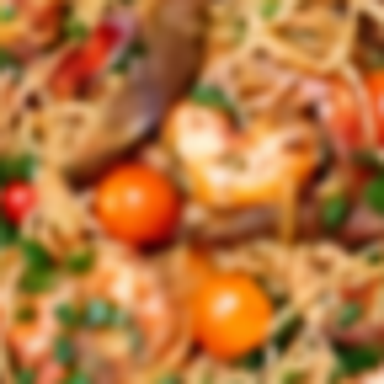}}\par\smallskip
DCAE only\par 0.053 bpp, \textbf{22.08\,dB}\par LPIPS 0.536
\end{minipage}
 &
\begin{minipage}[t]{0.190\textwidth}\centering
\parbox[c][\linewidth][c]{\linewidth}{\centering\includegraphics[width=\linewidth,height=\linewidth,keepaspectratio]{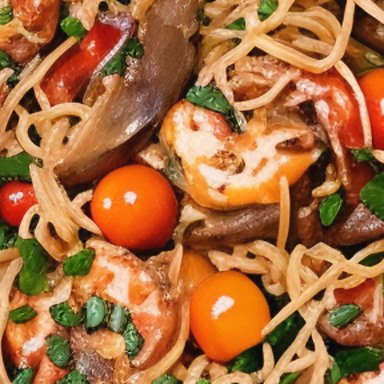}}\par\smallskip
DCAE $\to$ \ARCIR{}\par 0.053 bpp, 21.20\,dB\par \textbf{LPIPS 0.290}
\end{minipage}
 &
\begin{minipage}[t]{0.190\textwidth}\centering
\parbox[c][\linewidth][c]{\linewidth}{\centering\includegraphics[width=\linewidth,height=\linewidth,keepaspectratio]{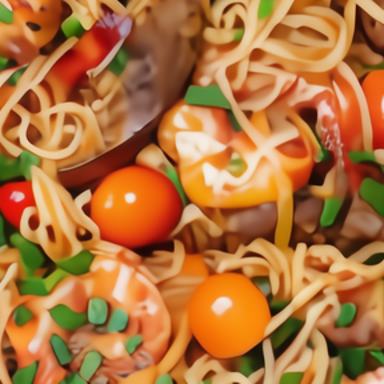}}\par\smallskip
\ARCIR{} $\to$ DCAE\par 0.127 bpp, 21.96\,dB\par LPIPS 0.342
\end{minipage}
 &
\begin{minipage}[t]{0.190\textwidth}\centering
\parbox[c][\linewidth][c]{\linewidth}{\centering\includegraphics[width=\linewidth,height=\linewidth,keepaspectratio]{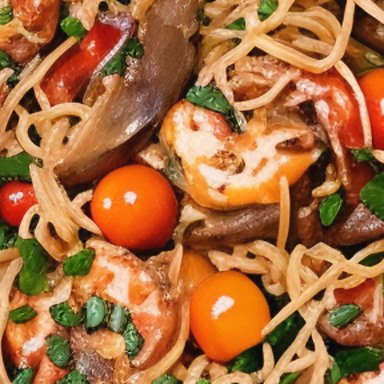}}\par\smallskip
\textbf{ARC (ours)}\par 0.053 bpp, 21.20\,dB\par \textbf{LPIPS 0.290}
\end{minipage}\\[3pt]
\begin{minipage}[t]{0.190\textwidth}\centering
\parbox[c][\linewidth][c]{\linewidth}{\centering\includegraphics[width=\linewidth,height=\linewidth,keepaspectratio]{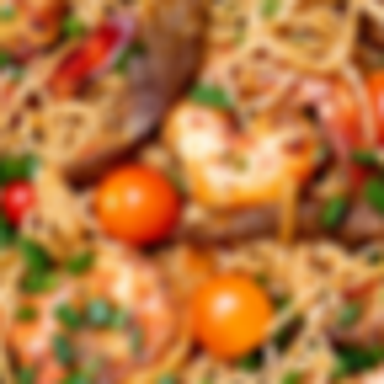}}\par\smallskip
Input crop\par Degraded\par \strut
\end{minipage}
 &
\begin{minipage}[t]{0.190\textwidth}\centering
\parbox[c][\linewidth][c]{\linewidth}{\centering\includegraphics[width=\linewidth,height=\linewidth,keepaspectratio]{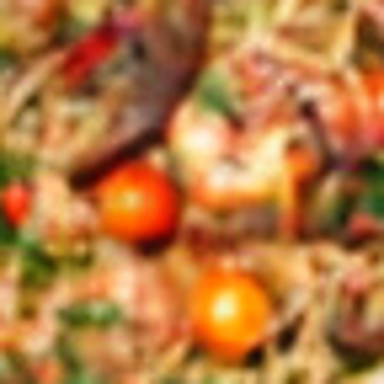}}\par\smallskip
CoD-Lite only\par \textbf{0.029 bpp}, 14.88\,dB\par LPIPS 0.655
\end{minipage}
 &
\begin{minipage}[t]{0.190\textwidth}\centering
\parbox[c][\linewidth][c]{\linewidth}{\centering\includegraphics[width=\linewidth,height=\linewidth,keepaspectratio]{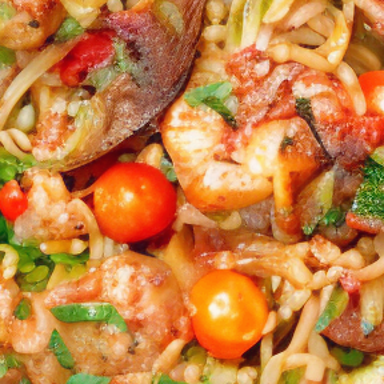}}\par\smallskip
CoD-Lite $\to$ \ARCIR{}\par 0.031 bpp, 18.27\,dB\par LPIPS 0.370
\end{minipage}
 &
\begin{minipage}[t]{0.190\textwidth}\centering
\parbox[c][\linewidth][c]{\linewidth}{\centering\includegraphics[width=\linewidth,height=\linewidth,keepaspectratio]{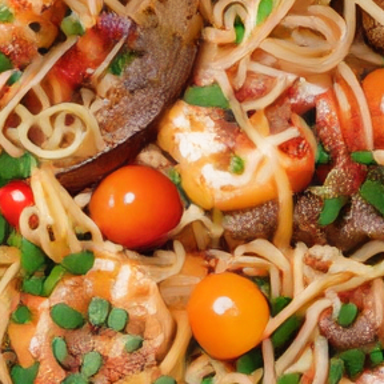}}\par\smallskip
\ARCIR{} $\to$ CoD-Lite\par 0.125 bpp, 21.42\,dB\par LPIPS 0.299
\end{minipage}
 &
\begin{minipage}[t]{0.190\textwidth}\centering
\parbox[c][\linewidth][c]{\linewidth}{\centering\includegraphics[width=\linewidth,height=\linewidth,keepaspectratio]{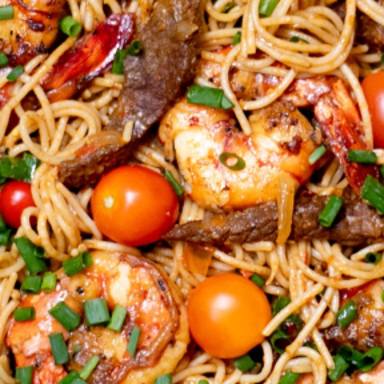}}\par\smallskip
Ground truth\par Clean reference\par \strut
\end{minipage}\\[3pt]
\multicolumn{5}{@{}l}{\rule{0pt}{4ex}\textbf{A3/098: defocus blur + JPEG artifacts}\quad Plan: JPEG removal (SwinIR) $\to$ \textbf{Compress (DCAE)} $\to$ Defocus deblur (IFAN) $\to$ Brighten (DiffPlugin)}\\[3pt]
\begin{minipage}[t]{0.190\textwidth}\centering
\parbox[c][\linewidth][c]{\linewidth}{\centering\includegraphics[width=\linewidth,height=\linewidth,keepaspectratio]{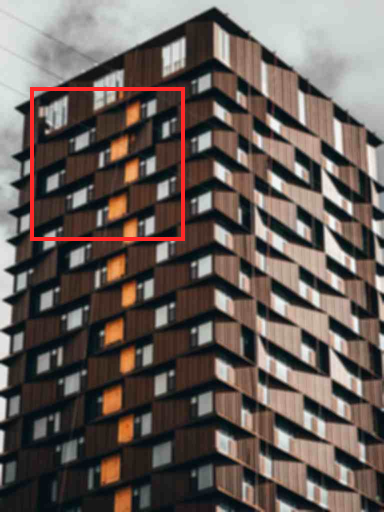}}\par\smallskip
Input\par Full frame\par \strut
\end{minipage}
 &
\begin{minipage}[t]{0.190\textwidth}\centering
\parbox[c][\linewidth][c]{\linewidth}{\centering\includegraphics[width=\linewidth,height=\linewidth,keepaspectratio]{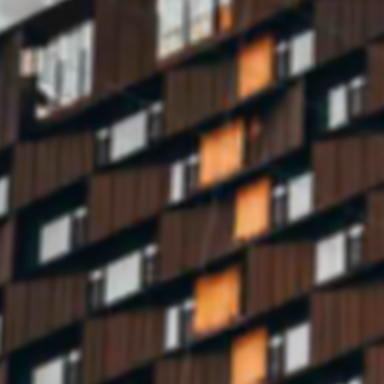}}\par\smallskip
DCAE only\par 0.209 bpp, \textbf{21.45\,dB}\par LPIPS 0.621
\end{minipage}
 &
\begin{minipage}[t]{0.190\textwidth}\centering
\parbox[c][\linewidth][c]{\linewidth}{\centering\includegraphics[width=\linewidth,height=\linewidth,keepaspectratio]{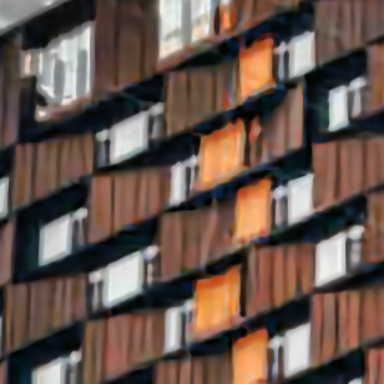}}\par\smallskip
DCAE $\to$ \ARCIR{}\par 0.209 bpp, 18.59\,dB\par LPIPS 0.403
\end{minipage}
 &
\begin{minipage}[t]{0.190\textwidth}\centering
\parbox[c][\linewidth][c]{\linewidth}{\centering\includegraphics[width=\linewidth,height=\linewidth,keepaspectratio]{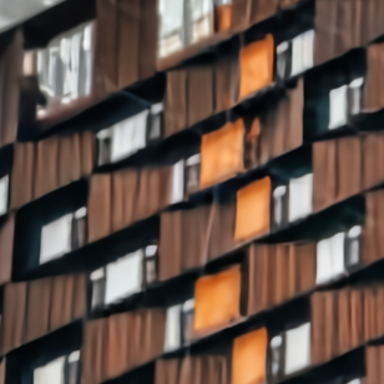}}\par\smallskip
\ARCIR{} $\to$ DCAE\par 0.185 bpp, 19.59\,dB\par LPIPS 0.368
\end{minipage}
 &
\begin{minipage}[t]{0.190\textwidth}\centering
\parbox[c][\linewidth][c]{\linewidth}{\centering\includegraphics[width=\linewidth,height=\linewidth,keepaspectratio]{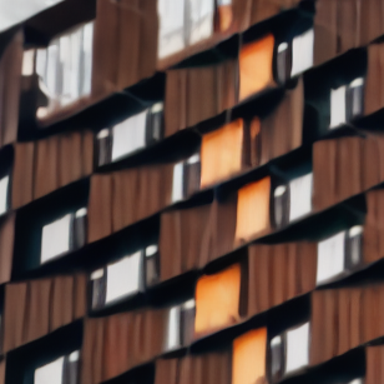}}\par\smallskip
\textbf{ARC (ours)}\par 0.201 bpp, 19.31\,dB\par \textbf{LPIPS 0.341}
\end{minipage}\\[3pt]
\begin{minipage}[t]{0.190\textwidth}\centering
\parbox[c][\linewidth][c]{\linewidth}{\centering\includegraphics[width=\linewidth,height=\linewidth,keepaspectratio]{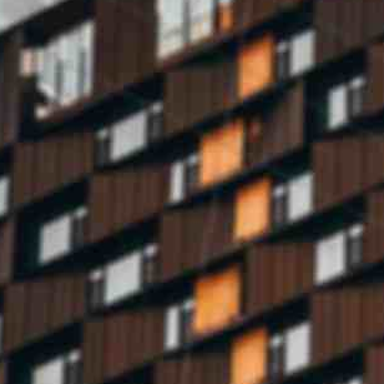}}\par\smallskip
Input crop\par Degraded\par \strut
\end{minipage}
 &
\begin{minipage}[t]{0.190\textwidth}\centering
\parbox[c][\linewidth][c]{\linewidth}{\centering\includegraphics[width=\linewidth,height=\linewidth,keepaspectratio]{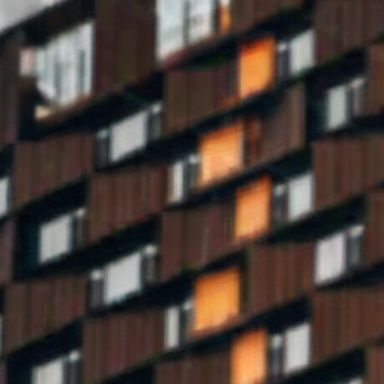}}\par\smallskip
CoD-Lite only\par \textbf{0.118 bpp}, 21.13\,dB\par LPIPS 0.561
\end{minipage}
 &
\begin{minipage}[t]{0.190\textwidth}\centering
\parbox[c][\linewidth][c]{\linewidth}{\centering\includegraphics[width=\linewidth,height=\linewidth,keepaspectratio]{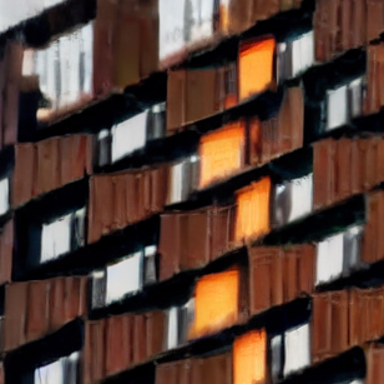}}\par\smallskip
CoD-Lite $\to$ \ARCIR{}\par 0.125 bpp, 18.85\,dB\par LPIPS 0.413
\end{minipage}
 &
\begin{minipage}[t]{0.190\textwidth}\centering
\parbox[c][\linewidth][c]{\linewidth}{\centering\includegraphics[width=\linewidth,height=\linewidth,keepaspectratio]{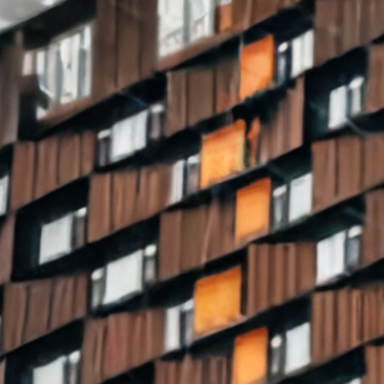}}\par\smallskip
\ARCIR{} $\to$ CoD-Lite\par 0.125 bpp, 19.41\,dB\par LPIPS 0.365
\end{minipage}
 &
\begin{minipage}[t]{0.190\textwidth}\centering
\parbox[c][\linewidth][c]{\linewidth}{\centering\includegraphics[width=\linewidth,height=\linewidth,keepaspectratio]{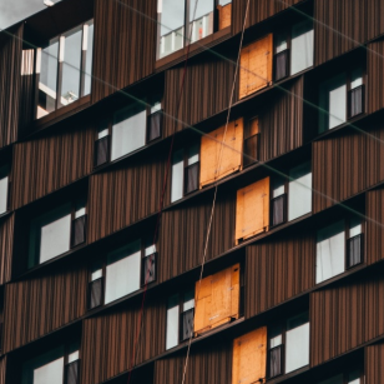}}\par\smallskip
Ground truth\par Clean reference\par \strut
\end{minipage}\\[3pt]
\end{tabular}
\endgroup
\caption{\textbf{Qualitative comparisons on MiO100, further examples.} \textbf{Bold}: lowest bpp, highest PSNR and lowest LPIPS per example. Each example shows the DCAE baselines in the first row and the CoD-Lite baselines in the second, with ARC directly above the ground truth. The row header gives the plan ARC executed, including the codec selected for the compression step. Low-resolution input crops are upsampled bicubically for display. Image B1/063 is an example where ARC chooses to put the compression step first, and therefore performs the same as DCAE $\to$ \ARCIR{} in this case.}
\label{fig:qualitative_comparison_v3_grid_b}
\end{figure*}

\begin{figure*}[t]
\centering
\begingroup
\setlength{\tabcolsep}{1.5pt}
\footnotesize
\begin{tabular}{@{}*{5}{c}@{}}
\multicolumn{5}{@{}l}{\rule{0pt}{2.5ex}\textbf{A4/053: motion blur + dark}\quad Plan: Motion deblur (MAXIM) $\to$ \textbf{Compress (CoD-Lite)} $\to$ Brighten (Gamma Correction)}\\[3pt]
\begin{minipage}[t]{0.190\textwidth}\centering
\parbox[c][\linewidth][c]{\linewidth}{\centering\includegraphics[width=\linewidth,height=\linewidth,keepaspectratio]{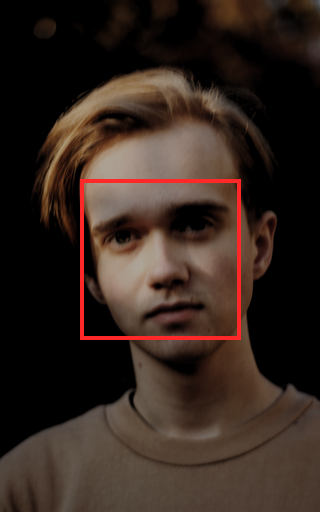}}\par\smallskip
Input\par Full frame\par \strut
\end{minipage}
 &
\begin{minipage}[t]{0.190\textwidth}\centering
\parbox[c][\linewidth][c]{\linewidth}{\centering\includegraphics[width=\linewidth,height=\linewidth,keepaspectratio]{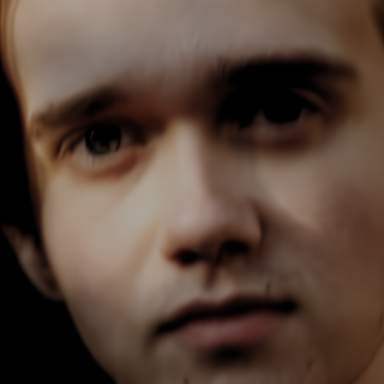}}\par\smallskip
DCAE only\par 0.073 bpp, 18.91\,dB\par LPIPS 0.289
\end{minipage}
 &
\begin{minipage}[t]{0.190\textwidth}\centering
\parbox[c][\linewidth][c]{\linewidth}{\centering\includegraphics[width=\linewidth,height=\linewidth,keepaspectratio]{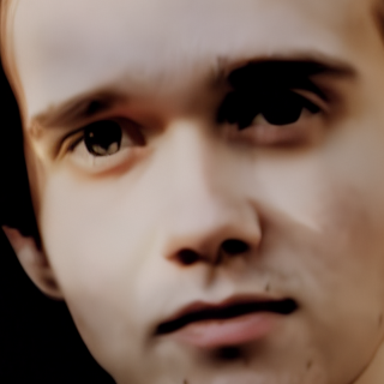}}\par\smallskip
DCAE $\to$ \ARCIR{}\par 0.073 bpp, 23.72\,dB\par LPIPS 0.261
\end{minipage}
 &
\begin{minipage}[t]{0.190\textwidth}\centering
\parbox[c][\linewidth][c]{\linewidth}{\centering\includegraphics[width=\linewidth,height=\linewidth,keepaspectratio]{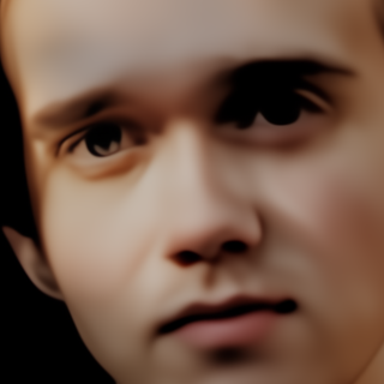}}\par\smallskip
\ARCIR{} $\to$ DCAE\par \textbf{0.025 bpp}, 23.68\,dB\par LPIPS 0.273
\end{minipage}
 &
\begin{minipage}[t]{0.190\textwidth}\centering
\parbox[c][\linewidth][c]{\linewidth}{\centering\includegraphics[width=\linewidth,height=\linewidth,keepaspectratio]{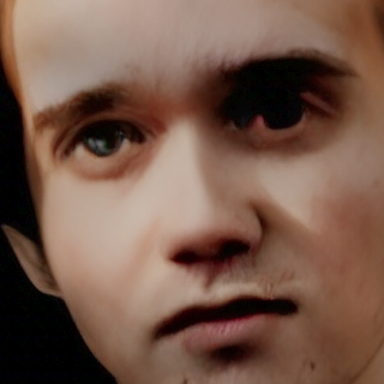}}\par\smallskip
\textbf{ARC (ours)}\par 0.031 bpp, \textbf{23.78\,dB}\par \textbf{LPIPS 0.205}
\end{minipage}\\[3pt]
\begin{minipage}[t]{0.190\textwidth}\centering
\parbox[c][\linewidth][c]{\linewidth}{\centering\includegraphics[width=\linewidth,height=\linewidth,keepaspectratio]{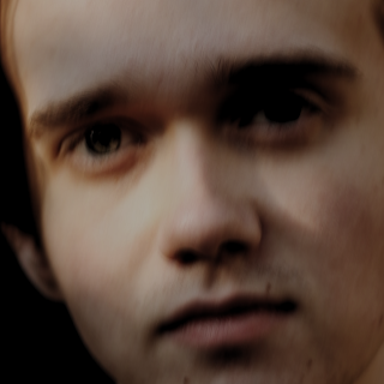}}\par\smallskip
Input crop\par Degraded\par \strut
\end{minipage}
 &
\begin{minipage}[t]{0.190\textwidth}\centering
\parbox[c][\linewidth][c]{\linewidth}{\centering\includegraphics[width=\linewidth,height=\linewidth,keepaspectratio]{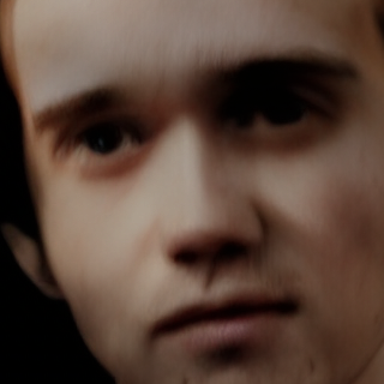}}\par\smallskip
CoD-Lite only\par 0.029 bpp, 18.93\,dB\par LPIPS 0.244
\end{minipage}
 &
\begin{minipage}[t]{0.190\textwidth}\centering
\parbox[c][\linewidth][c]{\linewidth}{\centering\includegraphics[width=\linewidth,height=\linewidth,keepaspectratio]{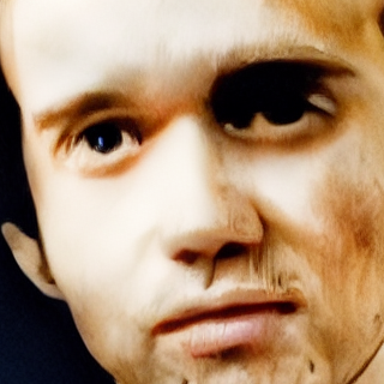}}\par\smallskip
CoD-Lite $\to$ \ARCIR{}\par 0.031 bpp, 17.38\,dB\par LPIPS 0.432
\end{minipage}
 &
\begin{minipage}[t]{0.190\textwidth}\centering
\parbox[c][\linewidth][c]{\linewidth}{\centering\includegraphics[width=\linewidth,height=\linewidth,keepaspectratio]{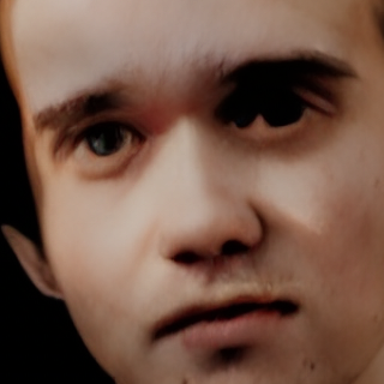}}\par\smallskip
\ARCIR{} $\to$ CoD-Lite\par 0.031 bpp, 23.42\,dB\par LPIPS 0.209
\end{minipage}
 &
\begin{minipage}[t]{0.190\textwidth}\centering
\parbox[c][\linewidth][c]{\linewidth}{\centering\includegraphics[width=\linewidth,height=\linewidth,keepaspectratio]{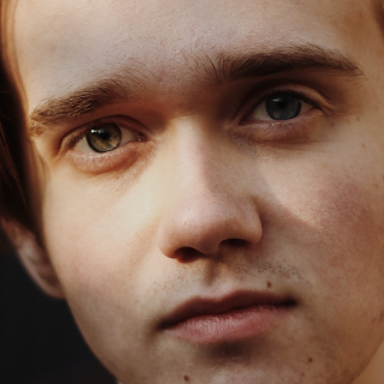}}\par\smallskip
Ground truth\par Clean reference\par \strut
\end{minipage}\\[3pt]
\multicolumn{5}{@{}l}{\rule{0pt}{4ex}\textbf{A4/080: motion blur + dark}\quad Plan: Motion deblur (Restormer) $\to$ \textbf{Compress (CoD-Lite)} $\to$ Brighten (CLAHE)}\\[3pt]
\begin{minipage}[t]{0.190\textwidth}\centering
\parbox[c][\linewidth][c]{\linewidth}{\centering\includegraphics[width=\linewidth,height=\linewidth,keepaspectratio]{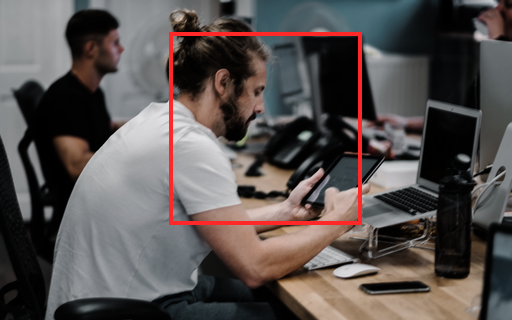}}\par\smallskip
Input\par Full frame\par \strut
\end{minipage}
 &
\begin{minipage}[t]{0.190\textwidth}\centering
\parbox[c][\linewidth][c]{\linewidth}{\centering\includegraphics[width=\linewidth,height=\linewidth,keepaspectratio]{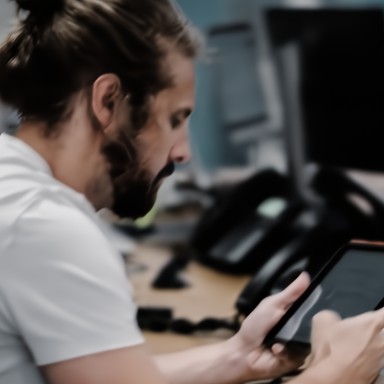}}\par\smallskip
DCAE only\par \textbf{0.097 bpp}, 20.00\,dB\par LPIPS 0.202
\end{minipage}
 &
\begin{minipage}[t]{0.190\textwidth}\centering
\parbox[c][\linewidth][c]{\linewidth}{\centering\includegraphics[width=\linewidth,height=\linewidth,keepaspectratio]{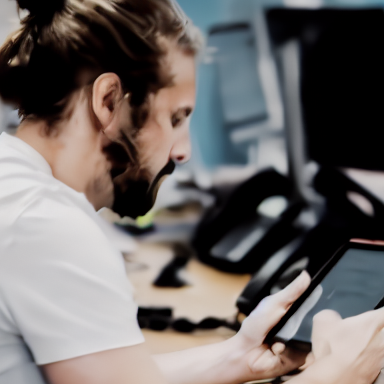}}\par\smallskip
DCAE $\to$ \ARCIR{}\par \textbf{0.097 bpp}, \textbf{24.75\,dB}\par LPIPS 0.169
\end{minipage}
 &
\begin{minipage}[t]{0.190\textwidth}\centering
\parbox[c][\linewidth][c]{\linewidth}{\centering\includegraphics[width=\linewidth,height=\linewidth,keepaspectratio]{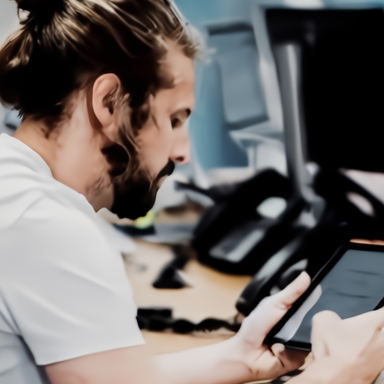}}\par\smallskip
\ARCIR{} $\to$ DCAE\par 0.115 bpp, 23.36\,dB\par LPIPS 0.174
\end{minipage}
 &
\begin{minipage}[t]{0.190\textwidth}\centering
\parbox[c][\linewidth][c]{\linewidth}{\centering\includegraphics[width=\linewidth,height=\linewidth,keepaspectratio]{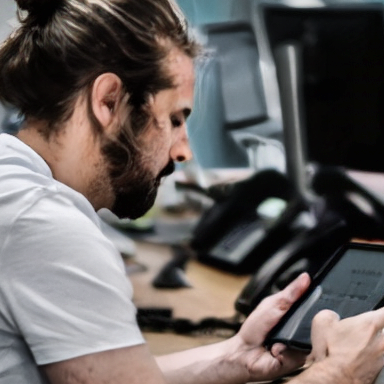}}\par\smallskip
\textbf{ARC (ours)}\par 0.125 bpp, 22.61\,dB\par LPIPS 0.161
\end{minipage}\\[3pt]
\begin{minipage}[t]{0.190\textwidth}\centering
\parbox[c][\linewidth][c]{\linewidth}{\centering\includegraphics[width=\linewidth,height=\linewidth,keepaspectratio]{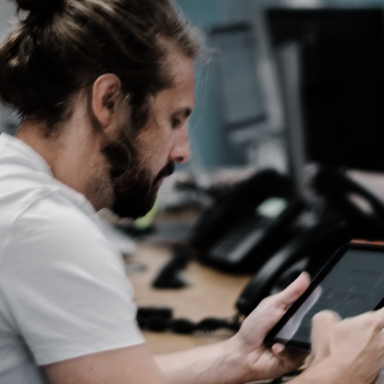}}\par\smallskip
Input crop\par Degraded\par \strut
\end{minipage}
 &
\begin{minipage}[t]{0.190\textwidth}\centering
\parbox[c][\linewidth][c]{\linewidth}{\centering\includegraphics[width=\linewidth,height=\linewidth,keepaspectratio]{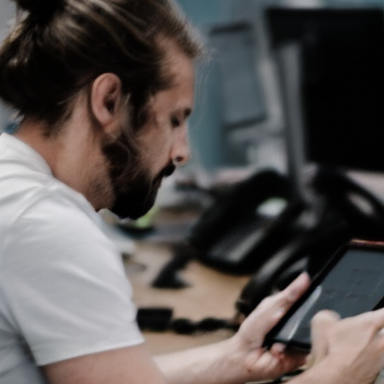}}\par\smallskip
CoD-Lite only\par 0.118 bpp, 20.09\,dB\par LPIPS 0.148
\end{minipage}
 &
\begin{minipage}[t]{0.190\textwidth}\centering
\parbox[c][\linewidth][c]{\linewidth}{\centering\includegraphics[width=\linewidth,height=\linewidth,keepaspectratio]{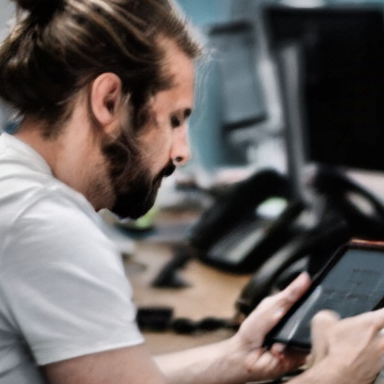}}\par\smallskip
CoD-Lite $\to$ \ARCIR{}\par 0.125 bpp, 23.07\,dB\par LPIPS 0.154
\end{minipage}
 &
\begin{minipage}[t]{0.190\textwidth}\centering
\parbox[c][\linewidth][c]{\linewidth}{\centering\includegraphics[width=\linewidth,height=\linewidth,keepaspectratio]{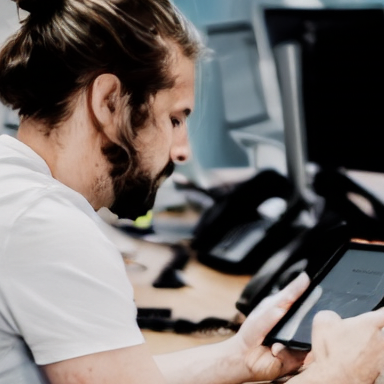}}\par\smallskip
\ARCIR{} $\to$ CoD-Lite\par 0.125 bpp, 23.13\,dB\par \textbf{LPIPS 0.137}
\end{minipage}
 &
\begin{minipage}[t]{0.190\textwidth}\centering
\parbox[c][\linewidth][c]{\linewidth}{\centering\includegraphics[width=\linewidth,height=\linewidth,keepaspectratio]{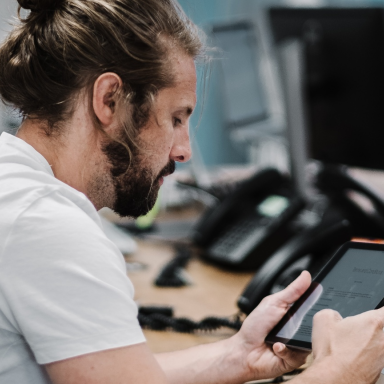}}\par\smallskip
Ground truth\par Clean reference\par \strut
\end{minipage}\\[3pt]
\end{tabular}
\endgroup
\caption{\textbf{Qualitative comparisons on MiO100, further examples.} \textbf{Bold}: lowest bpp, highest PSNR and lowest LPIPS per example. Each example shows the DCAE baselines in the first row and the CoD-Lite baselines in the second, with ARC directly above the ground truth. The row header gives the plan ARC executed, including the codec selected for the compression step. For image A4/080 the fixed pipeline using \ARCIR{} and CoD-Lite outperforms the plan ARC executes w.r.t. PSNR and LPIPS.}
\label{fig:qualitative_comparison_v3_grid_c}
\end{figure*}

\begin{figure*}[t]
\centering
\begingroup
\setlength{\tabcolsep}{1.5pt}
\footnotesize
\begin{tabular}{@{}*{5}{c}@{}}
\multicolumn{5}{@{}l}{\rule{0pt}{2.5ex}\textbf{A1/035: dark + noise}\quad Plan: Denoise (Restormer) $\to$ \textbf{Compress (CoD-Lite)}}\\[3pt]
\begin{minipage}[t]{0.190\textwidth}\centering
\parbox[c][\linewidth][c]{\linewidth}{\centering\includegraphics[width=\linewidth,height=\linewidth,keepaspectratio]{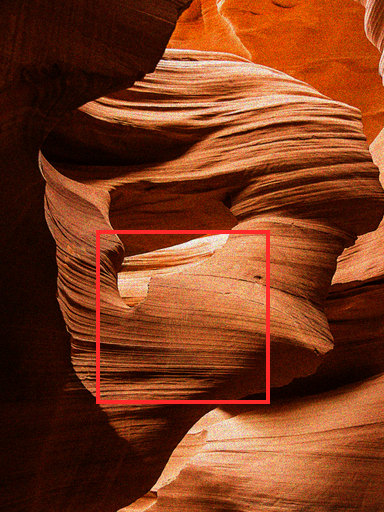}}\par\smallskip
Input\par Full frame\par \strut
\end{minipage}
 &
\begin{minipage}[t]{0.190\textwidth}\centering
\parbox[c][\linewidth][c]{\linewidth}{\centering\includegraphics[width=\linewidth,height=\linewidth,keepaspectratio]{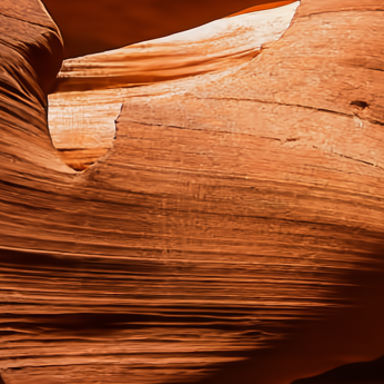}}\par\smallskip
DCAE only\par 0.173 bpp, 23.76\,dB\par LPIPS 0.219
\end{minipage}
 &
\begin{minipage}[t]{0.190\textwidth}\centering
\parbox[c][\linewidth][c]{\linewidth}{\centering\includegraphics[width=\linewidth,height=\linewidth,keepaspectratio]{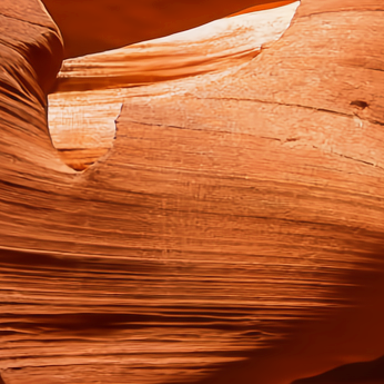}}\par\smallskip
DCAE $\to$ \ARCIR{}\par 0.173 bpp, \textbf{31.32\,dB}\par \textbf{LPIPS 0.169}
\end{minipage}
 &
\begin{minipage}[t]{0.190\textwidth}\centering
\parbox[c][\linewidth][c]{\linewidth}{\centering\includegraphics[width=\linewidth,height=\linewidth,keepaspectratio]{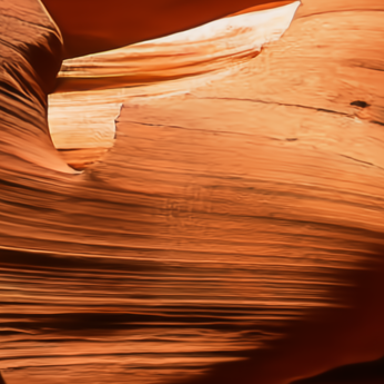}}\par\smallskip
\ARCIR{} $\to$ DCAE\par 0.161 bpp, 25.38\,dB\par LPIPS 0.237
\end{minipage}
 &
\begin{minipage}[t]{0.190\textwidth}\centering
\parbox[c][\linewidth][c]{\linewidth}{\centering\includegraphics[width=\linewidth,height=\linewidth,keepaspectratio]{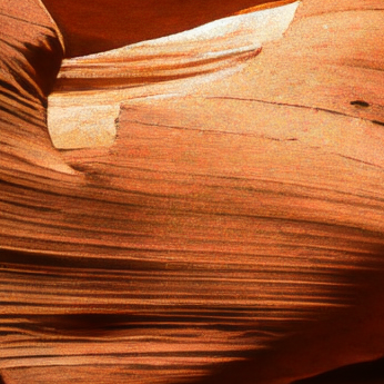}}\par\smallskip
\textbf{ARC (ours)}\par 0.125 bpp, 24.48\,dB\par LPIPS 0.266
\end{minipage}\\[3pt]
\begin{minipage}[t]{0.190\textwidth}\centering
\parbox[c][\linewidth][c]{\linewidth}{\centering\includegraphics[width=\linewidth,height=\linewidth,keepaspectratio]{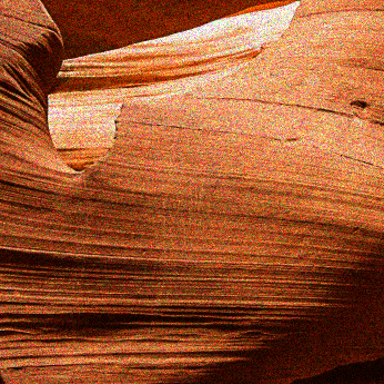}}\par\smallskip
Input crop\par Degraded\par \strut
\end{minipage}
 &
\begin{minipage}[t]{0.190\textwidth}\centering
\parbox[c][\linewidth][c]{\linewidth}{\centering\includegraphics[width=\linewidth,height=\linewidth,keepaspectratio]{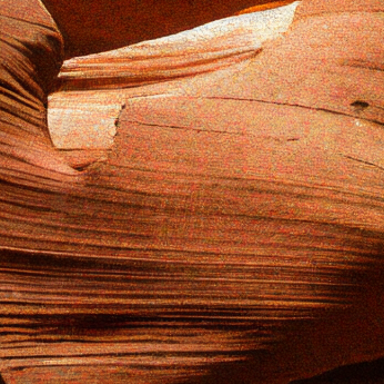}}\par\smallskip
CoD-Lite only\par \textbf{0.120 bpp}, 22.55\,dB\par LPIPS 0.453
\end{minipage}
 &
\begin{minipage}[t]{0.190\textwidth}\centering
\parbox[c][\linewidth][c]{\linewidth}{\centering\includegraphics[width=\linewidth,height=\linewidth,keepaspectratio]{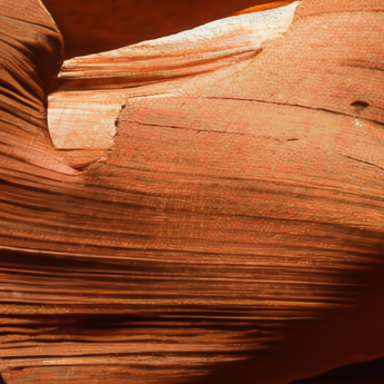}}\par\smallskip
CoD-Lite $\to$ \ARCIR{}\par 0.125 bpp, 23.12\,dB\par LPIPS 0.240
\end{minipage}
 &
\begin{minipage}[t]{0.190\textwidth}\centering
\parbox[c][\linewidth][c]{\linewidth}{\centering\includegraphics[width=\linewidth,height=\linewidth,keepaspectratio]{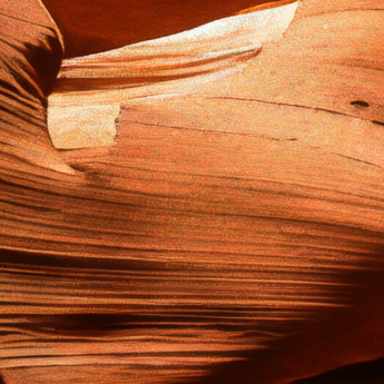}}\par\smallskip
\ARCIR{} $\to$ CoD-Lite\par 0.125 bpp, 24.48\,dB\par LPIPS 0.265
\end{minipage}
 &
\begin{minipage}[t]{0.190\textwidth}\centering
\parbox[c][\linewidth][c]{\linewidth}{\centering\includegraphics[width=\linewidth,height=\linewidth,keepaspectratio]{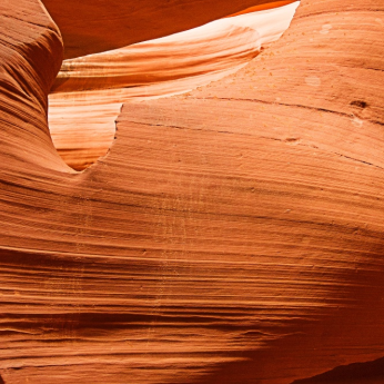}}\par\smallskip
Ground truth\par Clean reference\par \strut
\end{minipage}\\[3pt]
\end{tabular}
\endgroup
\caption{\textbf{Qualitative comparison on MiO100, failure case.} \textbf{Bold}: lowest bpp, highest PSNR and lowest LPIPS per example. Each example shows the DCAE baselines in the first row and the CoD-Lite baselines in the second, with ARC directly above the ground truth. The row header gives the plan ARC executed, including the codec selected for the compression step. Low-resolution input crops are upsampled bicubically for display. ARC's plan for A1/035 contains no brightening step, so its output stays darker than the reference, which accounts for most of its PSNR gap to DCAE $\to$ \ARCIR{}.}
\label{fig:qualitative_comparison_v3_grid_d}
\end{figure*}

Figures~\ref{fig:qualitative_comparison_v3_grid}--\ref{fig:qualitative_comparison_v3_grid_d} compare integrated ARC with each codec alone and both fixed orderings of the codec and restoration-only \ARCIR{}, for DCAE and CoD-Lite, on seven MiO100 examples, together with the plan ARC executed for each. The position of the compression step varies with the detected degradations according to the priorities detailed in Section~\ref{sec:toolbox} and potential rollbacks. Figure~\ref{fig:qualitative_comparison_v3_grid_d} shows a failure case where our approach is outperformed by a different tool order which uses compression first.

\section{Runtime}\label{sec:runtime}

Complementing the tool-call counts in Section~\ref{sec:overhead}, Table~\ref{tab:runtime} reports measured encoding times of ARC on MiO100. We measure these timings separately on a single GPU (NVIDIA RTX 4090).
The timings exclude \mbox{DepictQA} perception, which was run separately.
Over 80 images (five per subset), restoration and encoding takes 5.7 minutes per image on average. The large standard deviation comes mainly from images whose plan was rolled back due to the reflection threshold.
A compression step evaluates about 11 codec candidates which takes about 115\,s. Restoration calls vary strongly by tool: the median call takes 4.3\,s, while diffusion-based tools such as DiffBIR take about 80\,s.

For decoding, only the decompression step and a small number of restoration steps are needed, which takes 23.6\,s on average per image, most of which are restoration steps.

Checkpoint loading from HDD makes up a relevant fraction of the runtime and likely adds to the variance of the measured times.

\begin{table*}[ht]
    \centering
    \footnotesize
    \setlength{\tabcolsep}{3pt}
    \begin{tabular}{lcccc}
        \toprule
        Time (s) & $n$ & Mean & Std. & Median \\
        \midrule
        Full encoding, per image & 80 & 343.3 & 109.4 & 336.9 \\
        Restoration step, per call & 1027 & 11.5 & 17.6 & 4.3 \\
        All restoration steps, per image & 80 & 147.3 & 88.8 & 138.5 \\
        Compression step (codec-quality search) & 82$^\dagger$ & 114.7 & 33.4 & 112.9 \\
        \midrule
        Full decoding, per image & 80 & 23.6 & 27.9 & 13.1 \\
        Codec decoding, per image & 80 & 4.7 & 3.3 & 4.2 \\
        Restoration after decoding, per image & 80 & 18.9 & 28.1 & 9.0 \\
        \bottomrule
    \end{tabular}
    \caption{\textbf{Measured encoding and decoding time of ARC} on 80 MiO100 images (5 per subset). $n$ counts images, tool calls or compression steps. $^\dagger$ The number of codec-quality search steps can vary due to rollback.}\label{tab:runtime}
\end{table*}

\section{Exhaustive Path Search}\label{sec:oracle}

To separate the limits of greedy search from those of the reward itself, we exhaustively evaluate every restoration order, tool combination, and compression position on five images each of MiO100 subsets C1 and C3, and select the best complete path for each objective in hindsight (Table~\ref{tab:oracle_paths}).
The compression is fixed to DCAE using the checkpoint closest to achieving 0.2\,bpp to limit the search space. We exclude paths that exceed 0.3\,bpp.

Fully greedy search on the agent's reflection score $R$ and ARC's selection are compared against oracles on the full-reference metrics PSNR, LPIPS and DISTS. These hindsight oracles use ground truth and are not deployable baselines.

No single objective is best on all metrics. Oracles on $R$ (the reflection score) or $Q$ (the no-reference metric weighted average) reach the best no-reference scores (MUSIQ, NIQE) but not the best fidelity; on C3, selecting on $Q$ alone drops PSNR to 14.18\,dB, showing that a purely no-reference reward can be exploited. The PSNR oracle gives the highest PSNR in both subsets but the worst MUSIQ and NIQE. Placement of compression also depends on the objective: in the example paths, the PSNR oracle compresses last, whereas the $R$, $Q$ and LPIPS oracles compress mid-path and restore after decoding.
Replaying ARC's mean-rank tool selection on the same trees improves PSNR and LPIPS over fully greedy search on $R$ in both subsets (19.78 vs.\ 18.49\,dB on C1, 20.85 vs.\ 20.26\,dB on C3), but stays well below the fidelity oracles.

The oracle results suggest that the agent's greedy search, as well as its reliance on no-reference metrics, are likely limitations of its performance. Future work could explore more sophisticated search strategies and metrics that better correlate with full-reference quality.

\begin{table*}[t]
    \centering
    \footnotesize
    \setlength{\tabcolsep}{3.5pt}
    \begin{tabular}{l*{3}{>{\columncolor{frbg}}c}*{2}{>{\columncolor{nrbg}}c}c*{3}{>{\columncolor{frbg}}c}*{2}{>{\columncolor{nrbg}}c}c}
        \toprule
        & \multicolumn{6}{c}{Group C1} & \multicolumn{6}{c}{Group C3} \\
        \cmidrule(lr){2-7} \cmidrule(lr){8-13}
        Path & PSNR $\uparrow$ & LPIPS $\downarrow$ & DISTS $\downarrow$ & MUSIQ $\uparrow$ & NIQE $\downarrow$ & BPP & PSNR $\uparrow$ & LPIPS $\downarrow$ & DISTS $\downarrow$ & MUSIQ $\uparrow$ & NIQE $\downarrow$ & BPP \\
        \midrule
        Fully greedy on $R$ & 18.49 & 0.589 & 0.294 & 44.09 & 6.438 & 0.192 & 20.26 & 0.567 & \textit{0.251} & 46.18 & 5.285 & 0.213 \\
        ARC selection & \textit{19.78} & 0.540 & 0.275 & \textit{44.74} & 6.379 & 0.145 & 20.85 & \textit{0.547} & 0.281 & 50.23 & 5.738 & 0.182 \\
        \midrule
        Oracle on $R$ & 19.01 & \textit{0.525} & 0.268 & \textbf{51.97} & \textbf{5.167} & 0.213 & 17.67 & 0.581 & 0.293 & \textbf{60.46} & \textbf{4.273} & 0.235 \\
        Oracle on $Q$ & 19.01 & \textit{0.525} & 0.268 & \textbf{51.97} & \textbf{5.167} & 0.213 & 14.18 & 0.757 & 0.386 & \underline{57.93} & \underline{4.435} & 0.238 \\
        Oracle on PSNR & \textbf{23.87} & 0.560 & \textit{0.265} & 37.72 & 7.119 & 0.202 & \textbf{23.42} & \textit{0.547} & 0.269 & 35.67 & 7.593 & 0.142 \\
        Oracle on LPIPS & 18.66 & \textbf{0.456} & \underline{0.242} & \underline{46.30} & \underline{5.406} & 0.171 & \underline{21.41} & \textbf{0.442} & \underline{0.249} & 50.65 & 5.308 & 0.195 \\
        Oracle on DISTS & \underline{21.01} & \underline{0.496} & \textbf{0.226} & 42.01 & \textit{5.771} & 0.216 & \textit{21.30} & \underline{0.503} & \textbf{0.201} & \textit{51.55} & \textit{4.788} & 0.229 \\
        \midrule
        & \multicolumn{6}{c}{Selected path, C1/001} & \multicolumn{6}{c}{Selected path, C3/001} \\
        \cmidrule(lr){2-7} \cmidrule(lr){8-13}
        Fully greedy on $R$ & \multicolumn{6}{c}{defocus $\to$ brightening $\to$ JPEG $\to$ DCAE} & \multicolumn{6}{c}{motion $\to$ noise $\to$ defocus $\to$ DCAE} \\
        ARC selection & \multicolumn{6}{c}{JPEG $\to$ DCAE $\to$ defocus $\to$ brightening} & \multicolumn{6}{c}{noise $\to$ motion $\to$ DCAE $\to$ defocus} \\
        Oracle on $R$ & \multicolumn{6}{c}{JPEG $\to$ defocus $\to$ DCAE $\to$ brightening} & \multicolumn{6}{c}{motion $\to$ noise $\to$ DCAE $\to$ defocus} \\
        Oracle on $Q$ & \multicolumn{6}{c}{JPEG $\to$ defocus $\to$ DCAE $\to$ brightening} & \multicolumn{6}{c}{motion $\to$ noise $\to$ DCAE $\to$ defocus} \\
        Oracle on PSNR & \multicolumn{6}{c}{defocus $\to$ JPEG $\to$ brightening $\to$ DCAE} & \multicolumn{6}{c}{defocus $\to$ noise $\to$ motion $\to$ DCAE} \\
        Oracle on LPIPS & \multicolumn{6}{c}{JPEG $\to$ DCAE $\to$ defocus $\to$ brightening} & \multicolumn{6}{c}{noise $\to$ DCAE $\to$ motion $\to$ defocus} \\
        Oracle on DISTS & \multicolumn{6}{c}{DCAE $\to$ JPEG $\to$ defocus $\to$ brightening} & \multicolumn{6}{c}{motion $\to$ noise $\to$ defocus $\to$ DCAE} \\
        \bottomrule
    \end{tabular}
    \caption{\textbf{Exhaustive search over restoration orders, tools, and the compression position} on five images each of MiO100 subsets C1 and C3, with the degradations known and no super-resolution. Each oracle selects the best complete path per image in hindsight. Bold, underlined, and italic values mark the best, second-best, and third-best distinct values for each quality metric within each subset.}\label{tab:oracle_paths}
\end{table*}

\end{document}